%% file: preprint.tex
\documentclass{article} %
\usepackage{iclr2027_conference,times}

\input{math_commands.tex}

\usepackage{hyperref}
\usepackage{url}
\usepackage{algorithm}
\usepackage{algpseudocode}
\usepackage{graphicx} 
\usepackage{multirow}
\usepackage{booktabs}
\usepackage{diagbox}
\usepackage{amsmath}
\usepackage{xcolor}
\usepackage{amsthm}
\usepackage{bbm}
\usepackage{subcaption}
\usepackage{thmtools, thm-restate}
\usepackage{amssymb}
\usepackage{enumitem}
\usepackage{nicefrac}
\usepackage{xcolor}
\usepackage{xpatch}

\usepackage[T1]{fontenc}

\usepackage{booktabs,graphicx}
\usepackage{tabularx}
\usepackage{longtable}
\usepackage{placeins}

\title{D-Scope: Decomposing and Steering Diffusion Transformers with Sparse Autoencoders}

\author{%
\hspace*{-\tabcolsep}%
\parbox{\textwidth}{%
\centering
{\normalsize\bfseries
Xinyue Xu$^{1}$\thanks{Equal contribution.}\quad
Jiahao Zhang$^{2*}$\quad
Lijie Hu$^{2}$\thanks{Corresponding authors.}\quad
Peter Hase$^{3,4}$\quad
Hao Wang$^{5\dagger}$}\\[0.5em]
{\normalfont\small
$^{1}$Pivotal Research\quad
$^{2}$Mohamed bin Zayed University of Artificial Intelligence\\[0.15em]
$^{3}$Schmidt Sciences\quad
$^{4}$Stanford University\quad
$^{5}$University of Illinois at Urbana-Champaign}
}%
\hspace*{-\tabcolsep}%
}

\definecolor{newblue}{HTML}{2E75B6}

\iclrfinalcopy

\begin{document}

\maketitle
\lhead{Preprint}

\begin{abstract}
Sparse autoencoders (SAEs) reveal visual structure in diffusion transformers (DiTs), but interpreting a feature does not establish whether it can be used to control generation. We introduce \textbf{D-Scope (Diffusion Scope)}, a framework that connects feature interpretation to generation control through shared visual evidence. 
D-Scope aggregates SigLIP~2 embeddings of highly activating image patches into visual centroids. 
Matching target text descriptions against these visual centroids in the shared image-text embedding space then enables retrieval of individual features without per-feature text annotations. 
The underlying patches provide evidence for inspecting each selection, while spatially masked interventions test the corresponding decoder direction at varying strengths under fixed generation conditions. 
We characterize 150 SAEs across two model families and five layers, and introduce a benchmark of 100 target concepts with ten contexts each spanning under-specified and explicit-conflict conditions. Our empirical results show that high reconstruction fidelity can coexist with low dictionary utilization and limited visual-evidence coverage. Under per-case best-of-sweep strength selection, contrastive retrieval yields larger mean regional SigLIP~2 gains than direct retrieval across the tested steering configurations, without consistently improving outside-region preservation. 
D-Scope provides an inspectable framework for evaluating sparse DiT features through their visual evidence and the effects of their decoder directions on generation.
The demo is available at \url{https://jiahaozhang-public.github.io/d-scope/}.
\end{abstract}

\section{Introduction}
\label{sec:introduction}

\input{figs_compressed/fig1_teaser/fig1_teaser}

Understanding how visual concepts are encoded in diffusion transformers (DiTs)~\citep{peebles2023scalable} is important for explaining their generative behavior and for identifying interventions that change specific aspects of an image. Sparse autoencoders (SAEs) provide a direct way to examine these representations by decomposing dense activations into sparse combinations of learned features, each of which can be interpreted through its activating patches and used for intervention through its decoder direction. Prior work has shown that SAE features in diffusion models capture interpretable visual concepts~\citep{surkov2024one,shabalin2025interpreting,Tnaz2025EmergenceAE}, and that their decoder directions can steer generation, transfer visual attributes, and suppress concepts~\citep{surkov2024one,shabalin2025interpreting,cywinski2025saeuron,kim2025concept}. However, applying these capabilities to a specific, fine-grained image edit requires identifying an appropriate decoder direction from a large SAE dictionary.

Existing approaches select or interpret relevant SAE features primarily through textual descriptions or comparisons of paired activations. In the former, each feature is described based on the visual patterns shared by its highly activating images, either manually or using a vision-language model~\citep{surkov2024one,shabalin2025interpreting}. These descriptions can then be used to match a desired edit to candidate features, but require an additional annotation or description-generation step. Alternatively, relevant features can be identified by comparing SAE activations between paired source and target generations~\citep{surkov2024one}. This avoids explicit feature descriptions, but requires a suitable target generation and access to its internal activations. Thus, both approaches introduce additional information or processing beyond the source image itself when selecting a direction for a desired edit.

In this paper, we consider a cleaner setting in which a user specifies the desired fine-grained change through a text query such as ``red dress'' (Fig.~\ref{fig:teaser}).
The goal is to modify the queried attribute while preserving other image content.
Feature selection has access to a pretrained SAE dictionary and its activating image patches, \emph{without} relying on per-feature textual descriptions or target-generation activations like existing work.
This leads to our \textbf{central question:} \textit{can a target text query retrieve a relevant SAE decoder direction directly from the visual evidence associated with each feature?}

\begin{figure}[!ht]
    \centering
    \includegraphics[width=\linewidth]{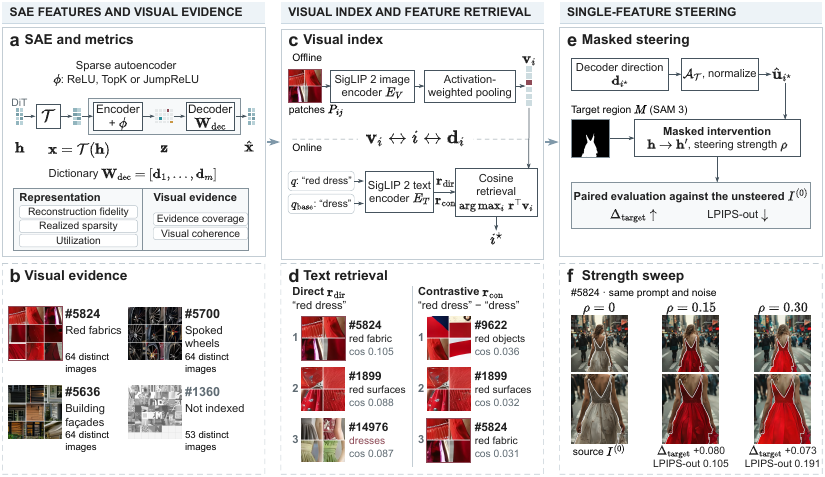}
    \caption{
    \textbf{D-Scope links the visual evidence of each SAE feature to its decoder direction.}
    Columns follow Sec.~\ref{sec:sae_features}--\ref{sec:single_feature_steering}; bottom panels show examples.
    \textbf{(a)} SAE training and dictionary metrics.
    \textbf{(b)} Activating patches; \#1360 spans too few distinct source images to be indexed.
    \textbf{(c)} Offline: patches are pooled into a visual centroid $\mathbf{v}_i$. Online: a target query, used directly or contrasted with a reference query, retrieves $i^\star$.
    \textbf{(d)} Contrasting ``red dress'' with ``dress'' removes the object-dominated feature \#14976.
    \textbf{(e)} The decoder direction of $i^\star$ is added within the target region, and the resulting image is compared with the unsteered image $I^{(0)}$.
    \textbf{(f)} With \#5824, raising $\rho$ from 0.15 to 0.30 does not increase $\Delta_{\mathrm{target}}$ but nearly doubles LPIPS-out.
    }
    \label{fig:framework}
\end{figure}

To answer this question, we introduce \textbf{D-Scope (Diffusion Scope)}, a method that uses each feature's highly activating image patches as \emph{visual evidence} to select SAE decoder directions for text-guided generation steering (Fig.~\ref{fig:framework}). D-Scope embeds these patches with SigLIP~2~\citep{tschannen2025siglip} and aggregates them through activation-weighted pooling into a \emph{visual centroid} representing each SAE feature. Given a target description, D-Scope retrieves relevant features by comparing the text embedding against these visual centroids. Because a description can contain both an object and the attribute to be changed, retrieval can be performed either directly using the target description or contrastively relative to a reference describing the generic object or its source appearance. The retrieved feature identifies a single SAE decoder direction, which is applied within a spatial mask while the source prompt, initial noise, and sampling settings remain fixed. D-Scope thus provides a direct path from a target description to an SAE intervention direction, using the activating patches themselves as visual evidence for feature retrieval.

\textbf{Key Observations.} 
Retrieval relies on dictionary features with sufficient visual evidence for semantic matching.
Our analysis of 150 SAEs shows that high reconstruction fidelity can coexist with low dictionary utilization and limited visual-evidence coverage, so reconstruction alone does not establish how broadly such evidence is available.
On our fine-grained steering benchmark, regional target gain measures the increase in SigLIP~2 alignment relative to the unsteered source image.
Under per-case best-of-sweep strength selection, contrastive retrieval yields higher mean gains than direct retrieval across the evaluated configurations, without consistently improving outside-region preservation.
Compared with the Oracle Dense, which uses paired source and target activations, a single retrieved decoder direction achieves up to 86\% of its regional target-alignment gain without using target-generation activations for feature selection.
Under task-specific selection protocols, individual features from the same pretrained dictionaries also support style suppression and semantic readout, extending their use beyond query-driven steering without updating the generator or SAE.
Our contributions are threefold:
\begin{itemize}[nosep,leftmargin=28pt]
    \item \textbf{A visual-evidence method for retrieving SAE decoder directions.}
    D-Scope matches target descriptions to activation-weighted visual centroids through direct and contrastive retrieval, identifying individual intervention directions without per-feature text annotations or target-generation activations.

    \item \textbf{A systematic analysis of SAE dictionaries for retrieval and intervention.}
    We compare 150 SAEs to characterize how training choices shape reconstruction fidelity, dictionary utilization, and visual evidence, showing that accurate reconstruction alone does not ensure sufficient evidence for feature retrieval.

    \item \textbf{A controlled benchmark for evaluating retrieved directions.}
    We introduce 1,000 fine-grained steering cases that measure regional target alignment and outside-region preservation under a fixed single-feature budget.
    Under per-case best-of-sweep strength selection, a single retrieved direction achieves up to 86\% of the Oracle Dense regional target-alignment gain.
\end{itemize}

\section{Related Work}
\label{sec:related_work}

\textbf{Sparse Autoencoders and Intervention Evaluation.}
Prior work has established interpretable sparse features and
generation interventions in diffusion
models~\citep{surkov2024one,shabalin2025interpreting,Tnaz2025EmergenceAE},
including temporal-aware dictionaries for DiTs~\citep{huang2025tidetemporalawaresparse}.
D-Scope connects dictionary training and analysis with semantic feature retrieval and intervention evaluation.
Our training study compares SAE architectures, input settings, and layers, assessing dictionary utilization and visual-evidence availability alongside reconstruction.
Related evaluations in language models demonstrate why these different aspects of feature utility should be examined separately.
SAEBench shows that improvements in reconstruction-sparsity proxies do not reliably translate into downstream gains~\citep{karvonen2025saebench}.
AxBench evaluates concept detection and model steering~\citep{wu2025axbench}, while \citet{arad2025saes} distinguish
features that activate in response to a concept from those whose interventions produce the corresponding generation behavior.
Since activation-based relevance does not by itself establish
steering effectiveness, D-Scope evaluates retrieved directions under a single-feature protocol.
Following editing benchmarks that separate edit fidelity from background preservation~\citep{ju2024pnp}, our evaluation measures regional target alignment and outside-region preservation separately, enabling controlled comparisons of dictionary and retrieval choices.

\textbf{Semantic Grounding and Feature Retrieval.}
Related approaches obtain semantic interpretations through automated feature descriptions~\citep{yu2025coe,ferrando2026language},
CLIP-based concept matching~\citep{oikarinen2022clip,rao2024discover}, or joint learning of visual and textual sparse codes~\citep{shen2025vl,gu2026lucid}.
Discover-then-Name matches dictionary directions learned in CLIP space directly to text embeddings~\citep{rao2024discover}.
Such direct matching requires compatible representations, whereas DiT decoder directions are not directly aligned with text embeddings.
SemanticLens represents model components through pooled embeddings of highly activating image regions, supporting natural-language retrieval~\citep{dreyer2025mechanistic}.
D-Scope constructs a visual index after SAE dictionary training in DiT, allowing text queries to select features from their activating patches without per-feature text annotations or joint vision-language SAE training.
Activation-weighted visual centroids support direct and contrastive matching, and each retrieved feature index identifies the corresponding decoder direction for generation intervention.
RIEBench uses a different selection mechanism: it selects transport features from paired source and target trajectories, requiring a target image to be generated and its SAE activations collected before feature selection~\citep{surkov2024one}.
D-Scope instead considers a setting in which the desired change is specified by a text query, with no target image or target-generation activations available for feature selection.
Appendix~\ref{app:additional-related-work} provides further discussion.

\section{D-Scope}
\label{sec:framework}

D-Scope connects target text queries to SAE decoder directions through a visual feature index (Fig.~\ref{fig:framework}c).
Each indexed feature $i$ is associated with its activating image patches, a visual centroid $\mathbf{v}_i$ representing these patches, and an SAE decoder direction $\mathbf{d}_i$.
Text queries are matched to the centroids in a shared image-text embedding space, while the correspondence $\mathbf{v}_i\leftrightarrow i\leftrightarrow\mathbf{d}_i$ identifies the associated SAE decoder direction.
This correspondence makes retrieval the link between the visual evidence used to select a feature and the direction used to intervene on generation.
D-Scope builds a visual index from a trained SAE by collecting highly activating patches for each feature and aggregating them into a visual centroid (Sec.~\ref{sec:sae_features}--\ref{sec:feature_retrieval}, Fig.~\ref{fig:framework}a,b,c).
A target query then retrieves a single feature from this index using direct or contrastive matching (Sec.~\ref{sec:feature_retrieval}, Fig.~\ref{fig:framework}c,d), and the corresponding decoder direction is adapted to the DiT residual space within the target region as a spatially masked intervention (Sec.~\ref{sec:single_feature_steering}, Fig.~\ref{fig:framework}e,f).
The dictionary and visual index are built offline and reused across queries, whereas retrieval and intervention are performed online for each query.
 
\textbf{Steering task.}
A steering case consists of a source prompt $c$, an initial noise $\boldsymbol{\epsilon}$, a target query $q$ describing the desired appearance, and a binary mask $M$ marking the target region.
The frozen DiT generates the source image $I^{(0)}$ from $c$ and $\boldsymbol{\epsilon}$; in Fig.~\ref{fig:framework}f, for example, $q$ is ``red dress'' and $M$ covers the dress in $I^{(0)}$.
Steering regenerates the image with $c$, $\boldsymbol{\epsilon}$, and the sampling settings unchanged, while applying the intervention direction derived from a single retrieved SAE decoder direction within $M$ at every denoising step.
The goal is to move the target region toward $q$ while preserving content outside $M$; 
with the intervention protocol fixed, steering reduces to selecting which feature to use and how strongly to apply it.

\subsection{SAE Features and Visual Evidence}
\label{sec:sae_features}

\textbf{Sparse features and decoder directions.}
An SAE decomposes residual-stream activations of visual tokens into sparse combinations of learned features (Fig.~\ref{fig:framework}a).
Let $\mathbf h\in\mathbb R^D$ denote the residual-stream activation
of a visual token after the chosen transformer block, where $D$ is
the number of activation channels.
A fixed input transformation $\mathcal T$ produces the SAE input
$\mathbf x=\mathcal T(\mathbf h)$.
The SAE encodes $\mathbf x$ into sparse coefficients
$\mathbf z\in\mathbb R^m$, where $m$ is the number of dictionary
features, and reconstructs the input as
\begin{equation*}
    \mathbf z
    =
    \phi_{\mathrm{SAE}}\!\left(
    \mathbf W_{\mathrm{enc}}
    (\mathbf x-\mathbf b_{\mathrm{dec}})
    +\mathbf b_{\mathrm{enc}}
    \right),
    \qquad
    \hat{\mathbf x}
    =
    \mathbf W_{\mathrm{dec}}\mathbf z+\mathbf b_{\mathrm{dec}}.
\end{equation*}
Here, $\phi_{\mathrm{SAE}}$ denotes the family-specific sparse activation rule.
The decoder matrix $\mathbf W_{\mathrm{dec}}=[\mathbf d_1,\ldots,\mathbf d_m]$
defines the SAE dictionary.
Each feature $i$ has an activation $z_i$ and a decoder direction
$\mathbf d_i$ expressed in the SAE input coordinates.
Feature activations identify the image patches that provide visual
evidence for retrieval.
After a feature is selected, its decoder direction is adapted for
intervention in the DiT residual stream.

\textbf{Visual evidence.}
We collect visual evidence from a fixed set of images generated offline
by the frozen DiT (Fig.~\ref{fig:framework}b).
For each image, the SAE encodes the visual-token activations at the
chosen block into a sparse feature vector $\mathbf z$ for each token.
For each feature $i$, we compare its activation $z_i$ across these
tokens and select the token with the largest activation.
If this activation is positive, we extract a candidate patch centered
at the token's corresponding image location, yielding at most one
candidate per feature per image.
A feature is eligible for indexing if it has positive candidates from
at least $J$ distinct images.
For each eligible feature, we retain the $J$ candidates with the largest
activations.
Let $P_{ij}$ denote retained patch $j$ and $a_{ij}>0$ the value of
$z_i$ at its selected token.
The patch-activation pairs $\{(P_{ij},a_{ij})\}_{j=1}^{J}$ constitute
the visual evidence for feature $i$.
We use $J=64$ throughout.

\subsection{Visual Index and Feature Retrieval}
\label{sec:feature_retrieval}

\textbf{Visual feature index.}
D-Scope constructs a visual index once from the visual evidence collected
in Sec.~\ref{sec:sae_features} and reuses it across queries
(Fig.~\ref{fig:framework}c).
Let $\mathcal I$ denote the features satisfying the evidence requirement
defined in Sec.~\ref{sec:sae_features}.

Using the frozen SigLIP~2 image encoder $E_V$~\citep{tschannen2025siglip}
and unit normalization
$\nu(\mathbf a)=\mathbf a/\|\mathbf a\|_2$, we summarize each indexed
feature by an activation-weighted visual centroid
\begin{equation*}
    \mathbf v_i
    =
    \nu\!\left(
    \sum_{j=1}^{J}
    a_{ij}\,
    \nu\!\left(E_V(P_{ij})\right)
    \right).
\end{equation*}

The visual centroid $\mathbf v_i$ and decoder direction $\mathbf d_i$
serve different roles: $\mathbf v_i$ represents the feature in the
shared image--text embedding space for semantic retrieval, whereas
$\mathbf d_i$ is the corresponding SAE direction used for intervention.
They are linked by the feature identity,
\[
    \mathbf v_i \leftrightarrow i \leftrightarrow \mathbf d_i.
\]
Thus, the feature identity provides a direct link between its visual representation and the corresponding SAE decoder direction, without requiring a learned mapping between the two latent spaces.

\textbf{Direct and contrastive retrieval.}
Retrieval operates entirely in the shared image-text embedding space.
A target query such as ``\textit{red dress}'' specifies both the object
(``\textit{dress}'') and the requested attribute (``\textit{red}'').
Direct retrieval matches the full query and can therefore favor features
associated with either component.
Contrastive retrieval represents the target query $q$ relative to a
reference query $q_{\mathrm{base}}$ that describes the generic object
or its source appearance (Fig.~\ref{fig:framework}d).
For a target query of ``\textit{red dress}'', the reference can be
``\textit{dress}'' when the source color is unspecified.
The comparison is intended to emphasize the requested attribute
or attribute change.

Let $\mathbf t(q)=\nu(E_T(q))$ denote the normalized embedding from the
frozen SigLIP~2 text encoder $E_T$. The two query representations are
\begin{equation*}
    \mathbf r_{\mathrm{dir}}
    =
    \mathbf t(q),
    \qquad
    \mathbf r_{\mathrm{con}}
    =
    \nu\!\left(
    \mathbf t(q)-\mathbf t(q_{\mathrm{base}})
    \right).
\end{equation*}
For either representation $\mathbf r$, cosine matching retrieves
\begin{equation*}
    i^\star
    =
    \arg\max_{i\in\mathcal I}
    \mathbf r^\top\mathbf v_i.
\end{equation*}
The selected identity $i^\star$ then links semantic retrieval back to the
SAE dictionary, identifying both its supporting visual evidence and the
decoder direction $\mathbf d_{i^\star}$ used in the subsequent steering
stage.

\subsection{Single-Feature Steering}
\label{sec:single_feature_steering}

\textbf{Steering direction.}
The retrieved decoder direction $\mathbf d_{i^\star}$ is defined in the SAE input coordinates, while steering modifies the original DiT residual-stream activation $\mathbf h$.
A fixed operator $\mathcal A_{\mathcal T}$ adjusts the decoder
direction according to the SAE input setting.
For the None setting, $\mathcal T(\mathbf h)=\mathbf h$ and
$\mathcal A_{\mathcal T}(\mathbf d_i)=\mathbf d_i\in\mathbb R^D$.
For LayerNorm, $\mathcal T$ centers and rescales each token's
activation.
We adjust the corresponding decoder direction by subtracting its
channel mean,
$\mathcal A_{\mathcal T}(\mathbf d_i)=\mathbf d_i-\bar d_i\mathbf 1$,
where $\bar d_i$ is the mean of the atom's $D$ channel values and
$\mathbf 1$ is the all-ones vector.
This adjustment removes the channel-mean component without
inverting the input normalization.
The unit steering direction is
\begin{equation*}
    \hat{\mathbf u}_{i^\star}
    =
    \nu\!\left(
        \mathcal A_{\mathcal T}(\mathbf d_{i^\star})
    \right).
\end{equation*}
Adjustment rules for all input settings are provided in
Appendix~\ref{app:steering-protocol}.

\textbf{Masked intervention.}
The target object in the source image $I^{(0)}$ is segmented with
SAM~3~\citep{carion2026sam} to obtain a binary mask $M$ (Fig.~\ref{fig:framework}e).
This mask is mapped to the visual-token grid, with $M_p=1$ for
tokens selected for intervention and $M_p=0$ otherwise.
At denoising step $t$, let $\mathbf h_{t,p}$ denote the
residual-stream activation of token $p$ after the
transformer block used to train the SAE.
The masked update is
\begin{equation*}
    \mathbf h'_{t,p}
    =
    \mathbf h_{t,p}
    +
    M_p\,\rho\,R(t)\,\hat{\mathbf u}_{i^\star},
\end{equation*}
where $\rho$ controls the relative intervention strength.
The reference scale $R(t)$ is the root mean square of raw residual-vector
$\ell_2$ norms over all spatial tokens in a fixed set of unmodified
source generations, computed at the same block and denoising step.
Since $\hat{\mathbf u}_{i^\star}$ has unit norm, $\rho R(t)$ determines
the update magnitude for each selected token.
The scales are precomputed and shared across feature choices and
intervention strengths.

The same direction is applied to all masked tokens at every denoising
step, with $\rho$ held constant within each generation.
The source prompt, initial noise, mask, and sampling settings remain
unchanged across generations with different intervention strengths
(Fig.~\ref{fig:framework}f).
Detailed calibration and intervention settings are provided in
Appendix~\ref{app:steering-protocol}.

\section{Experiments}

Our experiments evaluate whether D-Scope retrieves single SAE decoder directions that induce the requested regional changes while preserving surrounding content.
We first analyze 150 SAEs trained on two one-step DiT generators to characterize the visual evidence available for feature retrieval.
Steering evaluations span three generation settings on our fine-grained benchmark, comparing direct and contrastive retrieval and examining how dictionary quality relates to regional target alignment and outside-region preservation.
Additional experiments on style suppression and semantic readout assess the broader utility of the same dictionaries under task-specific feature selection in Appendix~\ref{app:broader-uses}.
We summarize practical takeaways from these analyses in Table~\ref{tab:takeaways}.

\subsection{Experimental Setup}
\textbf{Models and dictionaries.}
We train 150 SAEs on hidden activations from two frozen one-step DiT generators, SANA-Sprint and Nitro-1-PixArt, spanning five layers, three SAE families (\textbf{L1}, \textbf{TopK}, and \textbf{JumpReLU}), and five preprocessing or initialization settings (none, mean centering, geometric-median centering, LayerNorm, and RMS normalization).
Steering is evaluated in three model environments: \textbf{SANA-Sprint}~\citep{chen2025sana}, \textbf{Nitro-1-PixArt}~\citep{chen2024pixart}, and the \textbf{SANA teacher}~\citep{xie2025sana}.
For the teacher, we reuse the dictionaries trained on SANA-Sprint rather than training separate teacher SAEs.
Unless otherwise specified, steering uses middle-layer dictionaries;
full training configurations are provided in Appendix~\ref{app:sae-training-configurations} and Table~\ref{tab:sae-training-configurations}.

\textbf{Fine-grained steering benchmark.}
Our benchmark contains 1,000 cases comprising 100 target concepts across five attribute categories: color, material, texture or pattern, local appearance, and local state or geometry. Each concept is paired with ten source contexts. \textit{Under-specified} contexts leave the target attribute unspecified in the source prompt, while \textit{explicit-conflict} contexts specify a competing attribute. SAM~3~\citep{carion2026sam} extracts the target-region mask from the source image, and the mask remains fixed across output comparisons. 
Detailed intervention and evaluation protocols, including benchmark construction, are provided in Appendix~\ref{app:steering-protocol}, with the concept taxonomy in Table~\ref{tab:fgs-taxonomy}.

\textbf{Reference baselines.}
For evaluation only, each case additionally specifies a target prompt
$c'$ describing the edited scene.
\textbf{Oracle Prompt} regenerates the image from $c'$ using the same initial
noise and sampling settings.
\textbf{Oracle Dense} uses the paired source and target generations from $c$ and
$c'$ to construct a dense intervention direction from their activation
difference.
Thus, Oracle Dense has access to target-generation activations, whereas
D-Scope selects its direction using only the target query and the
precomputed visual index.
Detailed constructions are provided in
Appendix~\ref{app:steering-protocol}. 

\textbf{Evaluation.}
We measure target control with \emph{Region SigLIP $\Delta$}, the increase in target-query alignment within the target region relative to the unsteered source, and \emph{outside-region preservation with LPIPS-out}, the perceptual change outside the target region.
For each case, we select the steering strength that maximizes Region SigLIP $\Delta$ and report LPIPS-out on the same output.
Results are averaged over three generation seeds; metric and reporting details are provided in Appendix~\ref{app:steering-protocol}.

\subsection{SAE Dictionary Quality for Visual-Evidence Retrieval}
\label{sec:sae_dictionaries}

\begin{figure}[!ht]
    \centering
    \includegraphics[width=\linewidth]{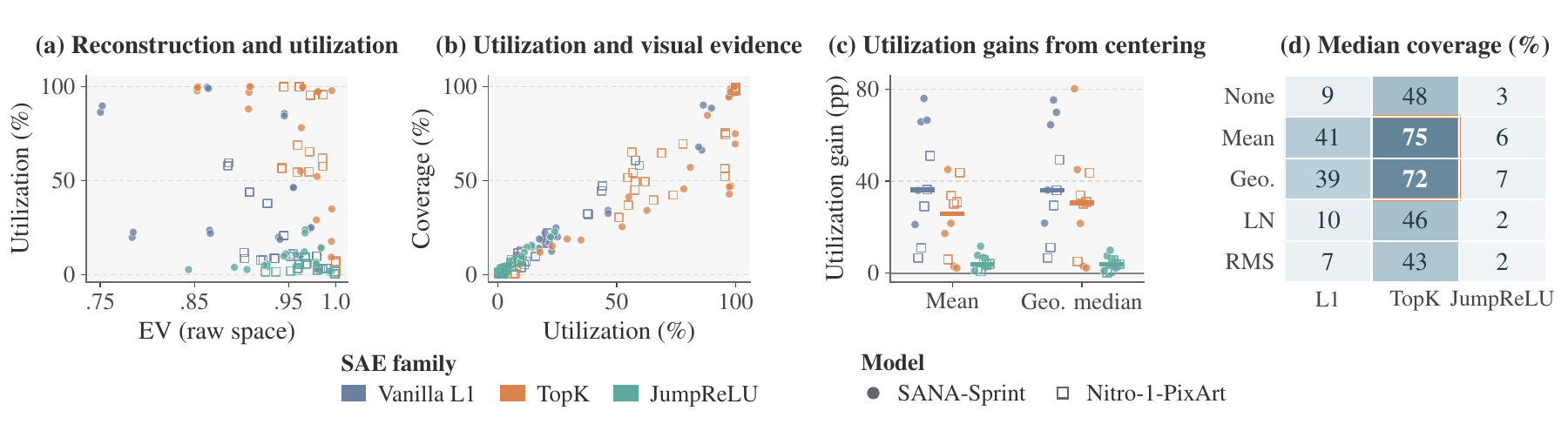}
    \caption{
    \textbf{Reconstruction fidelity alone does not characterize the feature pool available for visual-evidence retrieval.}
    \textbf{(a)} Raw-space explained variance (EV) versus dictionary utilization.
    \textbf{(b)} Dictionary utilization versus visual-evidence coverage.
    \textbf{(c)} Percentage-point change in utilization from mean or geometric-median centering relative to the matched uncentered configuration; horizontal bars denote medians.
    \textbf{(d)} Median visual-evidence coverage across models and layers for each SAE family and input setting; outlined cells indicate the two highest median coverage values.
    }
    \label{fig:training-evaluation}
\end{figure}

We first examine the SAE dictionaries that support feature retrieval. Specifically, we ask whether reconstruction fidelity alone is sufficient to characterize the extent to which dictionary features are utilized and supported by visual evidence.
Figure~\ref{fig:training-evaluation} summarizes these relationships across
150 SAEs; metric definitions, additional analyses, and complete results are
provided in Appendices~\ref{app:dictionary-evaluation}--\ref{app:sae-complete-results}.

\textbf{Reconstruction quality does not imply retrieval readiness.}
\textit{Dictionary utilization} measures the fraction of features active within
the 256,000 training tokens preceding the selected checkpoint.
\textit{Visual-evidence coverage} measures the fraction of sampled features
with positive activations in at least $J=64$ distinct images.
Reconstruction fidelity and dictionary utilization can differ substantially
(Fig.~\ref{fig:training-evaluation}a), while utilization is more closely
related to the availability of visual evidence
(Fig.~\ref{fig:training-evaluation}b).
For example, an uncentered TopK SAE at Nitro layer 2 achieves an explained
variance (EV) of 0.9997, while only 1.5\% of its features are utilized and
0.5\% of the sampled features provide sufficient visual evidence
(Appendix~\ref{app:sae-complete-results},
Table~\ref{tab:app-sae-nitro-results}).
Thus, reconstruction fidelity alone does not indicate whether a dictionary
exposes a broad set of features that can be represented from their
activating examples.

\textbf{Training choices substantially change visual-evidence availability.}
Centering~\citep{simon2026relationshipactivationoutliersfeature} 
generally increases dictionary utilization across models, layers,
and SAE families (Fig.~\ref{fig:training-evaluation}c), while centered TopK
achieves the highest median evidence coverage in our sweep
(Fig.~\ref{fig:training-evaluation}d).
These results show that SAE configurations with similar reconstruction quality
can expose substantially different feature pools for visual-evidence retrieval.
Features that remain inactive cannot accumulate the repeated positive activations required to enter the visual index, so \textit{low dictionary utilization can indicate a restricted candidate pool for retrieval}.

\subsection{Single-Feature Steering Performance}
\label{sec:steering-benchmark}

\begin{figure*}[t]
  \centering
  \input{figs_compressed/fig4_benchmarking/fig4_steering}
  \caption{
    \textbf{Single-feature steering with D-Scope across generators and SAE dictionaries.}
    \textbf{(a)} ``Wooden table'' interventions using Oracle Prompt, Oracle Dense,
    direct retrieval, and contrastive retrieval; dashed outlines mark the target
    region, and top activating patches are shown for the retrieved SAE features.
    \textbf{(b)} Regional target gain and outside-region change for middle-layer
    dictionaries, normalized by Oracle Dense.
    Faded and solid markers denote direct and contrastive retrieval, respectively,
    with matched dictionaries connected by lines.
    \textbf{(c)} Change in target gain from mean or geometric-median centering
    relative to the matched uncentered configuration; horizontal bars denote medians.
    }
  \label{fig:steering}
\end{figure*}
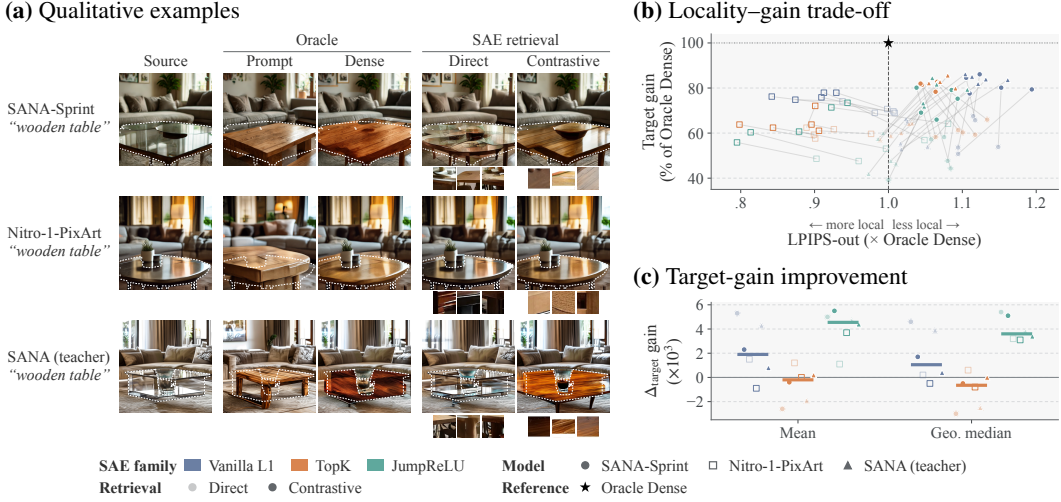

We next evaluate the central question of D-Scope: \textbf{whether a desired visual change can be realized using a single retrieved SAE decoder direction}.
We compare the retrieved direction with Oracle Dense, which uses a dense direction derived from paired source and target activations under the same spatial intervention protocol.
Figure~\ref{fig:steering}a,b summarizes the main results; complete quantitative results are reported in Table~\ref{tab:steering-main}, with additional results and analyses in Appendix~\ref{app:steering-benchmark}.

\textbf{Retrieved features produce the requested visual change.}
Across all evaluated dictionaries and model environments, intervening with the retrieved decoder direction yields a positive Region SigLIP $\Delta$, ranging from $13.8$ to $33.4$ ($\times 10^{-3}$), with a mean of $23.1$.
Figure~\ref{fig:steering}a illustrates this behavior for ``wooden table'':
although D-Scope uses only one SAE feature, the retrieved directions induce
attribute changes similar to those produced by the dense reference.
The corresponding visual-evidence patches also make the selected feature
directly inspectable.

\textbf{A single retrieved feature captures most of the dense-reference gain.}
With contrastive retrieval, the best dictionaries recover 86\%, 78\%, and 86\%
of the Oracle Dense target-alignment gain on SANA-Sprint, Nitro-1-PixArt, and
the SANA teacher, respectively (Fig.~\ref{fig:steering}b).
Averaged across all 15 dictionaries, the corresponding ratios remain 79\%,
69\%, and 82\%.
Thus, despite restricting each edit to a single query-retrieved decoder
direction, D-Scope captures a substantial fraction of the regional target-alignment gain achieved by Oracle Dense.

\textbf{Target control and locality are distinct.}
Figure~\ref{fig:steering}b also shows that larger target gains do not necessarily coincide with lower outside-region change.
We next examine how retrieval strategy and dictionary choice shape this trade-off.

\begin{table*}[t]
\centering
\caption{
\textbf{Complete fine-grained steering benchmark.}
We evaluate all 15 middle-layer SAE configurations with direct and
contrastive retrieval in three model environments; the SANA teacher reuses
the student-trained SANA dictionaries.
Region SigLIP improvement ($\Delta\uparrow$) and LPIPS-out ($\downarrow$)
are reported as mean $\pm$ sample SD across three generation-seed means,
with both metrics and SDs multiplied by $10^3$.
For each swept method and case-seed pair, steering strength maximizes
Region SigLIP $\Delta$; LPIPS-out is measured on the same output.
Oracle Prompt uses the target prompt, while Oracle Dense uses the
source-target activation difference.
\textbf{Bold} denotes the better retrieval strategy within each
dictionary, model, and metric based on unrounded means.
}
\label{tab:steering-main}

\begingroup
\fontsize{8.5}{10}\selectfont
\setlength{\tabcolsep}{2.8pt}
\renewcommand{\arraystretch}{1.06}

\newcommand{\SteeringResult}[2]{%
    \ensuremath{#1{\scriptstyle\,\pm\,#2}}%
}

\resizebox{\textwidth}{!}{%
\begin{tabular}{@{}llrrrrrrrrrrrr@{}}
\toprule
& &
\multicolumn{4}{c}{SANA Student} &
\multicolumn{4}{c}{Nitro Student} &
\multicolumn{4}{c}{SANA Teacher}
\\
\cmidrule(lr){3-6}
\cmidrule(lr){7-10}
\cmidrule(lr){11-14}

& &
\multicolumn{2}{c}{Direct} &
\multicolumn{2}{c}{Contrastive} &
\multicolumn{2}{c}{Direct} &
\multicolumn{2}{c}{Contrastive} &
\multicolumn{2}{c}{Direct} &
\multicolumn{2}{c}{Contrastive}
\\
\cmidrule(lr){3-4}
\cmidrule(lr){5-6}
\cmidrule(lr){7-8}
\cmidrule(lr){9-10}
\cmidrule(lr){11-12}
\cmidrule(lr){13-14}

Method & Input &
\multicolumn{1}{c}{$\Delta\uparrow$} &
\multicolumn{1}{c}{LPIPS$\downarrow$} &
\multicolumn{1}{c}{$\Delta\uparrow$} &
\multicolumn{1}{c}{LPIPS$\downarrow$} &
\multicolumn{1}{c}{$\Delta\uparrow$} &
\multicolumn{1}{c}{LPIPS$\downarrow$} &
\multicolumn{1}{c}{$\Delta\uparrow$} &
\multicolumn{1}{c}{LPIPS$\downarrow$} &
\multicolumn{1}{c}{$\Delta\uparrow$} &
\multicolumn{1}{c}{LPIPS$\downarrow$} &
\multicolumn{1}{c}{$\Delta\uparrow$} &
\multicolumn{1}{c}{LPIPS$\downarrow$}
\\
\midrule

No intervention & -- &
\SteeringResult{0.0}{0.0} &
\SteeringResult{0.0}{0.0} &
\SteeringResult{0.0}{0.0} &
\SteeringResult{0.0}{0.0} &
\SteeringResult{0.0}{0.0} &
\SteeringResult{0.0}{0.0} &
\SteeringResult{0.0}{0.0} &
\SteeringResult{0.0}{0.0} &
\SteeringResult{0.0}{0.0} &
\SteeringResult{0.0}{0.0} &
\SteeringResult{0.0}{0.0} &
\SteeringResult{0.0}{0.0}
\\

Oracle Prompt & -- &
\SteeringResult{37.9}{1.6} &
\SteeringResult{392.8}{8.6} &
\SteeringResult{37.9}{1.6} &
\SteeringResult{392.8}{8.6} &
\SteeringResult{36.4}{0.8} &
\SteeringResult{289.6}{6.0} &
\SteeringResult{36.4}{0.8} &
\SteeringResult{289.6}{6.0} &
\SteeringResult{39.3}{0.5} &
\SteeringResult{421.1}{10.7} &
\SteeringResult{39.3}{0.5} &
\SteeringResult{421.1}{10.7}
\\

Oracle Dense & -- &
\SteeringResult{38.8}{0.4} &
\SteeringResult{147.4}{2.5} &
\SteeringResult{38.8}{0.4} &
\SteeringResult{147.4}{2.5} &
\SteeringResult{29.0}{0.4} &
\SteeringResult{59.5}{1.1} &
\SteeringResult{29.0}{0.4} &
\SteeringResult{59.5}{1.1} &
\SteeringResult{34.8}{0.2} &
\SteeringResult{219.2}{7.6} &
\SteeringResult{34.8}{0.2} &
\SteeringResult{219.2}{7.6}
\\

\midrule

L1 & None &
\SteeringResult{20.9}{0.5} &
\SteeringResult{\mathbf{169.3}}{2.2} &
\SteeringResult{\mathbf{31.1}}{0.5} &
\SteeringResult{169.9}{4.8} &
\SteeringResult{20.0}{0.1} &
\SteeringResult{58.4}{4.4} &
\SteeringResult{\mathbf{22.6}}{0.2} &
\SteeringResult{\mathbf{54.3}}{2.7} &
\SteeringResult{18.8}{0.5} &
\SteeringResult{\mathbf{230.8}}{12.0} &
\SteeringResult{\mathbf{29.2}}{0.4} &
\SteeringResult{248.1}{9.1}
\\

& Mean center &
\SteeringResult{26.2}{0.7} &
\SteeringResult{\mathbf{164.3}}{3.2} &
\SteeringResult{\mathbf{33.4}}{0.6} &
\SteeringResult{165.6}{6.1} &
\SteeringResult{21.5}{0.5} &
\SteeringResult{55.8}{3.4} &
\SteeringResult{\mathbf{21.7}}{0.1} &
\SteeringResult{\mathbf{52.0}}{2.0} &
\SteeringResult{23.1}{0.5} &
\SteeringResult{\mathbf{224.4}}{12.7} &
\SteeringResult{\mathbf{30.0}}{0.2} &
\SteeringResult{242.0}{6.1}
\\

& Geo. center &
\SteeringResult{25.5}{0.4} &
\SteeringResult{164.7}{4.0} &
\SteeringResult{\mathbf{32.8}}{0.5} &
\SteeringResult{\mathbf{162.9}}{4.5} &
\SteeringResult{20.2}{0.1} &
\SteeringResult{59.9}{4.0} &
\SteeringResult{\mathbf{22.1}}{0.4} &
\SteeringResult{\mathbf{50.1}}{3.2} &
\SteeringResult{22.7}{0.6} &
\SteeringResult{\mathbf{224.8}}{11.3} &
\SteeringResult{\mathbf{29.6}}{0.5} &
\SteeringResult{244.0}{11.6}
\\

& LayerNorm &
\SteeringResult{19.7}{0.6} &
\SteeringResult{\mathbf{161.3}}{6.3} &
\SteeringResult{\mathbf{30.8}}{0.4} &
\SteeringResult{176.0}{5.7} &
\SteeringResult{20.0}{0.1} &
\SteeringResult{60.0}{2.2} &
\SteeringResult{\mathbf{22.0}}{0.1} &
\SteeringResult{\mathbf{54.1}}{3.8} &
\SteeringResult{18.4}{0.3} &
\SteeringResult{\mathbf{223.0}}{9.3} &
\SteeringResult{\mathbf{29.1}}{0.1} &
\SteeringResult{254.6}{9.7}
\\

& RMS &
\SteeringResult{20.8}{0.5} &
\SteeringResult{\mathbf{161.3}}{5.2} &
\SteeringResult{\mathbf{30.1}}{0.3} &
\SteeringResult{163.6}{3.8} &
\SteeringResult{20.5}{0.1} &
\SteeringResult{59.4}{3.2} &
\SteeringResult{\mathbf{22.6}}{0.4} &
\SteeringResult{\mathbf{55.3}}{2.4} &
\SteeringResult{19.3}{0.3} &
\SteeringResult{\mathbf{222.8}}{8.3} &
\SteeringResult{\mathbf{28.5}}{0.1} &
\SteeringResult{245.8}{10.9}
\\

\midrule

TopK & None &
\SteeringResult{25.6}{0.6} &
\SteeringResult{167.1}{4.0} &
\SteeringResult{\mathbf{32.1}}{0.8} &
\SteeringResult{\mathbf{157.1}}{3.3} &
\SteeringResult{16.7}{0.1} &
\SteeringResult{53.6}{3.2} &
\SteeringResult{\mathbf{18.5}}{0.2} &
\SteeringResult{\mathbf{47.5}}{2.6} &
\SteeringResult{22.4}{0.5} &
\SteeringResult{\mathbf{226.2}}{8.8} &
\SteeringResult{\mathbf{28.6}}{0.4} &
\SteeringResult{233.9}{8.7}
\\

& Mean center &
\SteeringResult{23.0}{0.2} &
\SteeringResult{164.7}{4.1} &
\SteeringResult{\mathbf{31.7}}{0.1} &
\SteeringResult{\mathbf{157.3}}{5.0} &
\SteeringResult{17.9}{0.4} &
\SteeringResult{55.1}{2.5} &
\SteeringResult{\mathbf{18.5}}{0.5} &
\SteeringResult{\mathbf{53.3}}{1.9} &
\SteeringResult{20.5}{0.4} &
\SteeringResult{\mathbf{219.9}}{9.8} &
\SteeringResult{\mathbf{28.8}}{0.6} &
\SteeringResult{234.5}{9.3}
\\

& Geo. center &
\SteeringResult{22.6}{0.3} &
\SteeringResult{156.8}{2.2} &
\SteeringResult{\mathbf{31.6}}{0.7} &
\SteeringResult{\mathbf{154.1}}{2.6} &
\SteeringResult{17.3}{0.3} &
\SteeringResult{58.1}{1.3} &
\SteeringResult{\mathbf{17.7}}{0.3} &
\SteeringResult{\mathbf{53.9}}{2.7} &
\SteeringResult{19.9}{0.4} &
\SteeringResult{\mathbf{216.7}}{9.3} &
\SteeringResult{\mathbf{28.6}}{0.4} &
\SteeringResult{231.2}{10.7}
\\

& LayerNorm &
\SteeringResult{23.4}{0.6} &
\SteeringResult{162.2}{4.8} &
\SteeringResult{\mathbf{30.4}}{0.4} &
\SteeringResult{\mathbf{156.8}}{4.6} &
\SteeringResult{18.1}{0.3} &
\SteeringResult{\mathbf{52.9}}{2.6} &
\SteeringResult{\mathbf{20.9}}{0.2} &
\SteeringResult{53.6}{1.4} &
\SteeringResult{20.9}{0.4} &
\SteeringResult{\mathbf{223.0}}{7.5} &
\SteeringResult{\mathbf{27.8}}{0.5} &
\SteeringResult{239.1}{9.3}
\\

& RMS &
\SteeringResult{23.8}{0.4} &
\SteeringResult{161.3}{3.3} &
\SteeringResult{\mathbf{31.8}}{0.4} &
\SteeringResult{\mathbf{153.8}}{1.9} &
\SteeringResult{17.2}{0.5} &
\SteeringResult{53.5}{2.7} &
\SteeringResult{\mathbf{18.1}}{0.2} &
\SteeringResult{\mathbf{50.2}}{3.0} &
\SteeringResult{22.3}{0.1} &
\SteeringResult{\mathbf{225.3}}{6.1} &
\SteeringResult{\mathbf{29.8}}{0.4} &
\SteeringResult{236.8}{9.4}
\\

\midrule

JumpReLU & None &
\SteeringResult{17.2}{0.6} &
\SteeringResult{159.8}{2.0} &
\SteeringResult{\mathbf{25.6}}{0.6} &
\SteeringResult{\mathbf{157.0}}{5.8} &
\SteeringResult{15.4}{0.3} &
\SteeringResult{59.3}{2.3} &
\SteeringResult{\mathbf{17.6}}{0.2} &
\SteeringResult{\mathbf{52.3}}{4.2} &
\SteeringResult{16.2}{0.4} &
\SteeringResult{\mathbf{221.4}}{10.6} &
\SteeringResult{\mathbf{25.0}}{0.4} &
\SteeringResult{229.8}{9.0}
\\

& Mean center &
\SteeringResult{22.2}{0.5} &
\SteeringResult{155.9}{1.8} &
\SteeringResult{\mathbf{31.1}}{0.4} &
\SteeringResult{\mathbf{153.0}}{4.2} &
\SteeringResult{16.5}{0.2} &
\SteeringResult{62.4}{3.9} &
\SteeringResult{\mathbf{21.3}}{0.2} &
\SteeringResult{\mathbf{56.2}}{3.5} &
\SteeringResult{20.9}{0.5} &
\SteeringResult{\mathbf{217.6}}{10.1} &
\SteeringResult{\mathbf{29.4}}{0.4} &
\SteeringResult{232.2}{9.7}
\\

& Geo. center &
\SteeringResult{22.6}{0.8} &
\SteeringResult{\mathbf{158.5}}{2.7} &
\SteeringResult{\mathbf{30.7}}{0.5} &
\SteeringResult{160.0}{1.9} &
\SteeringResult{18.6}{0.4} &
\SteeringResult{64.3}{1.8} &
\SteeringResult{\mathbf{20.7}}{0.3} &
\SteeringResult{\mathbf{54.9}}{3.0} &
\SteeringResult{20.0}{0.5} &
\SteeringResult{\mathbf{216.1}}{10.8} &
\SteeringResult{\mathbf{28.4}}{0.3} &
\SteeringResult{233.5}{10.5}
\\

& LayerNorm &
\SteeringResult{15.2}{0.1} &
\SteeringResult{\mathbf{147.4}}{1.3} &
\SteeringResult{\mathbf{26.8}}{0.5} &
\SteeringResult{153.9}{2.6} &
\SteeringResult{14.1}{0.4} &
\SteeringResult{53.7}{2.7} &
\SteeringResult{\mathbf{16.2}}{0.3} &
\SteeringResult{\mathbf{47.3}}{3.0} &
\SteeringResult{14.6}{0.1} &
\SteeringResult{\mathbf{213.3}}{11.4} &
\SteeringResult{\mathbf{25.5}}{0.4} &
\SteeringResult{229.5}{10.5}
\\

& RMS &
\SteeringResult{18.5}{0.5} &
\SteeringResult{\mathbf{159.2}}{3.5} &
\SteeringResult{\mathbf{29.2}}{0.7} &
\SteeringResult{161.2}{6.0} &
\SteeringResult{13.8}{0.0} &
\SteeringResult{57.1}{2.8} &
\SteeringResult{\mathbf{17.5}}{0.2} &
\SteeringResult{\mathbf{48.4}}{4.6} &
\SteeringResult{16.6}{0.4} &
\SteeringResult{\mathbf{219.9}}{12.0} &
\SteeringResult{\mathbf{27.6}}{0.5} &
\SteeringResult{242.6}{11.2}
\\

\bottomrule
\end{tabular}%
}

\endgroup
\end{table*}

\subsection{Effects of Retrieval Strategy and Dictionary Choice}
\label{sec:retrieval-analysis}

Figure~\ref{fig:steering}b compares retrieval strategies within matched dictionaries, while Fig.~\ref{fig:steering}c isolates the effect of centering on downstream target control.

\textbf{Contrastive retrieval consistently improves target control.}
Contrastive retrieval yields higher Region SigLIP $\Delta$ than direct
retrieval for all 15 dictionaries in each of the three generation
environments (Fig.~\ref{fig:steering}b).
Averaged over dictionaries, it increases target gain from
$21.8$ to $30.6$ on SANA-Sprint, from $17.9$ to $19.9$ on Nitro-1-PixArt,
and from $19.8$ to $28.4$ on the SANA teacher
(all values $\times10^{-3}$).
The improvement is therefore large in the two SANA environments
($+40\%$ and $+44\%$) but substantially smaller on Nitro ($+11\%$).

\textbf{Dictionary choice matters most when feature availability is limited.}
The weakest steering configurations are also those with the smallest visual
indices.
For L1 and JumpReLU, centering expands the searchable feature pool and improves
target alignment in most matched comparisons (Fig.~\ref{fig:steering}c).
In contrast, TopK already provides a large searchable set before centering, and
further increases in utilization do not consistently improve steering.
Thus, dictionary training is most consequential when it limits the candidate
features available for retrieval; beyond this regime, a larger feature pool
alone does not guarantee a better intervention direction.

\textbf{Improved target control does not guarantee better locality.}
Although contrastive retrieval consistently improves target alignment, its
effect on outside-region preservation varies across generators
(Fig.~\ref{fig:steering}b).
A plausible explanation is that D-Scope selects features according to the
visual evidence that activates them, whereas locality depends on what their
decoder directions write back into the model.
This distinction suggests that semantic relevance and intervention locality
are related but separate properties of a retrieved SAE feature.

\textbf{Low utilization can limit visual-evidence coverage and thereby constrain retrieval-based control.}
Recall from Sec.~\ref{sec:sae_dictionaries} that training choices affect dictionary utilization, and higher utilization is associated with broader visual-evidence coverage (Fig.~\ref{fig:training-evaluation}b). 
For L1 and JumpReLU, centering expands the visual indices by $1.8$--$4.4\times$ and improves mean target alignment in $22$ of $24$ matched comparisons (Table~\ref{tab:steering-main} and Appendix~\ref{app:visual-index} Table~\ref{tab:visual-index-size}).
This pattern is consistent with limited feature availability constraining retrieval-based control.
However, uncentered TopK already indexes more than $12{,}000$ features, and further expansion through centering does not consistently improve target gain.
Selection within the available pool also matters: with the dictionary and visual index fixed, contrastive retrieval consistently achieves higher mean target alignment than direct retrieval (Fig.~\ref{fig:steering}b).
Thus, training shapes the evidence-supported decoder directions available for retrieval, while the retrieval strategy selects among these directions and affects the resulting steering performance.

\section{Conclusion}
\label{sec:conclusion}

We presented D-Scope, a method for retrieving SAE decoder directions by matching target text queries to activation-weighted visual centroids of highly activating image patches. The selected features provide directions for spatially masked interventions without requiring per-feature text annotations or target-generation activations for selection. On our 1,000-case benchmark, a single retrieved direction achieves up to 86\% of the mean regional target-alignment gain of Oracle Dense under per-case best-of-sweep strength selection. Contrastive retrieval consistently improves mean target alignment over direct retrieval across the evaluated configurations, without consistently improving outside-region preservation. Our analysis of 150 SAEs further shows that reconstruction fidelity alone does not characterize the visual evidence available for retrieval. Training choices shape the candidate feature pool, but a larger pool does not guarantee stronger steering, highlighting the importance of selecting effective directions from the available features. D-Scope provides a direct path from a target description to a single SAE intervention, enabling feature selection to be examined through visual evidence and validated through its regional target effects and preservation costs.

\subsection*{AI use statement}

In this work, we used generative AI tools to assist with language polishing. We have reviewed all AI-assisted work, checking the text and figures for consistency with the experimental methods, reported results, and supporting evidence. We take responsibility for the final content of this work, including text, claims, and artifacts produced with the aid of generative AI.

\subsection*{Ethics statement}

D-Scope aims to improve the transparency of diffusion transformers by connecting internal features to visual evidence and generation effects. 
Our safety-oriented motivation parallels that of white-hat security research: we examine potentially sensitive model behavior through controlled evaluation to inform safeguards and support responsible use.
The nudity-related readout experiment demonstrates that a single SAE feature provides an interpretable signal for identifying nudity in generated images, which could assist safety screening and human review. All images used in this experiment are synthesized by the evaluated generator under controlled prompts. No nude photographs of real individuals are collected for this experiment.

Human annotations describe the visible content of generated images under the stated evaluation protocol. These labels do not express moral judgments about nudity, bodies, identities, or social groups, and the study is not intended to stigmatize or discriminate against any group. The presence of nudity is not treated as sufficient evidence that an image is harmful in every context. Sensitive regions in examples displayed in the paper and appendix have been pixelated to limit exposure to explicit content and reduce potential discomfort for readers.

The readout results constitute a controlled demonstration rather than validation of a general-purpose content moderation system. Pretrained models and human annotations may reflect biases, and broader use would require further evaluation of false positives, false negatives, robustness, and performance across populations and contexts. More generally, the ability to steer generation has potential for misuse. Our purpose in studying these capabilities is to support model auditing and inform safeguards. Downstream applications require evaluation appropriate to their intended use, safeguards against misuse, and human oversight.

\subsection*{Reproducibility statement}

Section~\ref{sec:framework} specifies SAE feature representations, visual evidence construction, feature retrieval, and masked single-feature interventions. Appendix~\ref{app:sae-training} documents training configurations, optimization, dictionary evaluation, and complete dictionary results. Appendix~\ref{app:visual-index} describes how feature cards and visual indices are constructed, including the catalog corpus, evidence selection, encoder settings, and the index size of each steering dictionary. Appendix~\ref{app:steering-protocol} details benchmark construction, reference methods, generation settings, evaluation metrics, strength selection, and statistical aggregation. Appendix~\ref{app:style-suppression} describes style-evaluator adaptation, feature selection, and intervention controls. Appendix~\ref{app:nudity-readout} documents human annotation, disjoint feature-discovery and test splits, and the single-feature readout protocol.

\subsubsection*{Acknowledgments}
This project was completed during Xinyue Xu's ongoing Pivotal AI Safety Fellowship. We gratefully acknowledge Pivotal Research Ltd.\ and Dr.\ Peter Hase for their generous financial support.

\bibliography{iclr2027_conference}
\bibliographystyle{iclr2027_conference}

\appendix

\section*{Appendix Guide}

\begingroup
\hypersetup{hidelinks}
\small
\setlength{\parindent}{0pt}
\setlength{\parskip}{0pt}

\newcommand{\appguidegroup}[2]{%
    \par\addvspace{0.9\baselineskip}
    \noindent
    \hyperref[#1]{\textbf{Appendix~\ref*{#1}\quad #2}}%
    \nobreak\dotfill\nobreak
    \hyperref[#1]{\pageref*{#1}}\par
    \nobreak\vspace{0.3\baselineskip}
}

\newcommand{\appguideitem}[2]{%
    \noindent\hspace*{1em}#1%
    \nobreak\hfill\nobreak #2\par
}

\noindent
\hyperref[tab:takeaways]{\textbf{Practical Takeaways}}%
\nobreak\dotfill\nobreak
\hyperref[tab:takeaways]{\pageref*{tab:takeaways}}\par
\nobreak\vspace{0.3\baselineskip}

\appguideitem{Recommendations for building and using DiT SAEs}{Table~\ref{tab:takeaways}}

\appguidegroup{app:additional-related-work}{Additional Related Work}

\appguideitem{Dictionary training, semantic retrieval, and steering evaluation}{}

\appguidegroup{app:sae-training}{SAE Training and Dictionary Evaluation}

\appguideitem{Training configurations}{Table~\ref{tab:sae-training-configurations}}
\appguideitem{Realized sparsity and visual evidence}{Figure~\ref{fig:app-sae-sparsity-evidence}}
\appguideitem{Dictionary quality across layers}{Figure~\ref{fig:app-sae-depth-effects}}
\appguideitem{Complete dictionary results}{Tables~\ref{tab:app-sae-sana-results} and~\ref{tab:app-sae-nitro-results}}

\appguidegroup{app:visual-index}{Feature Cards and Visual Index}

\appguideitem{Catalog construction settings}{Table~\ref{tab:visual-index-setup}}
\appguideitem{Visual index size per dictionary}{Table~\ref{tab:visual-index-size}}

\appguidegroup{app:steering-benchmark}{Fine-Grained Steering Benchmark}

\appguideitem{Benchmark and evaluation protocol}{Section~\ref{app:steering-protocol}}
\appguideitem{Concept taxonomy and example cases}{Tables~\ref{tab:fgs-taxonomy} and~\ref{tab:fgs-rose-case}}
\appguideitem{Concept categories and source contexts}{Figure~\ref{fig:fgs-topics}}

\appguidegroup{app:broader-uses}{Additional Applications of Learned SAE Features}

\appguideitem{Style suppression and semantic readout}{Figure~\ref{fig:broader-uses}}

\appguidegroup{app:style-suppression}{Style Suppression with General-Purpose SAEs}

\appguideitem{Evaluator validation and style selection}{Tables~\ref{tab:app-style-evaluator} and~\ref{tab:app-style-selection}}
\appguideitem{Single-feature suppression protocol}{Table~\ref{tab:app-style-protocol}}
\appguideitem{Complete and per-style results}{Tables~\ref{tab:app-style-all} and~\ref{tab:app-style-per-style}}
\appguideitem{Suppression and retention across dictionaries}{Figure~\ref{fig:app-style-configurations}}
\appguideitem{Successful suppression examples}{Figures~\ref{fig:style-success-1} and~\ref{fig:style-success-2}}
\appguideitem{Incomplete suppression examples}{Figure~\ref{fig:style-failures}}

\appguidegroup{app:nudity-readout}{Nudity-Related Feature Discovery and Readout}

\appguideitem{Dataset split and readout protocol}{Tables~\ref{tab:app-readout-split} and~\ref{tab:app-readout-protocol}}
\appguideitem{Label-guided feature discovery}{Algorithm~\ref{alg:nudity-feature-discovery}}
\appguideitem{Feature evidence and pooling sensitivity}{Figure~\ref{fig:app-nudity-feature-analysis}}
\appguideitem{Matched readout examples}{Figure~\ref{fig:app-nudity-matched-examples}}
\appguideitem{Hard-negative examples}{Figure~\ref{fig:app-nudity-hard-negative-examples}}

\endgroup

\begin{table}[!ht]
\centering
\caption{\textbf{Practical takeaways for building and using DiT SAEs.} Steering findings use middle-layer dictionaries and per-case best-of-sweep strength selection.}
\label{tab:takeaways}
\small
\setlength{\tabcolsep}{5pt}
\renewcommand{\arraystretch}{1.08}
\begin{tabularx}{\linewidth}{@{}>{\raggedright\arraybackslash}p{1.8cm}>{\raggedright\arraybackslash}X>{\raggedright\arraybackslash}X@{}}
\toprule
Decision & Empirical finding & Recommendation \\
\midrule
Dictionary selection &
High reconstruction EV can coexist with low utilization and evidence coverage (Fig.~\ref{fig:training-evaluation}a,b). The three weakest steering configurations in each generation--retrieval setting also have the smallest visual indices (Tables~\ref{tab:steering-main} and~\ref{tab:visual-index-size}). &
Assess utilization, coherence, and coverage alongside reconstruction fidelity. Validate steering performance separately, since a larger visual index does not guarantee stronger control. \\
\addlinespace
Bias initialization &
Centering increases utilization in every matched comparison. For L1 and JumpReLU, it expands visual indices by $1.8$--$4.4\times$ and improves mean target gain in 22 of 24 matched comparisons (Fig.~\ref{fig:training-evaluation}c; Tables~\ref{tab:steering-main} and~\ref{tab:visual-index-size}). &
Start with mean or geometric-median centering, then verify target alignment and preservation for the chosen generator and retrieval strategy. \\
\addlinespace
SAE family &
Centered TopK has the highest median evidence coverage. Mean-centered L1 achieves the highest mean target gain in five of six generator--retrieval combinations (Fig.~\ref{fig:training-evaluation}d; Table~\ref{tab:steering-main}). &
Start with centered TopK for visual inspection and mean-centered L1 for steering, checking preservation in both cases. \\
\addlinespace
Layer &
Layer 26 has lower input-space EV and equal or higher coverage than layer 2 in matched comparisons. Intermediate trends are not uniformly monotonic (Fig.~\ref{fig:app-sae-depth-effects}). &
Compare candidate layers jointly on reconstruction, utilization, coherence, and coverage. \\
\addlinespace
Retrieval &
Contrastive retrieval yields higher mean target gains in all 45 matched comparisons, without consistently improving preservation (Section~\ref{app:steering-analysis}). &
Use contrastive retrieval as a starting point for target alignment, and assess LPIPS-out separately. \\
\addlinespace
Dictionary reuse &
SANA student dictionaries retain positive target gains in the teacher, with higher LPIPS-out than in the student (Table~\ref{tab:steering-main}). &
Re-evaluate target alignment and preservation when transferring dictionaries to another generation setting. \\
\bottomrule
\end{tabularx}
\end{table}

\section{Additional Related Work}
\label{app:additional-related-work}

\textbf{SAE Training and Dictionary Evaluation.} Previous studies investigate how training configurations shape sparse representations in diffusion models. \citet{surkov2024one} compare activation layers, sparsity levels, and dictionary widths, while \citet{shabalin2025interpreting} examine dictionary learning methods, sparsity, expansion, and activation preprocessing in FLUX. The latter also report high explained variance alongside extensive feature inactivity, demonstrating that reconstruction alone does not fully characterize dictionary quality. Other work incorporates denoising time into dictionary learning: TIDE trains temporal-aware SAEs for DiTs and reports hierarchical semantics ranging from 3D structure to fine-grained concepts~\citep{huang2025tidetemporalawaresparse}, while residualized temporal SAEs learn from U-Net activation trajectories rather than isolated timesteps~\citep{yeung2026residualized}. In language models, SAEBench evaluates interpretability, feature disentanglement, and downstream performance, showing that improvements in proxy metrics do not reliably translate into practical gains~\citep{karvonen2025saebench}. Dictionaries trained on the same data can also learn different features across random seeds~\citep{paulo2025sparse}, and Matryoshka SAEs train nested dictionaries to organize features hierarchically~\citep{bussmann2025learning}. Feature absorption further shows that hierarchical concepts need not correspond to cleanly separated SAE features~\citep{chanin2026absorption}. Our training study compares SAE architectures, input settings, and activation layers across DiT backbones. The evaluation distinguishes reconstruction fidelity, realized sparsity, dictionary utilization, visual coherence, and evidence coverage. Reporting coherence together with coverage separates the consistency of features with sufficient visual evidence from the availability of such evidence across the dictionary. These measurements characterize the features available for visual inspection and semantic retrieval, whose intervention effects are evaluated separately.

\textbf{Visual Grounding and Semantic Retrieval.} Semantic interpretations of internal features can be obtained through feature annotation, concept matching, or cross-modal dictionary learning. CLIP-Dissect labels neurons by matching their activations on probe images to a concept set through CLIP~\citep{oikarinen2022clip}. CoE generates language descriptions from activating image patches and quantifies visual concept polysemanticity~\citep{yu2025coe}. \citet{ferrando2026language} elicit feature descriptions by steering a vision-language model's visual encoder and querying its language model. \citet{Tnaz2025EmergenceAE} use object detection and segmentation to associate diffusion SAE features with object concepts. For SAEs trained on CLIP embeddings, Discover-then-Name (DN-CBM) names each feature with the text whose embedding is most similar to its dictionary vector~\citep{rao2024discover}. This direct matching relies on dictionary directions sharing an embedding space with text. Complementary approaches establish semantic correspondences during dictionary learning: VL-SAE maps visual and textual representations into a unified concept set~\citep{shen2025vl}, while LUCID-SAE learns shared and modality-specific sparse codes with cross-modal alignment~\citep{gu2026lucid}. In DiTs, ConceptAttention produces concept saliency maps from the output space of attention layers, providing spatial grounding without a sparse dictionary~\citep{helbling2025conceptattention}. SemanticLens represents model components by pooling foundation-model embeddings of highly activating image regions, enabling semantic comparison and natural-language retrieval~\citep{dreyer2025mechanistic}. D-Scope applies visual-evidence indexing to SAE dictionaries trained on DiT activations. Since their decoder directions are not directly aligned with text embeddings, D-Scope represents each indexed feature through an activation-weighted centroid of its activating patches in the SigLIP~2 embedding space. Text queries can then retrieve features without per-feature text annotations or joint vision-language SAE training. A target description can contain both an object and the attribute to be changed. D-Scope therefore supports direct matching of the full description and contrastive matching relative to a reference describing the generic object or its source appearance. The retained patches provide visual evidence for assessing each selection, while the selected feature identifies a decoder direction that is adapted for intervention in the DiT residual stream.

\textbf{Feature Interventions and Steering Evaluation.} Sparse features have been used for localized control and broader concept manipulation. \citet{surkov2024one} evaluate spatial feature transport with RIEBench, selecting features by comparing SAE activations along paired source and target generation trajectories. This selection requires access to target-generation activations. Residualized temporal SAEs also evaluate transport on RIEBench with segmentation masks, comparing target CLIP similarity and LPIPS to the source image~\citep{yeung2026residualized}. \citet{shabalin2025interpreting} demonstrate activation-based steering in FLUX, while \citet{Tnaz2025EmergenceAE} examine how composition and style interventions vary across denoising stages. Other applications include concept unlearning with SAeUron~\citep{cywinski2025saeuron}, artistic style transfer with LouvreSAE~\citep{panda2025louvresae}, and safe editing and style transfer in DiTs with TIDE~\citep{huang2025tidetemporalawaresparse}. Concept Steerers instead controls generation through sparse codes of text embeddings~\citep{kim2025concept}. Beyond sparse features, Concept Sliders learn low-rank parameter directions whose strength can be continuously modulated~\citep{gandikota2024concept}, and FluxSpace performs training-free semantic editing through representations of rectified flow transformer blocks~\citep{dalva2025fluxspace}. In language models, AxBench finds that the evaluated SAE steering methods underperform prompting and simple baselines under its benchmark settings~\citep{wu2025axbench}. \citet{arad2025saes} distinguish features' responses to input concepts from their effects on generated outputs and show that selecting features by their output effects can improve steering. SAE-TS constructs steering vectors that target specific SAE features while limiting side effects~\citep{chalnev2024improving}. In image editing, PIE-Bench uses annotated masks to evaluate edit-region fidelity separately from background preservation~\citep{ju2024pnp}. D-Scope evaluates directions retrieved from target text queries and a precomputed visual index, without requiring target images or target-generation activations during feature selection. Its benchmark uses one retrieved decoder direction per intervention and measures regional target alignment separately from outside-region preservation across fine-grained concepts and source contexts. Since retrieval relies on activating visual evidence, this evaluation tests whether the selected direction produces the queried output effect. The common single-feature budget supports comparisons of dictionary and retrieval choices. For each case and generation seed, reported best-of-sweep results select the strength that maximizes regional target alignment and measure outside-region preservation on the same output.

\section{SAE Training and Dictionary Evaluation}
\label{app:sae-training}

\subsection{Training Configurations and Optimization}
\label{app:sae-training-configurations}

We train SAEs on post-block residual activations from frozen SANA-Sprint
and Nitro-1-PixArt. Our sweep covers five layers
(\(2,8,14,20,26\)), three SAE families (L1, TopK, and JumpReLU),
and five input settings, yielding 150 dictionaries in total.
All runs use the same dictionary width, data splits, and optimization
settings unless specified otherwise. Table~\ref{tab:sae-training-configurations}
summarizes the configurations required for reproduction.

\textbf{Input settings.}
Let $\mathbf{h}\in\mathbb{R}^{D}$, with $D=1{,}152$, denote a raw post-block residual-stream activation. The five input settings specify the input-processing operator $\mathcal{T}$, with $\mathbf{x}=\mathcal{T}(\mathbf{h})$, and the initialization of the shared bias $\mathbf{b}_{\mathrm{dec}}$:
\begin{equation*}
\mathcal T(\mathbf h)=
\begin{cases}
\mathbf h,
& \text{None, Mean center, Geometric median},\\[2pt]
\displaystyle
\sqrt{D}\,
\frac{\mathbf h-\bar h\mathbf 1}
{\sqrt{D^{-1}\|\mathbf h-\bar h\mathbf 1\|_2^2+\varepsilon}},
& \text{LayerNorm},\\[8pt]
\mathbf h/s,
& \text{RMS}.
\end{cases}
\end{equation*}
Here, $\bar{h}$ is the channel mean of $\mathbf{h}$, $\mathbf{1}$ is the all-ones vector, and $\varepsilon=10^{-6}$ ensures numerical stability. The LayerNorm setting therefore standardizes each token and applies an additional $\sqrt{D}$ rescaling, so that $\|\mathbf{x}\|_2\approx D$. The dataset-level RMS scale is
\begin{equation*}
s=\max\left\{
\left(
\frac{1}{ND}\sum_{n=1}^{N}\|\mathbf h_n\|_2^2
\right)^{1/2},
\varepsilon
\right\}.
\end{equation*}
The shared bias $\mathbf{b}_{\mathrm{dec}}$ is initialized to $\mathbf{0}$ except under Mean center and Geometric median, which use the activation mean $\boldsymbol{\mu}=N^{-1}\sum_{n=1}^{N}\mathbf{h}_n$ and the geometric median $\mathbf{g}=\arg\min_{\mathbf{y}}\sum_{n=1}^{N}\|\mathbf{h}_n-\mathbf{y}\|_2$, respectively; $\mathbf{g}$ is approximated with at most 128 Weiszfeld iterations initialized at $\boldsymbol{\mu}$. These statistics are estimated once from $N=65{,}536$ training activation tokens. The operator $\mathcal{T}$ remains fixed during SAE training, while $\mathbf{b}_{\mathrm{dec}}$ is optimized with the other SAE parameters. The corresponding map $\mathcal{A}_{\mathcal{T}}$ for decoder directions, used in steering, is given in Appendix~\ref{app:steering-protocol}.

\textbf{Training objectives.}
All families use the encoder and decoder of Section~\ref{sec:sae_features}, with pre-activations $\boldsymbol{\pi}=\mathbf{W}_{\mathrm{enc}}(\mathbf{x}-\mathbf{b}_{\mathrm{dec}})+\mathbf{b}_{\mathrm{enc}}$. For one activation vector, the reconstruction loss is $\mathcal{L}_{\mathrm{rec}}=D^{-1}\|\mathbf{x}-\hat{\mathbf{x}}\|_2^2$. The families differ in their activation rule and additional loss term:
\begin{align*}
\text{L1:}\quad
& \mathbf{z}=\operatorname{ReLU}(\boldsymbol{\pi}), &
\mathcal{L}
&=\mathcal{L}_{\mathrm{rec}}+\lambda_u\|\mathbf{z}\|_1,\\
\text{TopK:}\quad
& \mathbf{z}=\operatorname{TopK}_{k}\!\left(\operatorname{ReLU}(\boldsymbol{\pi})\right), &
\mathcal{L}
&=\mathcal{L}_{\mathrm{rec}}+\frac{\alpha}{D}\|\mathbf{e}-\hat{\mathbf{e}}\|_2^2,\\
\text{JumpReLU:}\quad
& z_i=\pi_i H(\pi_i-\theta_i), &
\mathcal{L}
&=\mathcal{L}_{\mathrm{rec}}+\lambda_u\sum_i H(\pi_i-\theta_i).
\end{align*}
The objectives are averaged over the minibatch. TopK retains the $k=64$ largest rectified activations and sets the remaining coordinates to zero. Its AuxK term~\citep{gao2025scaling} reconstructs the residual $\mathbf{e}=\mathbf{x}-\hat{\mathbf{x}}$, treated as a stop-gradient target. The auxiliary reconstruction $\hat{\mathbf{e}}$ uses the corresponding decoder columns and up to $k_{\mathrm{aux}}=64$ largest rectified activations among features marked inactive, that is, features with no positive activation in the preceding 256,000 training tokens. The coefficient is $\alpha=1/32$, and the auxiliary loss is zero when no features satisfy the inactivity criterion.
JumpReLU learns a positive threshold $\theta_i=\exp(\tau_i)$ per feature, with $H(a)=\mathbb{I}[a>0]$, and estimates gradients through $H$ with straight-through pseudo-derivatives~\citep{rajamanoharan2024jumping}.
For L1 and JumpReLU, the sparsity coefficient at training update $u$ is $\lambda_u=\lambda\min(1,u/10^4)$, and the sparsity hyperparameters are tuned toward a common sparsity range (Table~\ref{tab:sae-training-configurations}).

\begin{table}[t]
\centering
\caption{\textbf{SAE training configurations.}
All formal runs use a fresh initialization and one training seed.}
\label{tab:sae-training-configurations}
\small
\setlength{\tabcolsep}{4pt}
\renewcommand{\arraystretch}{1.05}
\begin{tabularx}{\linewidth}{
    @{}>{\raggedright\arraybackslash}p{0.30\linewidth}
    >{\raggedright\arraybackslash}X@{}}
\toprule
Setting & Configuration \\
\midrule

\multicolumn{2}{@{}l}{\textit{Experimental grid}} \\
Models & SANA-Sprint 0.6B; Nitro-1-PixArt 0.6B \\
Layers & 2, 8, 14, 20, 26 (0-indexed) \\
Activation & Post-block residual, one-step generation \\
Width & 1,152 input; 18,432 dictionary (16$\times$) \\

\midrule
\multicolumn{2}{@{}l}{\textit{Input settings}} \\
None & Raw residual activations; $\mathbf{b}_{\mathrm{dec}}$ initialized to $\mathbf{0}$ \\
Mean center & Raw activations; $\mathbf{b}_{\mathrm{dec}}$ initialized to the activation mean \\
Geometric median & Raw activations; $\mathbf{b}_{\mathrm{dec}}$ initialized to the geometric median \\
LayerNorm & Per-token standardization, rescaled by \(\sqrt{1152}\) \\
RMS & Division by a dataset-level RMS scale \\

\midrule
\multicolumn{2}{@{}l}{\textit{SAE families}} \\
L1 & ReLU codes; tuned L1 coefficient \(\lambda\) \\
TopK & \(k=64\); AuxK with \(k_{\mathrm{aux}}=64\) and \(\alpha=1/32\) \\
JumpReLU & Learned thresholds; rectangle-kernel pseudo-derivative bandwidth \(10^{-3}\); tuned \(\lambda\) and initial-\(\ell_0\) target \(\ell_0^{\mathrm{init}}\) \\

\midrule
\multicolumn{2}{@{}l}{\textit{Training}} \\
Prompts & 500k ReLAION-COCO prompts \\
Split & 475k / 12.5k / 12.5k train/val/eval \\
Calibration & 65,536 training activation tokens for the scale or bias initialization; 4,096 of them for JumpReLU threshold initialization \\
Sparsity tuning & Per generator, layer, and input setting: lowest validation MSE subject to validation \(\ell_0\in[64,100]\), using short 20k-update runs \\
Optimizer & AdamW (\(\beta_1{=}0.9\), \(\beta_2{=}0.999\), weight decay \(0.01\)); constant learning rate \(10^{-4}\) \\
Batch / updates & 4,096 activation tokens; 100k updates \\
Warmup & 10k-update linear warmup of \(\lambda\) for L1 and JumpReLU \\
Inactivity window & 256,000 training tokens (AuxK and utilization) \\
Checkpoint & Lowest validation MSE among checkpoints saved every 10k updates \\
Decoder & Unit-norm columns after each update \\

\bottomrule
\end{tabularx}
\end{table}

\subsection{Dictionary Evaluation}
\label{app:dictionary-evaluation}

Dictionary evaluation separates reconstruction fidelity, realized sparsity, utilization, visual coherence, and evidence coverage. For reconstruction and sparsity, we evaluate the validation-selected checkpoints of all 150 dictionaries on the same 1,000 held-out prompts within each generator. Fixed single-step generations provide all spatial-token activations, yielding 1,024,000 evaluation tokens for SANA-Sprint and 4,096,000 for Nitro-1-PixArt.

\textbf{Reconstruction and sparsity.} Let $\mathbf{X},\hat{\mathbf{X}}\in\mathbb{R}^{N\times D}$ denote the SAE inputs and their reconstructions, and let $\mathbf{Z}\in\mathbb{R}^{N\times m}$ contain the corresponding feature activations. Reconstruction fidelity is measured by explained variance:
\begin{equation*}
\mathrm{EV}
=
1-
\frac{\operatorname{Var}_{\mathrm{flat}}(\mathbf{X}-\hat{\mathbf{X}})}
{\operatorname{Var}_{\mathrm{flat}}(\mathbf{X})},
\end{equation*}
where $\operatorname{Var}_{\mathrm{flat}}$ denotes variance over all scalar entries, pooling the token and channel dimensions. Reconstruction is evaluated in the SAE input coordinates defined by $\mathcal{T}$, with raw-residual-space results reported separately where available. Realized sparsity is the mean number of positive feature activations per token:
\begin{equation*}
\ell_0
=
\frac{1}{N}
\sum_{n=1}^{N}\sum_{i=1}^{m}
\mathbb{I}[Z_{ni}>0],
\end{equation*}
where $\mathbb{I}[\cdot]$ is the indicator function.

\textbf{Dictionary utilization.} Utilization measures the fraction of dictionary features recorded as active within the preceding 256,000 training tokens at the selected checkpoint. Let $\mathcal{U}$ denote this set of features, identified using batch-level activity timestamps. The reported utilization is
\begin{equation*}
\mathrm{Utilization}=\frac{|\mathcal{U}|}{m}.
\end{equation*}
This statistic summarizes feature use during training and is distinct from the active-feature fraction on the held-out evaluation set.

\textbf{Visual coherence and evidence coverage.} For each dictionary, we deterministically and uniformly sample 184 feature indices before filtering for evidence availability. For each sampled feature, we identify its maximally activating token in each image and retain the $J=64$ distinct images with the highest positive responses. A feature is eligible only if all $J$ images provide positive activations. Evidence patches cover local regions of $4\times4$ visual-token cells around the selected locations, clipped at image boundaries.

Let $P_{ij}$ denote evidence patch $j$ for feature $i$. The frozen DINOv3 ViT-L/16 encoder $E_D$~\citep{simeoni2025dinov3} produces the unit-normalized patch embedding $\mathbf{e}_{ij}=\nu(E_D(P_{ij}))$, using the normalization operator defined in Section~\ref{sec:feature_retrieval}. Feature coherence is the mean cosine similarity over all distinct patch pairs:
\begin{equation*}
c_i
=
\frac{2}{J(J-1)}
\sum_{1\le j<k\le J}
\mathbf{e}_{ij}^{\top}\mathbf{e}_{ik}.
\end{equation*}
For the eligible subset $\mathcal{E}$ of the sampled features, dictionary-level coherence and evidence coverage are
\begin{equation*}
C
=
\frac{1}{|\mathcal{E}|}
\sum_{i\in\mathcal{E}} c_i,
\qquad
\mathrm{Coverage}
=
\frac{|\mathcal{E}|}{184}.
\end{equation*}
We also compute coherence using the frozen SigLIP~2 image encoder~\citep{tschannen2025siglip} on the same evidence patches as an auxiliary measure. If no sampled feature is eligible, coherence is undefined and reported as N/A, while coverage is zero. Coherence therefore describes visual consistency among eligible features, whereas coverage measures the availability of sufficient evidence across the sampled dictionary.

\subsection{Additional Analyses of Dictionary Quality}
\label{app:sae-quality-analysis}

\begin{figure}[t]
    \centering
    \includegraphics[width=\linewidth]{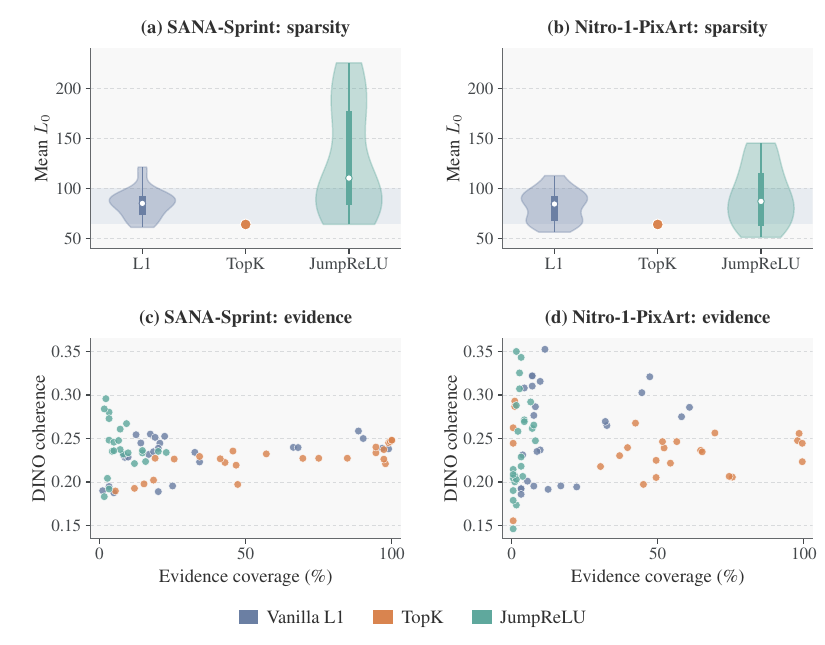}
    \caption{\textbf{Realized sparsity and visual evidence.} \textbf{(a,b)} Sparsity distributions across layers and input settings. White dots mark medians, thick bars indicate interquartile ranges, and thin bars span observed ranges. Shading marks the tuning target $[64,100]$. TopK is shown as a single point at $\ell_0=64$. \textbf{(c,d)} DINOv3 coherence versus evidence coverage, with each point representing a dictionary with eligible visual evidence.}
    \label{fig:app-sae-sparsity-evidence}
\end{figure}

\textbf{Realized sparsity.} Figure~\ref{fig:app-sae-sparsity-evidence}(a,b) shows the distributions of achieved sparsity separately for each generator. TopK achieves a mean $\ell_0$ of 64 in every configuration, whereas the mean number of active features ranges from 56.4 to 121.4 for L1 and from 51.0 to 225.6 for JumpReLU. Although pilot selection targets a common sparsity range, achieved sparsity varies after full training. Comparisons between SAE families should therefore be interpreted alongside their realized sparsity.

\textbf{Visual coherence and evidence availability.} Figure~\ref{fig:app-sae-sparsity-evidence}(c,d) relates dictionary-level DINOv3 coherence to evidence coverage. High coherence can coexist with limited coverage because coherence is computed only over features with sufficient visual evidence. Across the two generators and five layers, mean-centered TopK and mean-centered JumpReLU have median dictionary-level coherence of 0.225 and 0.234, respectively, while their median coverage is 75.3\% and 6.0\%. Coherence thus characterizes the consistency of eligible features, while coverage measures the fraction of sampled features for which such evidence is available.

\textbf{Scope of the training recommendation.} Across the two generators and five layers, mean-centered and geometric-median-centered TopK achieve median utilization of 98.6\% and 98.9\%, respectively, with median coverage of 75.3\% and 72.0\%, the two highest values among the evaluated training recipes. These results support centered TopK as a starting point for obtaining broadly utilized dictionaries with substantial visual evidence. The preferred recipe can nevertheless depend on the activation layer. At layer 2 of Nitro-1-PixArt, geometric-median-centered L1 achieves 12.5\% coverage, while both centered TopK variants achieve 0.5\%. Layer-specific evaluation therefore remains necessary when selecting a dictionary.

\begin{figure}[t]
    \centering
    \includegraphics[width=\linewidth]{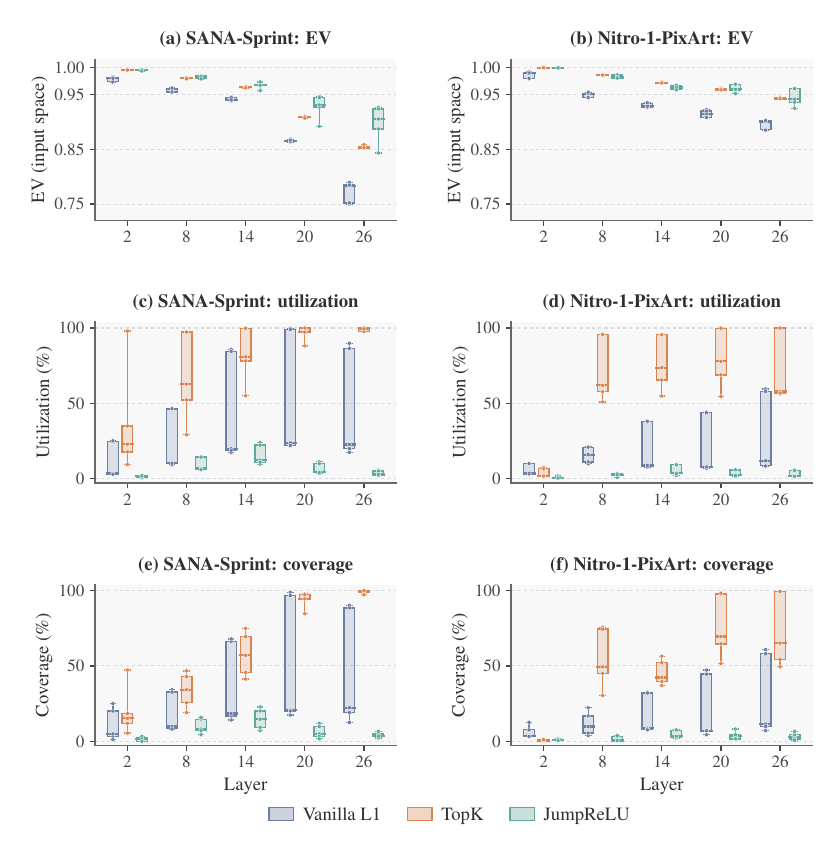}
    \caption{\textbf{Dictionary quality across transformer layers.} Columns correspond to generators. Rows show SAE input-space EV, dictionary utilization, and evidence coverage. Each box summarizes five input settings within a layer and SAE family. Boxes indicate interquartile ranges, center lines mark medians, whiskers span observed ranges, and dots show individual settings.}
    \label{fig:app-sae-depth-effects}
\end{figure}

\textbf{Effects of dictionary depth.} Figure~\ref{fig:app-sae-depth-effects} shows reconstruction EV, utilization, and evidence coverage across all five layers, with each box summarizing the five input settings for a given generator and SAE family. In a comparison of matched configurations, layer 26 has lower SAE input-space EV and higher utilization than layer 2 in all 30 pairs. Evidence coverage increases in 29 pairs and remains unchanged in one. These endpoint results indicate that stronger reconstruction does not necessarily coincide with broader dictionary use or greater evidence availability.

The changes across intermediate layers are not uniformly monotonic. For mean-centered TopK in Nitro-1-PixArt, coverage decreases from 75.5\% at layer 8 to 52.2\% at layer 14, then increases to 97.8\% at layer 20 and 99.5\% at layer 26. Dictionary selection therefore benefits from examining reconstruction, utilization, and evidence coverage jointly at each candidate layer.

\subsection{Complete Dictionary Evaluation Results}
\label{app:sae-complete-results}

Tables~\ref{tab:app-sae-sana-results} and~\ref{tab:app-sae-nitro-results} report all configurations for SANA-Sprint and Nitro-1-PixArt. Utilization and coverage are expressed as percentages. DINO and SigLIP denote dictionary-level mean coherence over eligible features. N/A indicates unavailable raw-space evaluation or undefined coherence when no sampled feature meets the evidence requirement. Recipe-level summaries and matched comparisons are computed from unrounded values.

\begingroup
\footnotesize
\setlength{\tabcolsep}{3pt}
\renewcommand{\arraystretch}{1.05}
\begin{longtable}{@{}rllrrrrrrr@{}}
\caption{\textbf{Complete SAE results for SANA-Sprint.}}\label{tab:app-sae-sana-results}\\
\toprule
Layer & SAE & Input & EV (input) & EV (raw) & Mean $\ell_0$ & Util. (\%) & Cov. (\%) & DINO & SigLIP \\
\midrule
\endfirsthead
\multicolumn{10}{@{}l}{\tablename\ \thetable\ (continued)}\\
\toprule
Layer & SAE & Input & EV (input) & EV (raw) & Mean $\ell_0$ & Util. (\%) & Cov. (\%) & DINO & SigLIP \\
\midrule
\endhead
\midrule
\multicolumn{10}{r@{}}{\textit{Continued on next page.}}\\
\endfoot
\bottomrule
\endlastfoot
\addlinespace[2pt]
2 & L1 & None & 0.9824 & 0.9824 & 94.0 & 3.4 & 3.3 & 0.195 & 0.726 \\*
 & & Mean center & 0.9726 & 0.9726 & 115.0 & 24.5 & 25.0 & 0.195 & 0.732 \\*
 & & Geo. center & 0.9738 & 0.9738 & 121.4 & 25.1 & 20.1 & 0.189 & 0.731 \\*
 & & LayerNorm & 0.9823 & N/A & 95.3 & 3.4 & 4.9 & 0.188 & 0.724 \\*
 & & RMS & 0.9803 & 0.9803 & 81.7 & 2.9 & 1.1 & 0.190 & 0.720 \\
\addlinespace[2pt]
2 & TopK & None & 0.9956 & 0.9956 & 64.0 & 17.7 & 12.0 & 0.193 & 0.730 \\*
 & & Mean center & 0.9958 & 0.9958 & 64.0 & 35.0 & 18.5 & 0.202 & 0.731 \\*
 & & Geo. center & 0.9958 & 0.9958 & 64.0 & 97.9 & 47.3 & 0.197 & 0.728 \\*
 & & LayerNorm & 0.9957 & N/A & 64.0 & 22.9 & 15.2 & 0.198 & 0.733 \\*
 & & RMS & 0.9953 & 0.9953 & 64.0 & 9.3 & 5.4 & 0.190 & 0.728 \\
\addlinespace[2pt]
2 & JumpReLU & None & 0.9942 & 0.9942 & 76.7 & 0.9 & 0.0 & N/A & N/A \\*
 & & Mean center & 0.9955 & 0.9955 & 65.8 & 1.9 & 2.7 & 0.204 & 0.742 \\*
 & & Geo. center & 0.9955 & 0.9955 & 65.8 & 1.9 & 3.3 & 0.192 & 0.726 \\*
 & & LayerNorm & 0.9956 & N/A & 92.5 & 0.9 & 1.6 & 0.183 & 0.742 \\*
 & & RMS & 0.9936 & 0.9936 & 82.2 & 2.0 & 0.0 & N/A & N/A \\
\midrule
\addlinespace[2pt]
8 & L1 & None & 0.9619 & 0.9619 & 84.2 & 10.3 & 8.7 & 0.228 & 0.742 \\*
 & & Mean center & 0.9551 & 0.9551 & 73.1 & 46.3 & 34.2 & 0.223 & 0.744 \\*
 & & Geo. center & 0.9550 & 0.9550 & 73.3 & 46.4 & 32.6 & 0.234 & 0.749 \\*
 & & LayerNorm & 0.9605 & N/A & 84.3 & 9.2 & 9.8 & 0.229 & 0.738 \\*
 & & RMS & 0.9624 & 0.9624 & 85.1 & 10.2 & 8.2 & 0.231 & 0.743 \\
\addlinespace[2pt]
8 & TopK & None & 0.9801 & 0.9801 & 64.0 & 52.3 & 25.5 & 0.226 & 0.741 \\*
 & & Mean center & 0.9811 & 0.9811 & 64.0 & 97.3 & 46.7 & 0.219 & 0.742 \\*
 & & Geo. center & 0.9811 & 0.9811 & 64.0 & 97.3 & 42.9 & 0.222 & 0.740 \\*
 & & LayerNorm & 0.9793 & N/A & 64.0 & 62.7 & 34.2 & 0.229 & 0.737 \\*
 & & RMS & 0.9795 & 0.9795 & 64.0 & 29.2 & 19.0 & 0.227 & 0.739 \\
\addlinespace[2pt]
8 & JumpReLU & None & 0.9794 & 0.9794 & 83.3 & 6.8 & 8.2 & 0.232 & 0.739 \\*
 & & Mean center & 0.9845 & 0.9845 & 112.6 & 14.5 & 14.7 & 0.234 & 0.741 \\*
 & & Geo. center & 0.9845 & 0.9845 & 112.7 & 14.3 & 15.8 & 0.224 & 0.733 \\*
 & & LayerNorm & 0.9833 & N/A & 136.3 & 6.3 & 7.1 & 0.237 & 0.738 \\*
 & & RMS & 0.9792 & 0.9792 & 87.0 & 5.9 & 4.3 & 0.235 & 0.750 \\
\midrule
\addlinespace[2pt]
14 & L1 & None & 0.9393 & 0.9393 & 61.9 & 19.9 & 16.8 & 0.232 & 0.740 \\*
 & & Mean center & 0.9456 & 0.9456 & 87.0 & 85.7 & 66.3 & 0.240 & 0.750 \\*
 & & Geo. center & 0.9453 & 0.9453 & 87.2 & 84.5 & 67.9 & 0.240 & 0.742 \\*
 & & LayerNorm & 0.9392 & N/A & 61.0 & 17.3 & 14.1 & 0.245 & 0.744 \\*
 & & RMS & 0.9409 & 0.9409 & 69.0 & 18.8 & 18.5 & 0.235 & 0.743 \\
\addlinespace[2pt]
14 & TopK & None & 0.9635 & 0.9635 & 64.0 & 78.2 & 45.7 & 0.236 & 0.736 \\*
 & & Mean center & 0.9649 & 0.9649 & 64.0 & 99.8 & 75.0 & 0.227 & 0.732 \\*
 & & Geo. center & 0.9649 & 0.9649 & 64.0 & 99.8 & 69.6 & 0.227 & 0.731 \\*
 & & LayerNorm & 0.9643 & N/A & 64.0 & 80.9 & 57.1 & 0.232 & 0.737 \\*
 & & RMS & 0.9627 & 0.9627 & 64.0 & 55.0 & 41.3 & 0.227 & 0.731 \\
\addlinespace[2pt]
14 & JumpReLU & None & 0.9668 & 0.9668 & 110.3 & 12.3 & 14.7 & 0.236 & 0.729 \\*
 & & Mean center & 0.9676 & 0.9676 & 94.4 & 24.0 & 22.8 & 0.234 & 0.737 \\*
 & & Geo. center & 0.9674 & 0.9674 & 94.2 & 22.2 & 20.1 & 0.235 & 0.738 \\*
 & & LayerNorm & 0.9737 & N/A & 169.9 & 9.6 & 7.1 & 0.261 & 0.755 \\*
 & & RMS & 0.9575 & 0.9575 & 64.2 & 10.9 & 9.2 & 0.267 & 0.759 \\
\midrule
\addlinespace[2pt]
20 & L1 & None & 0.8660 & 0.8660 & 90.9 & 23.7 & 20.7 & 0.245 & 0.742 \\*
 & & Mean center & 0.8637 & 0.8637 & 79.8 & 99.7 & 98.9 & 0.238 & 0.741 \\*
 & & Geo. center & 0.8648 & 0.8648 & 82.3 & 99.1 & 96.7 & 0.239 & 0.741 \\*
 & & LayerNorm & 0.8688 & N/A & 92.8 & 21.8 & 17.4 & 0.255 & 0.756 \\*
 & & RMS & 0.8669 & 0.8669 & 91.6 & 22.0 & 20.1 & 0.239 & 0.737 \\
\addlinespace[2pt]
20 & TopK & None & 0.9079 & 0.9079 & 64.0 & 97.0 & 94.6 & 0.234 & 0.731 \\*
 & & Mean center & 0.9090 & 0.9090 & 64.0 & 100.0 & 97.8 & 0.221 & 0.723 \\*
 & & Geo. center & 0.9090 & 0.9090 & 64.0 & 100.0 & 97.3 & 0.226 & 0.729 \\*
 & & LayerNorm & 0.9099 & N/A & 64.0 & 97.2 & 94.6 & 0.240 & 0.736 \\*
 & & RMS & 0.9074 & 0.9074 & 64.0 & 88.1 & 84.8 & 0.227 & 0.728 \\
\addlinespace[2pt]
20 & JumpReLU & None & 0.9277 & 0.9277 & 165.5 & 4.4 & 4.9 & 0.236 & 0.735 \\*
 & & Mean center & 0.9469 & 0.9469 & 180.1 & 11.2 & 12.0 & 0.221 & 0.729 \\*
 & & Geo. center & 0.9449 & 0.9449 & 177.7 & 10.0 & 9.8 & 0.234 & 0.743 \\*
 & & LayerNorm & 0.9314 & N/A & 197.3 & 3.4 & 3.3 & 0.280 & 0.761 \\*
 & & RMS & 0.8924 & 0.8924 & 73.8 & 4.0 & 1.6 & 0.284 & 0.754 \\
\midrule
\addlinespace[2pt]
26 & L1 & None & 0.7838 & 0.7838 & 94.7 & 19.9 & 22.3 & 0.253 & 0.743 \\*
 & & Mean center & 0.7502 & 0.7502 & 67.0 & 86.4 & 90.2 & 0.250 & 0.742 \\*
 & & Geo. center & 0.7520 & 0.7520 & 68.5 & 89.8 & 88.6 & 0.259 & 0.746 \\*
 & & LayerNorm & 0.7895 & N/A & 90.9 & 17.5 & 19.0 & 0.251 & 0.745 \\*
 & & RMS & 0.7849 & 0.7849 & 94.4 & 22.7 & 12.5 & 0.254 & 0.745 \\
\addlinespace[2pt]
26 & TopK & None & 0.8527 & 0.8527 & 64.0 & 97.8 & 98.9 & 0.245 & 0.733 \\*
 & & Mean center & 0.8537 & 0.8537 & 64.0 & 100.0 & 99.5 & 0.246 & 0.730 \\*
 & & Geo. center & 0.8536 & 0.8536 & 64.0 & 100.0 & 100.0 & 0.248 & 0.730 \\*
 & & LayerNorm & 0.8586 & N/A & 64.0 & 97.8 & 97.3 & 0.237 & 0.731 \\*
 & & RMS & 0.8531 & 0.8531 & 64.0 & 99.4 & 100.0 & 0.248 & 0.731 \\
\addlinespace[2pt]
26 & JumpReLU & None & 0.9057 & 0.9057 & 218.0 & 2.8 & 3.3 & 0.248 & 0.730 \\*
 & & Mean center & 0.9271 & 0.9271 & 222.2 & 5.6 & 4.9 & 0.246 & 0.733 \\*
 & & Geo. center & 0.9242 & 0.9242 & 225.6 & 5.0 & 6.5 & 0.248 & 0.736 \\*
 & & LayerNorm & 0.8874 & N/A & 198.8 & 2.1 & 2.2 & 0.296 & 0.782 \\*
 & & RMS & 0.8437 & 0.8437 & 95.2 & 2.7 & 3.3 & 0.273 & 0.747 \\
\end{longtable}
\endgroup

\begingroup
\footnotesize
\setlength{\tabcolsep}{3pt}
\renewcommand{\arraystretch}{1.05}
\begin{longtable}{@{}rllrrrrrrr@{}}
\caption{\textbf{Complete SAE results for Nitro-1-PixArt.}}\label{tab:app-sae-nitro-results}\\
\toprule
Layer & SAE & Input & EV (input) & EV (raw) & Mean $\ell_0$ & Util. (\%) & Cov. (\%) & DINO & SigLIP \\
\midrule
\endfirsthead
\multicolumn{10}{@{}l}{\tablename\ \thetable\ (continued)}\\
\toprule
Layer & SAE & Input & EV (input) & EV (raw) & Mean $\ell_0$ & Util. (\%) & Cov. (\%) & DINO & SigLIP \\
\midrule
\endhead
\midrule
\multicolumn{10}{r@{}}{\textit{Continued on next page.}}\\
\endfoot
\bottomrule
\endlastfoot
\addlinespace[2pt]
2 & L1 & None & 0.9922 & 0.9922 & 92.7 & 3.4 & 3.3 & 0.193 & 0.782 \\*
 & & Mean center & 0.9798 & 0.9798 & 67.3 & 9.9 & 7.6 & 0.195 & 0.781 \\*
 & & Geo. center & 0.9801 & 0.9801 & 67.0 & 10.0 & 12.5 & 0.192 & 0.784 \\*
 & & LayerNorm & 0.9909 & N/A & 83.3 & 2.9 & 3.3 & 0.192 & 0.780 \\*
 & & RMS & 0.9899 & 0.9899 & 84.4 & 3.4 & 3.3 & 0.186 & 0.777 \\
\addlinespace[2pt]
2 & TopK & None & 0.9997 & 0.9997 & 64.0 & 1.5 & 0.5 & 0.155 & 0.767 \\*
 & & Mean center & 0.9997 & 0.9997 & 64.0 & 7.4 & 0.5 & 0.245 & 0.811 \\*
 & & Geo. center & 0.9997 & 0.9997 & 64.0 & 6.5 & 0.5 & 0.262 & 0.838 \\*
 & & LayerNorm & 0.9997 & N/A & 64.0 & 1.8 & 1.1 & 0.293 & 0.808 \\*
 & & RMS & 0.9997 & 0.9997 & 64.0 & 1.7 & 1.1 & 0.287 & 0.806 \\
\addlinespace[2pt]
2 & JumpReLU & None & 0.9993 & 0.9993 & 51.0 & 0.3 & 1.1 & 0.200 & 0.778 \\*
 & & Mean center & 0.9995 & 0.9995 & 111.4 & 1.8 & 0.5 & 0.214 & 0.781 \\*
 & & Geo. center & 0.9997 & 0.9997 & 80.8 & 0.6 & 1.6 & 0.174 & 0.777 \\*
 & & LayerNorm & 0.9995 & N/A & 61.6 & 0.5 & 1.1 & 0.207 & 0.784 \\*
 & & RMS & 0.9992 & 0.9992 & 61.3 & 0.4 & 0.5 & 0.179 & 0.766 \\
\midrule
\addlinespace[2pt]
8 & L1 & None & 0.9520 & 0.9520 & 95.4 & 9.8 & 3.8 & 0.231 & 0.793 \\*
 & & Mean center & 0.9448 & 0.9448 & 112.9 & 20.8 & 16.8 & 0.196 & 0.782 \\*
 & & Geo. center & 0.9450 & 0.9450 & 112.7 & 20.9 & 22.3 & 0.194 & 0.784 \\*
 & & LayerNorm & 0.9522 & N/A & 96.8 & 15.9 & 9.8 & 0.237 & 0.799 \\*
 & & RMS & 0.9548 & 0.9548 & 102.7 & 11.1 & 5.4 & 0.201 & 0.783 \\
\addlinespace[2pt]
8 & TopK & None & 0.9861 & 0.9861 & 64.0 & 61.9 & 49.5 & 0.205 & 0.785 \\*
 & & Mean center & 0.9864 & 0.9864 & 64.0 & 95.7 & 75.5 & 0.206 & 0.785 \\*
 & & Geo. center & 0.9865 & 0.9865 & 64.0 & 95.7 & 74.5 & 0.206 & 0.786 \\*
 & & LayerNorm & 0.9862 & N/A & 64.0 & 50.9 & 30.4 & 0.218 & 0.789 \\*
 & & RMS & 0.9862 & 0.9862 & 64.0 & 57.6 & 45.1 & 0.197 & 0.781 \\
\addlinespace[2pt]
8 & JumpReLU & None & 0.9804 & 0.9804 & 94.2 & 2.4 & 0.5 & 0.146 & 0.773 \\*
 & & Mean center & 0.9863 & 0.9863 & 80.6 & 3.3 & 3.3 & 0.218 & 0.783 \\*
 & & Geo. center & 0.9865 & 0.9865 & 85.0 & 3.3 & 3.8 & 0.206 & 0.789 \\*
 & & LayerNorm & 0.9818 & N/A & 87.0 & 0.8 & 0.5 & 0.190 & 0.770 \\*
 & & RMS & 0.9808 & 0.9808 & 96.9 & 2.6 & 0.5 & 0.205 & 0.787 \\
\midrule
\addlinespace[2pt]
14 & L1 & None & 0.9346 & 0.9346 & 66.2 & 8.7 & 8.7 & 0.235 & 0.797 \\*
 & & Mean center & 0.9271 & 0.9271 & 63.3 & 37.8 & 32.6 & 0.265 & 0.807 \\*
 & & Geo. center & 0.9276 & 0.9276 & 67.6 & 38.1 & 32.1 & 0.270 & 0.811 \\*
 & & LayerNorm & 0.9292 & N/A & 56.4 & 7.7 & 8.2 & 0.287 & 0.816 \\*
 & & RMS & 0.9354 & 0.9354 & 68.0 & 8.7 & 7.6 & 0.277 & 0.814 \\
\addlinespace[2pt]
14 & TopK & None & 0.9721 & 0.9721 & 64.0 & 65.5 & 39.7 & 0.240 & 0.794 \\*
 & & Mean center & 0.9731 & 0.9731 & 64.0 & 95.5 & 52.2 & 0.239 & 0.797 \\*
 & & Geo. center & 0.9731 & 0.9731 & 64.0 & 95.5 & 56.5 & 0.246 & 0.802 \\*
 & & LayerNorm & 0.9718 & N/A & 64.0 & 73.7 & 42.4 & 0.268 & 0.807 \\*
 & & RMS & 0.9717 & 0.9717 & 64.0 & 54.8 & 37.0 & 0.230 & 0.793 \\
\addlinespace[2pt]
14 & JumpReLU & None & 0.9606 & 0.9606 & 62.6 & 3.6 & 2.2 & 0.258 & 0.805 \\*
 & & Mean center & 0.9677 & 0.9677 & 61.4 & 9.4 & 7.1 & 0.262 & 0.800 \\*
 & & Geo. center & 0.9675 & 0.9675 & 61.0 & 9.2 & 7.6 & 0.265 & 0.803 \\*
 & & LayerNorm & 0.9634 & N/A & 91.3 & 2.1 & 2.7 & 0.307 & 0.817 \\*
 & & RMS & 0.9592 & 0.9592 & 58.7 & 3.3 & 3.3 & 0.229 & 0.790 \\
\midrule
\addlinespace[2pt]
20 & L1 & None & 0.9228 & 0.9228 & 97.7 & 7.7 & 7.1 & 0.310 & 0.830 \\*
 & & Mean center & 0.9086 & 0.9086 & 79.8 & 44.0 & 47.3 & 0.321 & 0.831 \\*
 & & Geo. center & 0.9086 & 0.9086 & 79.5 & 43.8 & 44.6 & 0.303 & 0.824 \\*
 & & LayerNorm & 0.9156 & N/A & 86.2 & 6.9 & 7.1 & 0.322 & 0.830 \\*
 & & RMS & 0.9201 & 0.9201 & 92.9 & 7.8 & 4.3 & 0.308 & 0.832 \\
\addlinespace[2pt]
20 & TopK & None & 0.9596 & 0.9596 & 64.0 & 68.9 & 64.7 & 0.236 & 0.798 \\*
 & & Mean center & 0.9616 & 0.9616 & 64.0 & 99.9 & 97.8 & 0.248 & 0.803 \\*
 & & Geo. center & 0.9616 & 0.9616 & 64.0 & 99.9 & 98.4 & 0.256 & 0.805 \\*
 & & LayerNorm & 0.9595 & N/A & 64.0 & 77.8 & 69.6 & 0.256 & 0.808 \\*
 & & RMS & 0.9592 & 0.9592 & 64.0 & 54.6 & 51.6 & 0.246 & 0.802 \\
\addlinespace[2pt]
20 & JumpReLU & None & 0.9589 & 0.9589 & 111.7 & 2.3 & 3.3 & 0.343 & 0.832 \\*
 & & Mean center & 0.9684 & 0.9684 & 115.2 & 5.8 & 8.2 & 0.247 & 0.803 \\*
 & & Geo. center & 0.9695 & 0.9695 & 121.7 & 5.9 & 4.3 & 0.271 & 0.805 \\*
 & & LayerNorm & 0.9600 & N/A & 139.0 & 1.6 & 1.6 & 0.288 & 0.804 \\*
 & & RMS & 0.9523 & 0.9523 & 86.8 & 2.1 & 1.6 & 0.350 & 0.853 \\
\midrule
\addlinespace[2pt]
26 & L1 & None & 0.9031 & 0.9031 & 92.3 & 8.6 & 11.4 & 0.353 & 0.840 \\*
 & & Mean center & 0.8862 & 0.8862 & 64.9 & 59.5 & 58.2 & 0.275 & 0.812 \\*
 & & Geo. center & 0.8853 & 0.8853 & 63.7 & 57.9 & 60.9 & 0.286 & 0.815 \\*
 & & LayerNorm & 0.9011 & N/A & 88.3 & 8.3 & 9.8 & 0.316 & 0.833 \\*
 & & RMS & 0.9027 & 0.9027 & 91.2 & 11.8 & 7.1 & 0.322 & 0.840 \\
\addlinespace[2pt]
26 & TopK & None & 0.9428 & 0.9428 & 64.0 & 56.3 & 65.2 & 0.235 & 0.795 \\*
 & & Mean center & 0.9450 & 0.9450 & 64.0 & 100.0 & 99.5 & 0.223 & 0.792 \\*
 & & Geo. center & 0.9451 & 0.9451 & 64.0 & 100.0 & 99.5 & 0.245 & 0.799 \\*
 & & LayerNorm & 0.9429 & N/A & 64.0 & 57.8 & 49.5 & 0.225 & 0.794 \\*
 & & RMS & 0.9429 & 0.9429 & 64.0 & 56.8 & 54.3 & 0.222 & 0.793 \\
\addlinespace[2pt]
26 & JumpReLU & None & 0.9364 & 0.9364 & 117.9 & 1.5 & 2.7 & 0.325 & 0.816 \\*
 & & Mean center & 0.9614 & 0.9614 & 143.8 & 5.6 & 4.3 & 0.269 & 0.800 \\*
 & & Geo. center & 0.9613 & 0.9613 & 145.3 & 5.5 & 6.5 & 0.292 & 0.813 \\*
 & & LayerNorm & 0.9421 & N/A & 143.8 & 1.6 & 1.6 & 0.203 & 0.787 \\*
 & & RMS & 0.9254 & 0.9254 & 79.2 & 1.5 & 0.5 & 0.209 & 0.786 \\
\end{longtable}
\endgroup

\section{Feature Cards and Visual Index}
\label{app:visual-index}

Each of the 30 layer-14 dictionaries used for steering has a feature catalog, built once from a shared catalog corpus with the settings in Table~\ref{tab:visual-index-setup}. Construction uses no text annotations, benchmark prompts, or steering outcomes.

For feature $i$ and catalog image $n$, let $s_{in}$ be the maximal SAE activation over the visual tokens of $n$. The $J=64$ images with the largest positive $s_{in}$ supply the evidence of $i$, one patch per image, so that no image contributes more than once. These patches form the feature card of $i$; figures show its highest-ranked patches. They are also pooled into the visual centroid $\mathbf{v}_i$ as in Section~\ref{sec:feature_retrieval}, with $a_{ij}=s_{in}$ for the image $n$ containing $P_{ij}$. Feature $i$ enters the index $\mathcal{I}$ only if all $J$ evidence activations are positive, that is, if it activates in at least $J$ distinct catalog images; features with fewer positive images keep their cards and centroids but are never retrieved. For the dictionary in Figure~\ref{fig:framework}, 16,232 of 18,432 features have cards and appear in the feature atlas, and 15,754 are indexed. Index size ranges from 373 to 18,298 features across dictionaries, and 41.3\% of all 552,960 features are indexed (Table~\ref{tab:visual-index-size}). The SANA teacher reuses the SANA-Sprint indices, and the card in Figure~\ref{fig:app-nudity-feature-card} is taken from the same catalog.

\textbf{Relation to evidence coverage.} Coverage (Appendix~\ref{app:dictionary-evaluation}) applies the same rule to 184 sampled features over only the first 1,000 catalog images. It thus requires activation in at least 6.4\% of images rather than 0.64\%, a stricter criterion than index membership. Accordingly, the indexed share exceeds coverage for 25 of the 30 dictionaries, and the remaining five lie within one standard error of the 184-feature estimate. The $J$ evidence images of a card are distinct from the $K$ tokens pooled within each image in Appendix~\ref{app:nudity-readout}.

\textbf{Relation to steering.}
In every model-retrieval setting, the three weakest steering dictionaries (JumpReLU with the None, LayerNorm, and RMS settings; Table~\ref{tab:steering-main}) have the three smallest visual indices, with at most 2,255 features on SANA-Sprint and 645 on Nitro. Centering enlarges the indices of L1 and JumpReLU by $1.8$--$4.4\times$ but those of TopK by at most $1.5\times$, since uncentered TopK already indexes more than 12,000 features. Centering improves mean target alignment in 22 of 24 matched L1 and JumpReLU comparisons, while its effects on TopK are not consistently positive (Fig.~\ref{fig:steering}c; Table~\ref{tab:steering-main}). Among the remaining 12 dictionaries in each generation setting, larger indices are not consistently associated with higher mean target gains: the Spearman correlation between index size and $\Delta_{\mathrm{target}}$ ranges from 0.14 to 0.69 in the SANA environments but from $-0.76$ to $-0.48$ on Nitro. Index size is almost perfectly rank-correlated with utilization (Spearman 0.99--1.00), so these associations do not separate the size of the retrieval pool from other properties of the dictionary.

\begin{table}[t]
\centering
\caption{\textbf{Feature-catalog construction.} Settings shared by all 30 layer-14 dictionaries.}
\label{tab:visual-index-setup}
\small
\begin{tabularx}{\linewidth}{@{}lX@{}}
\toprule
Component & Setting \\
\midrule
Prompts & First 10,000 prompts of the evaluation split (Table~\ref{tab:sae-training-configurations}), disjoint from training and validation prompts; the first 1,000, with the same seeds, form the dictionary-evaluation set \\
Generation & $1024\times1024$, single step, guidance 0 (timestep 400 for Nitro-1-PixArt); \\
Activations & Layer-14 post-block residual of all visual tokens ($32\times32$ for SANA-Sprint, $64\times64$ for Nitro-1-PixArt), mapped by $\mathcal{T}$ and encoded by the SAE \\
Evidence & $J=64$ images with the largest positive per-image maximum; one token per image; ties broken by image order \\
Patch & Square of four token cells centered on the selected token, clipped at the image boundary \\
Card & Patches in an $8\times8$ grid by descending activation, for every feature with a positive image \\
Encoder & SigLIP~2 So400M/16 at $384\times384$; pooled 1,152-dimensional image embedding \\
Index & Features whose $J$ evidence activations are all positive \\
\bottomrule
\end{tabularx}
\end{table}

\begin{table}[t]
\centering
\caption{\textbf{Visual index size.} Indexed features out of 18,432 in each layer-14 dictionary, with percentages in parentheses. Retrieval considers only indexed features; the SANA teacher uses the SANA-Sprint indices.}
\label{tab:visual-index-size}
\small
\begin{tabular}{@{}llrr@{}}
\toprule
SAE & Input & SANA-Sprint & Nitro-1-PixArt \\
\midrule
L1 & None & 3,661 (19.9) & 1,583 (8.6) \\
 & Mean center & 15,754 (85.5) & 6,983 (37.9) \\
 & Geo.\ center & 15,525 (84.2) & 6,976 (37.8) \\
 & LayerNorm & 3,188 (17.3) & 1,422 (7.7) \\
 & RMS & 3,432 (18.6) & 1,567 (8.5) \\
\midrule
TopK & None & 14,206 (77.1) & 12,411 (67.3) \\
 & Mean center & 18,298 (99.3) & 18,189 (98.7) \\
 & Geo.\ center & 18,290 (99.2) & 18,173 (98.6) \\
 & LayerNorm & 14,660 (79.5) & 14,081 (76.4) \\
 & RMS & 10,041 (54.5) & 10,317 (56.0) \\
\midrule
JumpReLU & None & 2,255 (12.2) & 645 (3.5) \\
 & Mean center & 4,397 (23.9) & 1,726 (9.4) \\
 & Geo.\ center & 4,082 (22.1) & 1,689 (9.2) \\
 & LayerNorm & 1,765 (9.6) & 373 (2.0) \\
 & RMS & 1,995 (10.8) & 585 (3.2) \\
\bottomrule
\end{tabular}
\end{table}

\section{Fine-Grained Steering Benchmark}
\label{app:steering-benchmark}

This section provides additional details for the fine-grained steering
benchmark in Section~\ref{sec:steering-benchmark}. We describe the benchmark,
feature retrieval and intervention protocol, and evaluation procedure, followed
by analyses of retrieval strategy and student-to-teacher transfer.

\subsection{Benchmark and Evaluation Protocol}
\label{app:steering-protocol}
 
\textbf{Benchmark construction.} The benchmark contains 100 target concepts across five categories: color, material, texture or pattern, local appearance, and local state or geometry. Each concept is evaluated in five under-specified and five explicit-conflict contexts, yielding 1,000 cases. Under-specified contexts leave the target attribute unspecified in the source prompt, while explicit-conflict contexts specify a competing attribute. Each case provides a source prompt $c$, a target query $q$ describing the requested regional appearance, a reference query $q_{\mathrm{base}}$, and a target region. A full target prompt $c'$ is provided for the oracle references. Table~\ref{tab:fgs-taxonomy} summarizes the benchmark, and Table~\ref{tab:fgs-rose-case} presents a matched pair of contexts.
 
\begin{table}[t]
    \centering
    \caption{\textbf{Fine-grained steering benchmark.} Each target concept has five under-specified and five explicit-conflict contexts.}
    \label{tab:fgs-taxonomy}
    \small
    \begin{tabularx}{\linewidth}{@{}XrrX@{}}
        \toprule
        Category & Targets & Cases & Example change \\
        \midrule
        Color & 20 & 200 & Green dress $\rightarrow$ red dress \\
        Material & 20 & 200 & Wooden chair $\rightarrow$ metal chair \\
        Texture or pattern & 20 & 200 & Plain shirt $\rightarrow$ striped shirt \\
        Local appearance & 20 & 200 & Straight hair $\rightarrow$ curly hair \\
        Local state or geometry & 20 & 200 & Closed umbrella $\rightarrow$ open umbrella \\
        \midrule
        Total & 100 & 1,000 & \\
        \bottomrule
    \end{tabularx}
\end{table}
 
\begin{table*}[t]
    \centering
    \caption{\textbf{Matched rose-blooming cases.} The scene and target query are shared, while the source state and reference query differ.}
    \label{tab:fgs-rose-case}
    \small
    \setlength{\tabcolsep}{6pt}
    \renewcommand{\arraystretch}{1.18}
    \begin{tabularx}{\textwidth}{@{}p{0.18\textwidth}>{\raggedright\arraybackslash}X>{\raggedright\arraybackslash}X@{}}
        \toprule
        \multicolumn{3}{@{}l@{}}{\textbf{Category:} Local state or geometry \qquad \textbf{Target region:} rose} \\
        \multicolumn{3}{@{}l@{}}{\textbf{Steering instruction:} make the rose bloom} \\
        \multicolumn{3}{@{}l@{}}{\textbf{Target query $q$:} rose in full bloom} \\
        \midrule
        & \textbf{Under-specified} & \textbf{Explicit-conflict} \\
        \midrule
        \addlinespace[3pt]
        Source prompt $c$
        & a single rose in a garden with morning dew
        & a single \textbf{closed rose bud} in a garden with morning dew \\
        \addlinespace[3pt]
        Source state
        & Blooming state is unspecified.
        & A closed bud is explicitly specified. \\
        \addlinespace[3pt]
        Reference query $q_{\mathrm{base}}$
        & rose
        & rose bud \\
        \midrule
        \multicolumn{3}{@{}p{\textwidth}@{}}{\textbf{Shared target prompt $c'$:} a single rose \textbf{in full bloom} in a garden with morning dew} \\
        \bottomrule
    \end{tabularx}
\end{table*}
 
\textbf{Generators and dictionaries.} Steering is evaluated with SANA-Sprint, Nitro-1-PixArt, and the SANA teacher at $1024\times1024$ resolution. Interventions are applied to the post-block residual stream at zero-indexed layer 14. For each student generator, the evaluation includes the 15 dictionaries at this layer, spanning three SAE families and five input settings (Appendix~\ref{app:sae-training-configurations}), for 30 student-trained dictionaries in total. The SANA teacher reuses the corresponding SANA-Sprint dictionaries and visual indices.
 
\textbf{Feature retrieval.} We use the direct and contrastive retrieval rules defined in Section~\ref{sec:feature_retrieval}. The reference query describes the generic target object in under-specified contexts and the competing source attribute in explicit-conflict contexts. Retrieval uses only the text queries and the precomputed visual index, without access to target images or steering outcomes, and ties are resolved by the smaller feature index. For a given dictionary, case, and retrieval strategy, the selected feature is fixed across steering strengths and generation seeds.
 
\textbf{Target region.} SAM~3 localizes the target object in the unmodified source image $I^{(0)}$. Valid instance masks are combined into the target region $M$ and mapped to the visual-token grid to obtain $M_p\in\{0,1\}$. The target region remains fixed across methods and steering strengths within each case and generation seed.

\textbf{Direction adjustment.}
Let $\mathbf d_i\in\mathbb R^D$ denote the decoder atom of feature $i$,
where $D$ is the number of activation channels and
$(\mathbf d_i)_k$ is its scalar value in channel $k$.
The channel mean is the scalar
\begin{equation*}
    \bar d_i
    =
    \frac{1}{D}\sum_{k=1}^{D}(\mathbf d_i)_k.
\end{equation*}
This average is taken over the channels of one decoder atom.

For the five input settings considered here, the direction adjustment is
\begin{equation*}
    \mathcal A_{\mathcal T}(\mathbf d_i)
    =
    \begin{cases}
        \mathbf d_i-\bar d_i\mathbf 1,
        & \text{if the input setting is LayerNorm},\\[2pt]
        \mathbf d_i,
        & \text{otherwise},
    \end{cases}
\end{equation*}
where $\mathbf 1\in\mathbb R^D$ is the all-ones vector.
Thus, None, Mean center, Geo. center, and RMS leave the decoder atom
unchanged, whereas LayerNorm subtracts the same channel mean from
every component of the atom.

RMS rescaling contributes only a positive scalar factor, which cancels
under unit normalization.
Decoder bias terms are excluded because they cancel in differences
between reconstructed activations.
The LayerNorm adjustment removes the channel-mean component and does
not invert the input normalization.
The selected feature's unit steering direction is
\begin{equation*}
    \hat{\mathbf u}_{i^\star}
    =
    \nu\!\left(
        \mathcal A_{\mathcal T}(\mathbf d_{i^\star})
    \right).
\end{equation*}

\textbf{Masked intervention.} The masked update of Section~\ref{sec:single_feature_steering} is applied at every denoising step with constant strength. The reference scale $R(t)$ is the root mean square of raw residual $\ell_2$ norms over all spatial tokens in 200 fixed, unmodified source generations at the intervened layer and step. Calibration is performed separately for each generation setting. The steering strength is swept over
\begin{equation*}
\mathcal{G}
=
\{0.05,0.10,0.15,0.20,0.30,0.40,0.50,0.75,1.00,1.50\}.
\end{equation*}
 
\textbf{Reference methods.} \emph{Source} denotes the unmodified generation $I^{(0)}$. \emph{Oracle Dense} constructs a target-informed direction from paired source- and target-prompt generations with the same initial noise. The direction is obtained by subtracting the mean activation over the target region in the source generation from the corresponding mean in the target generation and normalizing the difference. It is applied using the same masked intervention protocol and strength grid as the SAE directions. \emph{Oracle Prompt} generates directly from $c'$ using the same initial noise and sampling settings. The oracle references access the full target prompt, and Oracle Dense additionally accesses target-generation activations.
 
\textbf{Regional target alignment.}
We measure target control using Region SigLIP $\Delta$.
Let $C_M(I)$ denote a crop derived from the bounding box of the target
region $M$. The bounding box is expanded by 15\% of its width and height
on each side and clipped to the image boundaries, with identical crop
coordinates used for the steered image $I$ and the unmodified source
image $I^{(0)}$.
Let $S(C_M(I),q)$ denote the cosine similarity between the SigLIP~2
embeddings of the crop and the target query $q$.
Regional target gain is
\begin{equation*}
\Delta_{\mathrm{target}}(I)
=
S(C_M(I),q)
-
S(C_M(I^{(0)}),q).
\end{equation*}
We refer to $\Delta_{\mathrm{target}}$ as Region SigLIP $\Delta$ in the
main text. Positive values indicate increased regional alignment with
the target query relative to the unmodified source image.

\textbf{Outside-region preservation.}
We measure outside-region preservation using LPIPS-out.
The target region $M$ is dilated by 12 pixels, and LPIPS-out averages
the spatial AlexNet LPIPS~\citep{zhang2018unreasonable} distance map
between $I$ and $I^{(0)}$ over the complement of the dilated region.
Lower values indicate better preservation of content outside the target
region.
 
\textbf{Strength selection.} For each case and generation seed, strength is selected separately for every fixed combination of SAE dictionary and retrieval strategy:
\begin{equation*}
\rho^\star
\in
\operatorname*{arg\,max}_{\rho\in\mathcal{G}}
\Delta_{\mathrm{target}}(I(\rho)).
\end{equation*}
Ties are resolved by lower LPIPS-out and then smaller $\rho$. Both metrics are reported from the same selected output $I(\rho^\star)$. Oracle Dense uses the same strength-selection procedure, whereas Oracle Prompt has no strength sweep. SigLIP~2 is used for feature retrieval, regional target scoring, and strength selection. The resulting comparisons describe per-case best-of-sweep performance.
 
\textbf{Statistical aggregation.} Evaluation uses three generation seeds. Within each generation setting and seed, all methods are compared on their common intersection of valid cases. Metrics are first averaged over cases, then summarized by the mean and sample standard deviation of the three seed-level averages. Reported target gains and LPIPS-out values are multiplied by $10^3$.
 
\subsection{Additional Analysis}
\label{app:steering-analysis}

\textbf{Direct versus Contrastive retrieval.}
Across all three generation environments, Contrastive SAE retrieval increases
mean Region SigLIP improvement relative to Direct SAE for all 15 evaluated
dictionaries. Its effect on source preservation is less uniform: LPIPS-out
decreases for 8 of 15 SANA-Sprint dictionaries and 14 of 15 Nitro
dictionaries, but increases for all 15 SANA-teacher dictionaries. Thus,
improved target control does not necessarily imply improved preservation.

\begin{figure*}[t]
\centering
\includegraphics[width=\textwidth]{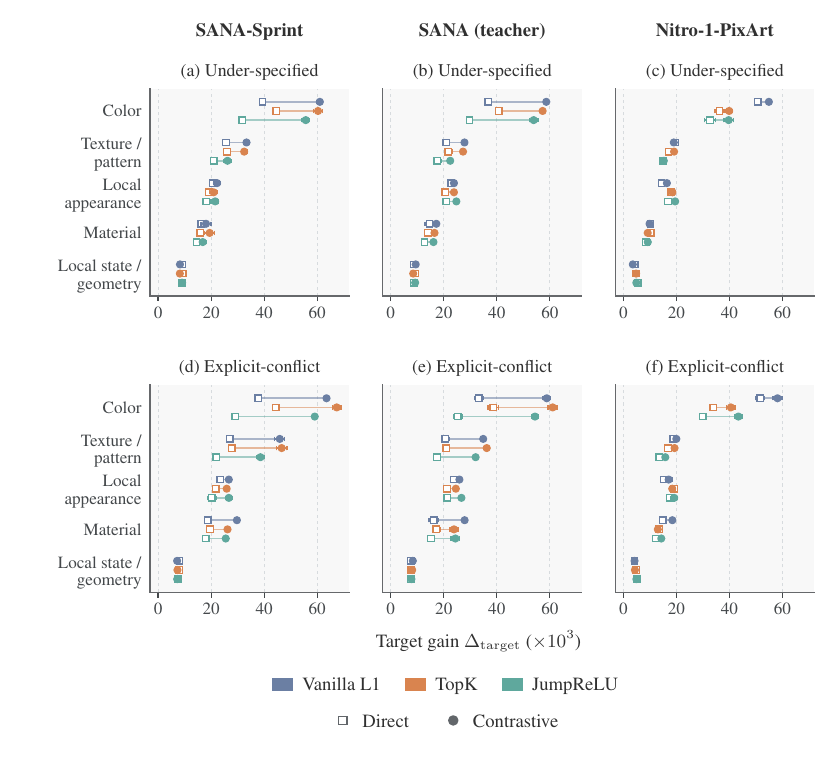}
\caption{
\textbf{Target alignment by concept category and source context.}
Columns correspond to the three generation environments:
SANA-Sprint (student), SANA (teacher), and Nitro-1-PixArt;
the top and bottom rows show under-specified and explicit-conflict source
contexts, respectively.
Colors denote SAE families (L1, TopK, and JumpReLU), while open squares
and filled circles denote Direct and Contrastive direction retrieval.
Lines connect the two direction sources within the same SAE family and concept
category.
For each point, we first average over the five input configurations within each
generation seed, then show the mean with sample SD across three seeds.
Target-alignment improvement $\Delta_{\mathrm{target}}$ is reported in units of $10^{-3}$.
}
\label{fig:fgs-topics}
\end{figure*}

\textbf{Concept category and source context.}
Figure~\ref{fig:fgs-topics} resolves target-control performance by semantic
category, source-context regime, SAE family, and generation environment.
Within each SAE family, we first average the five input configurations within
each generation seed, and then report the mean and sample SD across three
seeds. Color and texture or pattern are comparatively easy to steer, whereas
local state or geometry is more difficult using a single SAE direction.
Contrastive SAE retrieval generally improves target alignment over Direct SAE
retrieval, with particularly pronounced gains for explicit-conflict cases in
the SANA environments.

\textbf{Student-to-teacher transfer.}
The SANA teacher uses the same student-trained dictionaries, feature indices,
and retrieved feature identities as SANA-Sprint. The positive steering results
therefore show that student-trained SAE directions remain actionable in the
longer teacher generation trajectory, although the control--preservation
operating point changes. For example, mean-centered TopK with Contrastive SAE
achieves Region SigLIP improvements of \(31.7\) and \(28.8\) on the SANA
student and teacher, respectively, while LPIPS-out changes from \(157.3\) to
\(234.5\).

\section{Additional Applications of Learned SAE Features}
\label{app:broader-uses}

Beyond steering toward a target concept, individual SAE features can serve as lightweight interfaces for both controlling and reading out model behavior. 
We study these complementary roles through style suppression and single-feature semantic readout. 
In both cases, the pretrained generators and SAEs remain frozen.

\begin{figure*}[ht]
    \centering

    \begin{subfigure}[t]{0.215\textwidth}
        \includegraphics[width=\linewidth]{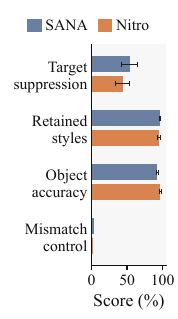}
        \caption{}
        \label{fig:broader-uses-a}
    \end{subfigure}\hfill
    \begin{subfigure}[t]{0.205\textwidth}
        \includegraphics[width=\linewidth]{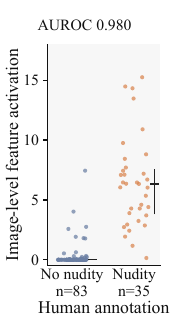}
        \caption{}
        \label{fig:broader-uses-b}
    \end{subfigure}\hfill
    \begin{subfigure}[t]{0.235\textwidth}
        \includegraphics[width=\linewidth]{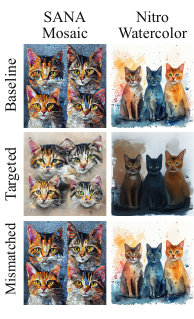}
        \caption{}
        \label{fig:broader-uses-c}
    \end{subfigure}\hfill
    \begin{subfigure}[t]{0.295\textwidth}
        \includegraphics[width=\linewidth]{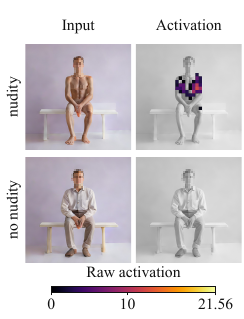}
        \caption{}
        \label{fig:broader-uses-d}
    \end{subfigure}

    \caption{
    \textbf{Learned SAE features as intervention and readout interfaces.}
    \textbf{(a)} Suppressing a style-associated feature reduces the targeted
    style while retaining non-target styles and object content; mismatched
    interventions have little effect.
    \textbf{(b)} Image-level activation of a nudity-associated feature on
    held-out annotated examples, yielding an AUROC of $0.980$.
    \textbf{(c)} Qualitative examples of targeted and mismatched style
    suppression for SANA and Nitro.
    \textbf{(d)} Spatial activation maps of the selected feature for examples
    with and without nudity.
    }
    \label{fig:broader-uses}
\end{figure*}

\textbf{Style suppression.}
Following SAeUron~\citep{cywinski2025saeuron}, we intervene on a single style-associated feature during generation in SANA-Sprint and Nitro-1-PixArt; unlike the steering benchmark, this intervention rescales the selected SAE code rather than adding a unit-normalized direction (Appendix~\ref{app:style-suppression-protocol}).
Because the released UnlearnCanvas~\citep{zhang2024unlearncanvas} classifier transfers poorly to these generators, we train generator-specific evaluators and restrict evaluation to ten styles that are classified reliably across validation and external baseline images (Tables~\ref{tab:app-style-evaluator} and~\ref{tab:app-style-selection}).
Using matched prompts, we compare targeted suppression with unmodified generations and mismatched-feature interventions (Figure~\ref{fig:broader-uses}a, c). 
Under a preservation constraint on non-target styles and object content (Fig.~\ref{fig:app-style-configurations}; Table~\ref{tab:app-style-all}), the selected configurations achieve target-style suppression scores of 53.8\% for SANA and 44.0\% for Nitro, compared with 3.5\% and 1.2\% under mismatched interventions.
These results show that existing dictionary features can selectively attenuate a visual style, although suppression remains incomplete and varies substantially across styles (Table~\ref{tab:app-style-per-style}; Fig.~\ref{fig:style-failures}).

\textbf{Nudity feature readout.}
We next ask whether a single frozen SAE feature can also provide a useful semantic readout. 
Using a labeled discovery set (Table~\ref{tab:app-readout-split}), we select one feature from a SANA-Sprint SAE according to its association with human-annotated nudity (Algorithm~\ref{alg:nudity-feature-discovery}), then evaluate that feature on images from disjoint scenes and template families without fitting an additional probe.
We aggregate its top-$K$ token-wise responses into an image-level activation score; the selected feature reaches an AUROC of 0.980 on 118 held-out images (Figure~\ref{fig:broader-uses}b) and is recovered consistently across pooling sizes (Fig.~\ref{fig:app-nudity-feature-analysis}b), while spatial activation maps visualize where the feature responds (Figure~\ref{fig:broader-uses}d; Fig.~\ref{fig:app-nudity-matched-examples} and~\ref{fig:app-nudity-hard-negative-examples}).
This provides
evidence that a single SAE feature can expose a semantically useful signal
in addition to supporting intervention. 
These results also highlight the potential of learned SAE features for safety-relevant monitoring and content detection.

\section{Style Suppression with General-Purpose SAEs}
\label{app:style-suppression}

This section provides the experimental details for the style-suppression
study in Section~\ref{app:broader-uses}. We describe the adapted style
evaluators, the single-feature intervention protocol, quantitative results
across SAE configurations, and additional qualitative examples.

\subsection{Style Evaluator Adaptation}

The released UnlearnCanvas~\citep{zhang2024unlearncanvas} style classifier
transfers poorly to generations from SANA-Sprint and Nitro-1-PixArt,
achieving only 9.9\% and 9.5\% top-1 accuracy on our external evaluation
sets. We therefore train a generator-specific ViT-L/16 style evaluator for
each model. Each evaluator is trained on 50 UnlearnCanvas styles together
with a plain-image class, using prompt-disjoint training and validation
splits. Table~\ref{tab:app-style-evaluator} summarizes evaluator performance.

\begin{table}[t]
\centering
\caption{
\textbf{Style-evaluator accuracy (\%).}
Validation contains 2,040 images per generator and the external screen
contains 1,224. External top-1 results compare the released classifier,
the evaluator trained on the other generator, and the matched evaluator.
}
\label{tab:app-style-evaluator}
\small
\setlength{\tabcolsep}{5pt}
\begin{tabular}{@{}lrrrrrr@{}}
\toprule
& \multicolumn{2}{c}{Validation}
& \multicolumn{3}{c}{External Top-1}
& External \\
\cmidrule(lr){2-3}\cmidrule(lr){4-6}
Image domain
& Top-1 & Top-5
& Released & Other & Matched
& Top-5 \\
\midrule
SANA-Sprint    & 60.2 & 83.3 & 9.9 & 51.1 & 77.5 & 95.1 \\
Nitro-1-PixArt & 51.7 & 74.8 & 9.5 & 52.6 & 66.7 & 88.4 \\
\bottomrule
\end{tabular}
\end{table}

To avoid evaluating suppression on styles that are poorly recognized even
before intervention, we retain only styles with at least 80\% per-class
accuracy on both validation and external images for both generators. This
yields ten shared styles: Blossom Season, Comic Etch, Mosaic, Neon Lines,
Pencil Drawing, Pop Art, Red Blue Ink, Ukiyoe, Van Gogh, and Watercolor.
The set is determined from unmodified generations before inspecting SAE
interventions; all ten subsequently achieve at least 90\% recognition on
the downstream baselines.

\begin{table}[t]
\centering
\caption{
\textbf{Recognition accuracy (\%) for the ten retained styles.}
Style inclusion requires at least 80\% accuracy on both validation and
external images for both generators. Baseline denotes recognition on the
unmodified generations used in the suppression experiment.
}
\label{tab:app-style-selection}
\small
\setlength{\tabcolsep}{4pt}
\begin{tabular}{@{}lrrrrrr@{}}
\toprule
& \multicolumn{3}{c}{SANA-Sprint}
& \multicolumn{3}{c}{Nitro-1-PixArt} \\
\cmidrule(lr){2-4}\cmidrule(lr){5-7}
Style & Val. & Ext. & Base. & Val. & Ext. & Base. \\
\midrule
Blossom Season & 87.5 & 91.7 & 90.0 & 85.0 & 87.5 & 90.0 \\
Comic Etch & 85.0 & 100.0 & 90.0 & 92.5 & 100.0 & 100.0 \\
Mosaic & 90.0 & 91.7 & 92.5 & 82.5 & 100.0 & 100.0 \\
Neon Lines & 95.0 & 100.0 & 100.0 & 95.0 & 100.0 & 100.0 \\
Pencil Drawing & 90.0 & 100.0 & 100.0 & 97.5 & 100.0 & 100.0 \\
Pop Art & 80.0 & 95.8 & 97.5 & 85.0 & 95.8 & 97.5 \\
Red Blue Ink & 97.5 & 100.0 & 100.0 & 87.5 & 100.0 & 100.0 \\
Ukiyoe & 85.0 & 100.0 & 95.0 & 92.5 & 100.0 & 100.0 \\
Van Gogh & 87.5 & 100.0 & 100.0 & 82.5 & 100.0 & 100.0 \\
Watercolor & 92.5 & 100.0 & 100.0 & 95.0 & 95.8 & 100.0 \\
\bottomrule
\end{tabular}
\end{table}

\subsection{Single-Feature Suppression Protocol}
\label{app:style-suppression-protocol}

We evaluate all 30 layer-14 dictionaries, comprising three SAE families and five normalization and initialization settings for each generator. Both generators and SAEs remain frozen. Each target style is evaluated across eight object contexts and five generation seeds. Baseline, targeted, and mismatched conditions share the same prompt, initial noise, and sampling settings. Table~\ref{tab:app-style-protocol} summarizes the evaluation setup.

\begin{table}[t]
    \centering
    \caption{\textbf{Style-suppression evaluation setup.}}
    \label{tab:app-style-protocol}
    \small
    \setlength{\tabcolsep}{5pt}
    \begin{tabularx}{\linewidth}{@{}lX@{}}
        \toprule
        Component & Setting \\
        \midrule
        SAE site & Layer-14 post-block residual, 18,432 features \\
        SAE families & L1, TopK, JumpReLU \\
        Input settings & None, mean centering, geometric-median centering, LayerNorm, RMS scaling \\
        Target styles & 10 shared styles \\
        Object contexts & Architectures, Birds, Cats, Dogs, Flowers, Horses, Human, Trees \\
        Generation seeds & Five \\
        SANA-Sprint & Two inference steps, embedded guidance 4.5 \\
        Nitro-1-PixArt & One inference step, guidance 0, timestep 400 \\
        Conditions & Baseline, targeted feature, mismatched feature \\
        \bottomrule
    \end{tabularx}
\end{table}

\textbf{Feature selection.} We adapt the activation-based feature selection of SAeUron~\citep{cywinski2025saeuron} to select one dictionary feature for each target style. Let $\mathcal{S}$ denote the set of ten styles.
For each style $\omega\in\mathcal{S}$, let $\bar{z}_{\omega,i}$ denote the mean activation of feature $i$ over all visual tokens from the first transformer forward pass of eight object-anchor generations, including zero activations. The corresponding mean over the other styles is
\begin{equation*}
\bar{z}_{\neg\omega,i}
=
\frac{1}{|\mathcal{S}|-1}
\sum_{\substack{\omega'\in\mathcal{S}\\\omega'\neq\omega}}
\bar{z}_{\omega',i}.
\end{equation*}
The selected feature maximizes the difference between its normalized activation shares for the target and remaining styles:
\begin{equation*}
i_\omega
=
\arg\max_{1\leq i\leq m}
\left[
\frac{\bar{z}_{\omega,i}}
{\sum_{j=1}^{m}\bar{z}_{\omega,j}+10^{-8}}
-
\frac{\bar{z}_{\neg\omega,i}}
{\sum_{j=1}^{m}\bar{z}_{\neg\omega,j}+10^{-8}}
\right].
\end{equation*}
Ties are resolved by the smaller feature index. Feature selection is performed independently for each dictionary.

\textbf{Activation-gated intervention.}
The anchor statistics determine a threshold and scaling coefficient for the selected feature:
\begin{equation*}
\kappa_\omega
=
\frac{1}{|\mathcal{S}|}
\sum_{\omega'\in\mathcal{S}}
\bar{z}_{\omega',i_\omega},
\qquad
\beta_\omega=-\bar{z}_{\omega,i_\omega}.
\end{equation*}
For each current token, the SAE encodes the processed residual activation $\mathbf{x}=\mathcal{T}(\mathbf{h})$ into $\mathbf{z}$. Only the selected coordinate is modified:
\begin{equation*}
z'_{i_\omega}
=
\begin{cases}
\beta_\omega z_{i_\omega}, & z_{i_\omega}>\kappa_\omega,\\
z_{i_\omega}, & \text{otherwise},
\end{cases}
\qquad
z'_j=z_j \quad (j\neq i_\omega).
\end{equation*}
The threshold determines whether the intervention is applied, and the scaling acts on the entire selected activation. The feature index, threshold, and scaling coefficient remain fixed throughout generation, while current residual activations are re-encoded at every denoising step. The mismatched control uses the complete intervention rule of another style, including its feature index, threshold, and scale, selected through a fixed cyclic mapping.

The edited code contributes a decoded difference in SAE input space:
\begin{equation*}
\delta\mathbf{x}
=
\mathbf{W}_{\mathrm{dec}}(\mathbf{z}'-\mathbf{z}),
\qquad
\mathbf{x}'=\mathbf{x}+\delta\mathbf{x}.
\end{equation*}
Because $\hat{\mathbf{x}}=\mathbf{W}_{\mathrm{dec}}\mathbf{z}+\mathbf{b}_{\mathrm{dec}}$, the edited input can equivalently be written as
\begin{equation*}
\mathbf{x}'
=
\mathbf{W}_{\mathrm{dec}}\mathbf{z}'
+
\mathbf{b}_{\mathrm{dec}}
+
(\mathbf{x}-\hat{\mathbf{x}}).
\end{equation*}
The original SAE reconstruction residual is therefore retained in the edited input.

For None and the two centering settings, the raw residual update is $\mathbf{h}'=\mathbf{h}+\delta\mathbf{x}$. 
RMS scaling restores the fitted scale $s$ defined in Appendix~\ref{app:sae-training-configurations}, giving $\mathbf{h}'=\mathbf{h}+s\,\delta\mathbf{x}$.
For LayerNorm, the edited SAE input is recentered and normalized before restoring the original token mean and scale. Reconstruction-residual preservation applies to the SAE input-space edit, before this nonlinear restoration. The intervention retains the magnitude of the decoded difference and does not apply the unit-direction normalization or $\rho R(t)$ scaling used in the steering benchmark.

\textbf{Evaluation.} Each target-style intervention is evaluated on 40 target-style generations, 360 generations from the other nine styles, and 40 plain-object generations. 
We adapt the UA, IRA, and CRA metrics from UnlearnCanvas~\citep{zhang2024unlearncanvas} to our style-suppression setting.
Target suppression (UA) is one minus target-style classification accuracy. Retained-style accuracy (IRA) measures classification accuracy on the other nine styles under the same intervention. Object accuracy (CRA) measures object classification accuracy on plain-object prompts. UA and IRA use the corresponding generator-specific style classifier, while CRA uses the object classifier. All three metrics are averaged equally over the ten target styles, with higher values indicating better performance. CRA assesses preservation of object category rather than exact instance identity or composition.

\subsection{Quantitative Results}

\begin{figure*}[t]
    \centering
    \includegraphics{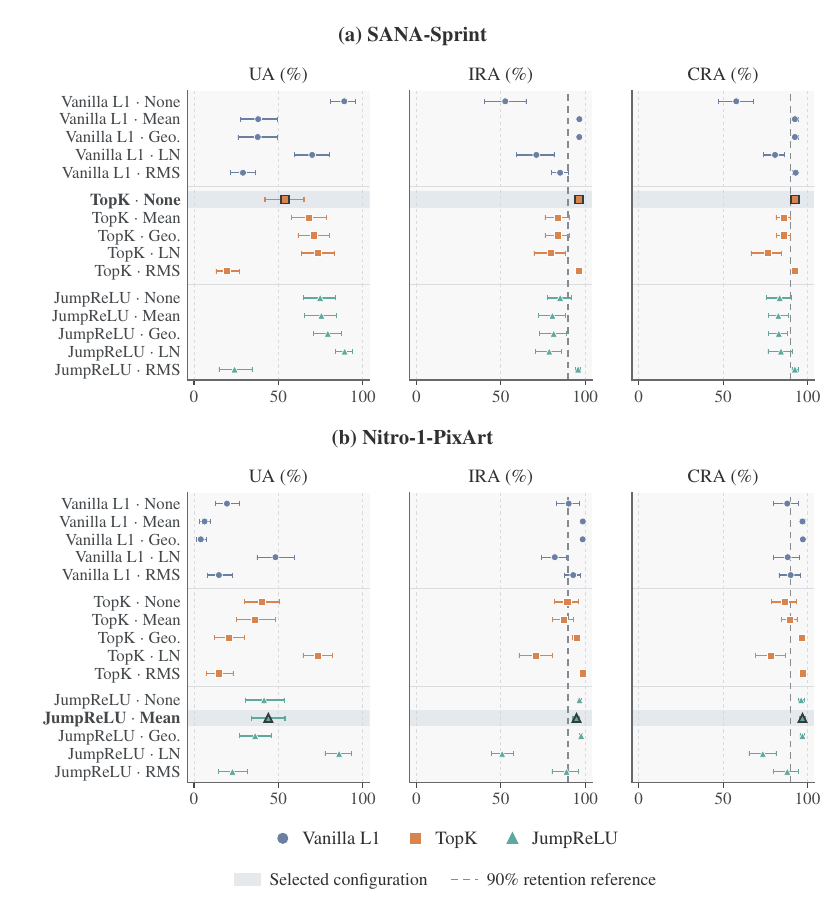}
    \caption{
    \textbf{Style-suppression trade-offs across SAE configurations.}
    Point estimates show target-style suppression (UA), retained-style
    accuracy (IRA), and plain-prompt object accuracy (CRA) for all
    layer-14 dictionaries in SANA-Sprint and Nitro-1-PixArt; error bars
    denote 95\% bootstrap intervals.
    Shaded rows indicate the configurations selected by maximizing UA
    subject to mean IRA and CRA of at least 90\%; dashed lines mark the
    90\% retention reference.
    }
    \label{fig:app-style-configurations}
\end{figure*}

For each generator, we select the configuration with the highest mean UA
among those with both mean IRA and CRA of at least 90\%.
Figure~\ref{fig:app-style-configurations} summarizes the resulting
suppression-retention trade-off across all 30 SAE configurations.
This selects unnormalized TopK for SANA-Sprint and mean-centered JumpReLU
for Nitro-1-PixArt.

Table~\ref{tab:app-style-all} reports the corresponding exact values.
Higher suppression is achievable for several configurations, but typically
at substantial cost to non-target preservation.

\begin{table*}[t]
\centering
\caption{Complete style-suppression results (\%). Each row averages ten target styles, five generation seeds, and eight objects. UA, IRA, and CRA are measured after targeted intervention; control UA uses the cyclic mismatched feature. Bold rows maximize UA subject to mean IRA and CRA of at least 90\% within each model. This choice uses the reported evaluation data. Intervals for all targeted metrics are shown in Figure~\ref{fig:app-style-configurations}. Mean and Geo.\ median denote the two bias-initialization conventions.}
\label{tab:app-style-all}
\small
\setlength{\tabcolsep}{7pt}
\begin{tabular}{llrrrr}
\toprule
SAE family & Input convention & UA $\uparrow$ & IRA $\uparrow$ & CRA $\uparrow$ & Control UA \\
\midrule
\multicolumn{6}{l}{\textbf{SANA-Sprint}} \\
L1 & None & 89.0 & 52.6 & 57.8 & 49.5 \\
L1 & Mean & 38.0 & 96.6 & 92.5 & 3.5 \\
L1 & Geo. median & 37.8 & 96.5 & 92.5 & 3.5 \\
L1 & LayerNorm & 70.0 & 71.1 & 80.7 & 26.7 \\
L1 & RMS & 29.0 & 85.2 & 93.0 & 15.7 \\
\textbf{TopK} & \textbf{None} & \textbf{53.8} & \textbf{96.4} & \textbf{92.5} & \textbf{3.5} \\
TopK & Mean & 68.0 & 83.6 & 85.8 & 15.0 \\
TopK & Geo. median & 71.0 & 83.6 & 86.0 & 14.5 \\
TopK & LayerNorm & 73.5 & 79.5 & 76.2 & 22.5 \\
TopK & RMS & 19.5 & 96.5 & 92.5 & 3.5 \\
JumpReLU & None & 74.8 & 85.2 & 83.5 & 13.5 \\
JumpReLU & Mean & 75.5 & 80.6 & 82.8 & 17.2 \\
JumpReLU & Geo. median & 79.2 & 81.4 & 83.0 & 19.0 \\
JumpReLU & LayerNorm & 89.2 & 78.7 & 84.2 & 21.5 \\
JumpReLU & RMS & 24.0 & 95.8 & 92.5 & 3.2 \\
\midrule
\multicolumn{6}{l}{\textbf{Nitro-1-PixArt}} \\
L1 & None & 19.5 & 90.3 & 88.0 & 11.2 \\
L1 & Mean & 6.2 & 98.6 & 97.0 & 1.5 \\
L1 & Geo. median & 4.0 & 98.5 & 97.2 & 1.5 \\
L1 & LayerNorm & 48.3 & 82.0 & 88.2 & 11.5 \\
L1 & RMS & 14.8 & 92.9 & 90.0 & 11.5 \\
TopK & None & 40.2 & 89.4 & 86.8 & 10.2 \\
TopK & Mean & 36.0 & 87.3 & 89.5 & 11.7 \\
TopK & Geo. median & 20.5 & 94.9 & 96.5 & 2.7 \\
TopK & LayerNorm & 73.5 & 70.9 & 78.5 & 23.5 \\
TopK & RMS & 14.5 & 98.6 & 97.0 & 1.2 \\
JumpReLU & None & 41.5 & 96.6 & 96.2 & 1.7 \\
\textbf{JumpReLU} & \textbf{Mean} & \textbf{44.0} & \textbf{94.9} & \textbf{97.0} & \textbf{1.2} \\
JumpReLU & Geo. median & 36.2 & 97.7 & 97.0 & 1.2 \\
JumpReLU & LayerNorm & 86.0 & 50.8 & 73.5 & 50.7 \\
JumpReLU & RMS & 22.7 & 88.9 & 88.0 & 6.0 \\
\bottomrule
\end{tabular}
\end{table*}

Suppression varies substantially across styles and generators.
Table~\ref{tab:app-style-per-style} therefore reports the ten target styles
for the two representative configurations. For example, SANA suppresses
Mosaic strongly but has little effect on Neon Lines, whereas Nitro exhibits
the opposite pattern. The aggregate preservation constraint likewise does
not guarantee uniformly high retention for every individual target.

\begin{table*}[!t]
\centering
\caption{Per-style results for the representative dictionaries (\%). SANA uses TopK without input normalization; Nitro uses mean-centered JumpReLU. Each target evaluates 40 target-style, 360 retained-style, and 40 plain-object images per intervention condition. Control columns report mismatched-feature UA. Mean retention above 90\% does not imply that every target style meets this constraint.}
\label{tab:app-style-per-style}
\small
\setlength{\tabcolsep}{5pt}
\begin{tabular}{lrrrrrrrr}
\toprule
 & \multicolumn{4}{c}{SANA-Sprint} & \multicolumn{4}{c}{Nitro-1-PixArt} \\
\cmidrule(lr){2-5}\cmidrule(lr){6-9}
Target style & UA & IRA & CRA & Control & UA & IRA & CRA & Control \\
\midrule
Blossom Season & 70.0 & 96.9 & 92.5 & 10.0 & 10.0 & 99.7 & 97.5 & 10.0 \\
Comic Etch & 12.5 & 97.2 & 92.5 & 10.0 & 22.5 & 98.6 & 97.5 & 0.0 \\
Mosaic & 100.0 & 96.9 & 92.5 & 7.5 & 32.5 & 95.8 & 97.5 & 0.0 \\
Neon Lines & 2.5 & 95.0 & 92.5 & 0.0 & 90.0 & 96.1 & 97.5 & 0.0 \\
Pencil Drawing & 100.0 & 96.4 & 92.5 & 0.0 & 87.5 & 77.8 & 100.0 & 0.0 \\
Pop Art & 5.0 & 96.4 & 92.5 & 2.5 & 92.5 & 92.8 & 95.0 & 2.5 \\
Red Blue Ink & 5.0 & 96.1 & 92.5 & 0.0 & 0.0 & 98.6 & 95.0 & 0.0 \\
Ukiyoe & 75.0 & 96.7 & 92.5 & 5.0 & 2.5 & 98.6 & 97.5 & 0.0 \\
Van Gogh & 100.0 & 96.1 & 92.5 & 0.0 & 52.5 & 91.9 & 95.0 & 0.0 \\
Watercolor & 67.5 & 96.1 & 92.5 & 0.0 & 50.0 & 98.6 & 97.5 & 0.0 \\
\bottomrule
\end{tabular}
\end{table*}

\newlength{\AppStyleLabelWidth}
\newlength{\AppStyleImageWidth}
\newlength{\AppStyleImageGap}

\newcommand{\AppStyleSetup}{%
  \setlength{\AppStyleLabelWidth}{0.18\linewidth}%
  \setlength{\AppStyleImageWidth}{0.26\linewidth}%
  \setlength{\AppStyleImageGap}{0.02\linewidth}%
  \setlength{\parindent}{0pt}%
  \setlength{\parskip}{0pt}%
}

\newcommand{\AppStyleHeader}{%
  \noindent\hspace*{\AppStyleLabelWidth}%
  \makebox[\AppStyleImageWidth]{%
    \fontsize{8.5}{10}\selectfont Baseline}%
  \hspace{\AppStyleImageGap}%
  \makebox[\AppStyleImageWidth]{%
    \fontsize{8.5}{10}\selectfont Targeted}%
  \hspace{\AppStyleImageGap}%
  \makebox[\AppStyleImageWidth]{%
    \fontsize{8.5}{10}\selectfont Mismatched}%
  \par\vspace{1.5mm}%
}

\begingroup

\makeatletter
\setlength{\@fptop}{0pt}
\makeatother

\subsection{Qualitative Examples}
\label{app:style-qualitative}

Figures~\ref{fig:style-success-1} and~\ref{fig:style-success-2}
show selected visual success cases, and
Figure~\ref{fig:style-failures} shows examples of incomplete
suppression. All examples use Cats, with identical
prompts and initial noise within each triplet. The SAE configuration
varies across rows, as indicated. This grouping reflects qualitative
assessment of the visible output and does not change the
classifier-based metrics above. In particular, a non-target style
prediction can coexist with residual target-style attributes.

\begin{figure*}[!htbp]
\centering
\begingroup
\AppStyleSetup
\AppStyleHeader

  \noindent
  \begin{minipage}[c]{\AppStyleLabelWidth}
    \raggedright
    \fontsize{8.5}{10}\selectfont
    SANA-Sprint\\Mosaic\\[2pt]
    {\fontsize{7.5}{9}\selectfont None\\TopK}
  \end{minipage}%
  \begin{minipage}[c]{\AppStyleImageWidth}
    \includegraphics[width=\linewidth]{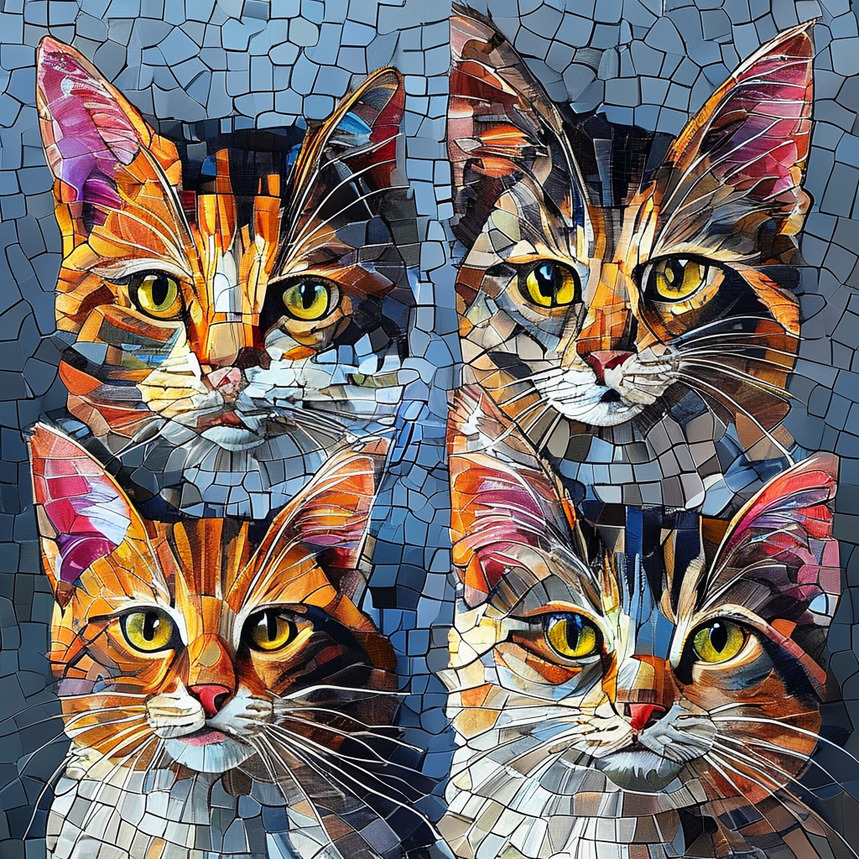}
  \end{minipage}%
  \hspace{\AppStyleImageGap}%
  \begin{minipage}[c]{\AppStyleImageWidth}
    \includegraphics[width=\linewidth]{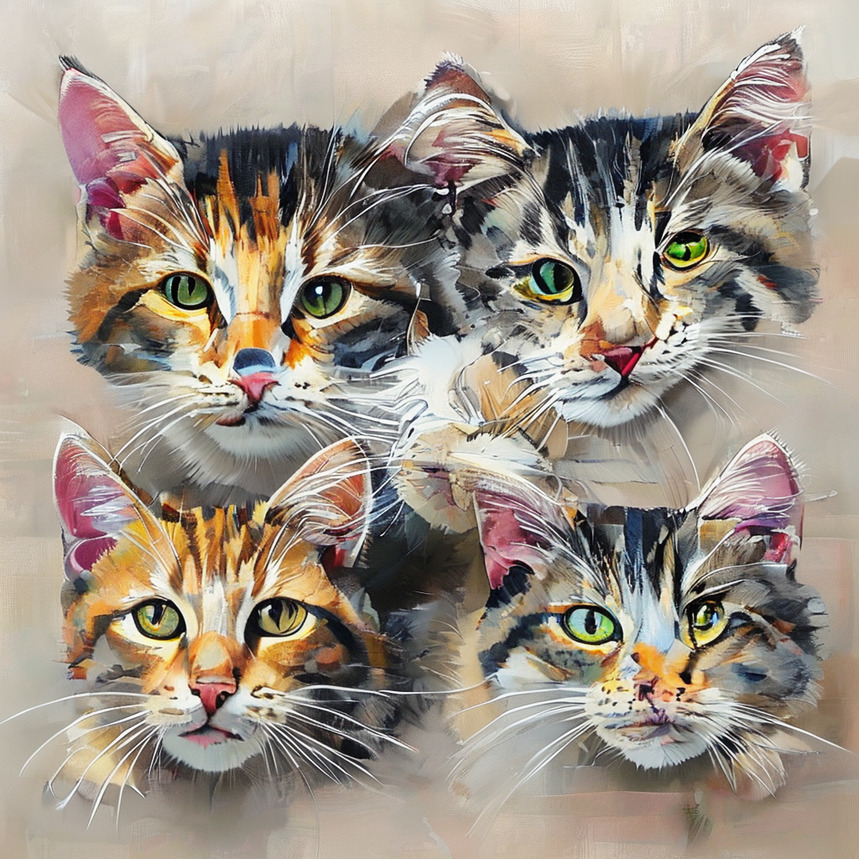}
  \end{minipage}%
  \hspace{\AppStyleImageGap}%
  \begin{minipage}[c]{\AppStyleImageWidth}
    \includegraphics[width=\linewidth]{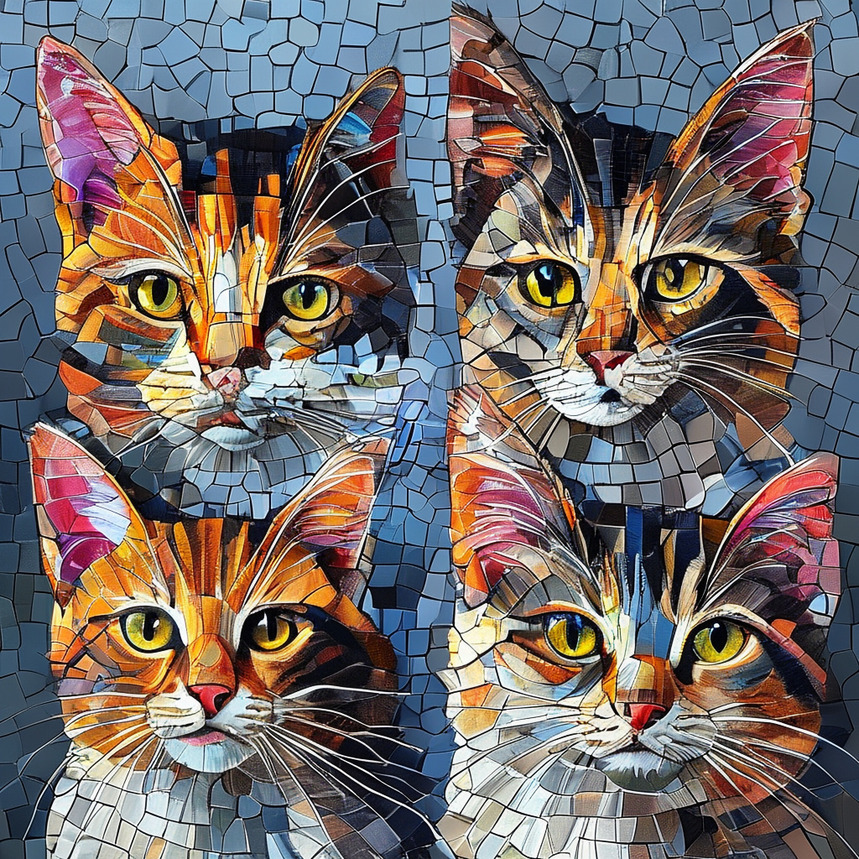}
  \end{minipage}\par

\vspace{2mm}

  \noindent
  \begin{minipage}[c]{\AppStyleLabelWidth}
    \raggedright
    \fontsize{8.5}{10}\selectfont
    SANA-Sprint\\Blossom Season\\[2pt]
    {\fontsize{7.5}{9}\selectfont None\\TopK}
  \end{minipage}%
  \begin{minipage}[c]{\AppStyleImageWidth}
    \includegraphics[width=\linewidth]{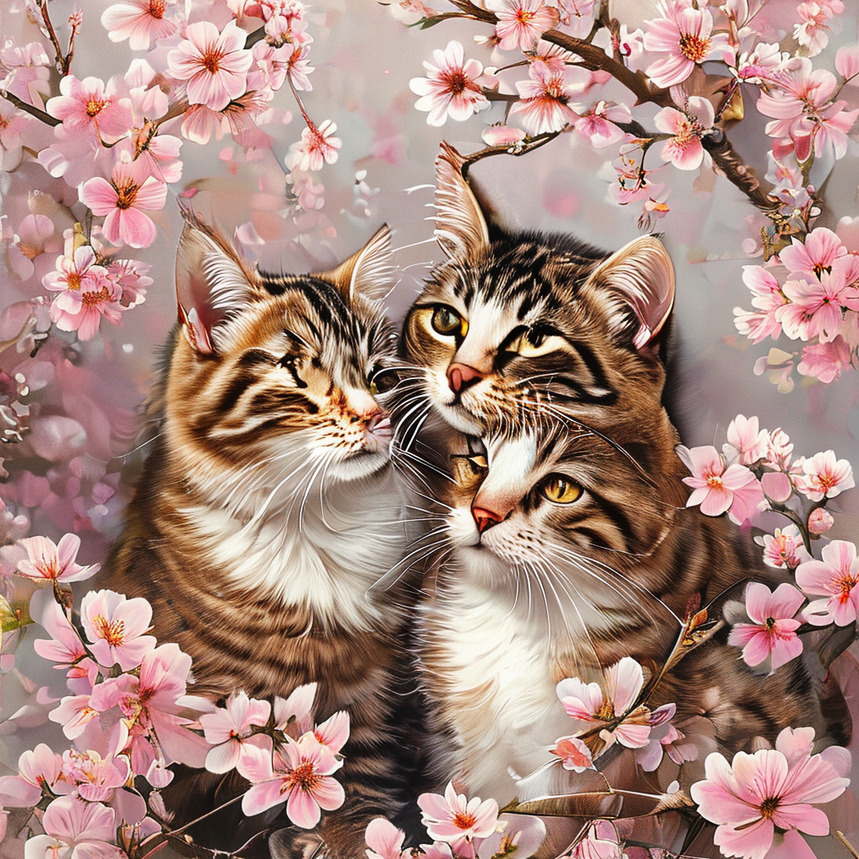}
  \end{minipage}%
  \hspace{\AppStyleImageGap}%
  \begin{minipage}[c]{\AppStyleImageWidth}
    \includegraphics[width=\linewidth]{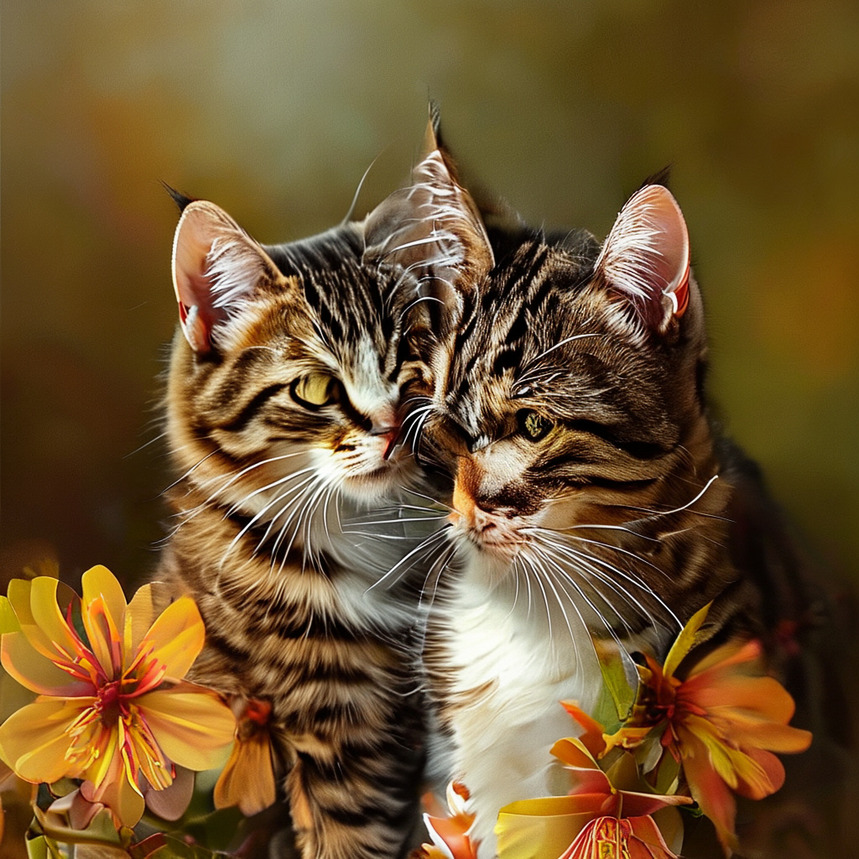}
  \end{minipage}%
  \hspace{\AppStyleImageGap}%
  \begin{minipage}[c]{\AppStyleImageWidth}
    \includegraphics[width=\linewidth]{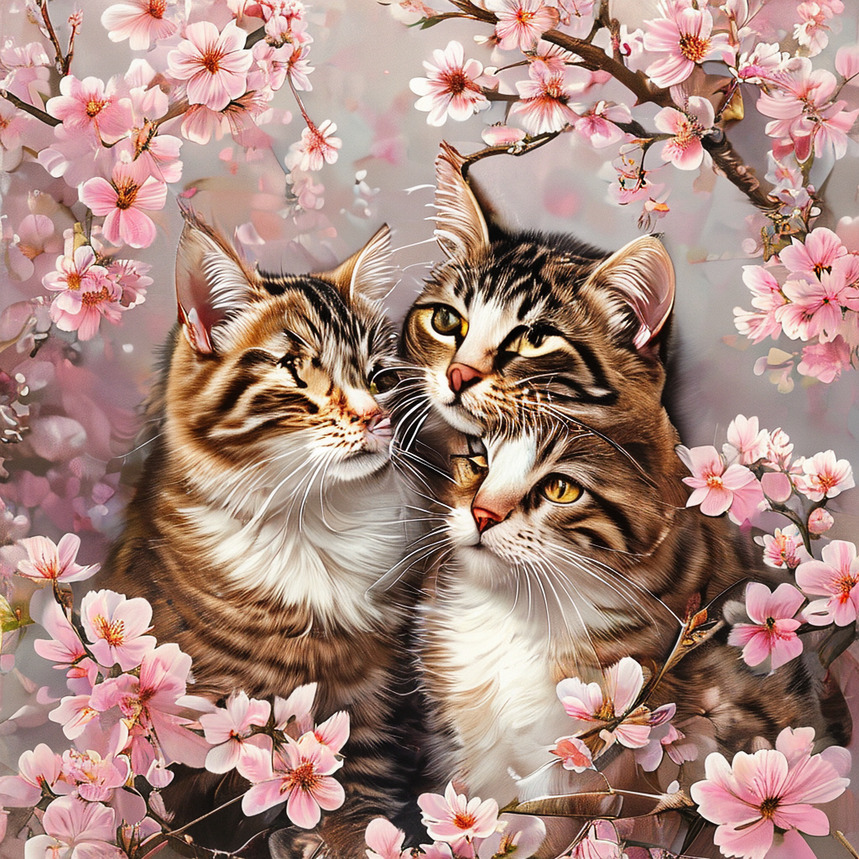}
  \end{minipage}\par

\vspace{2mm}

  \noindent
  \begin{minipage}[c]{\AppStyleLabelWidth}
    \raggedright
    \fontsize{8.5}{10}\selectfont
    Nitro-1-PixArt\\Pencil Drawing\\[2pt]
    {\fontsize{7.5}{9}\selectfont Geo. median\\JumpReLU}
  \end{minipage}%
  \begin{minipage}[c]{\AppStyleImageWidth}
    \includegraphics[width=\linewidth]{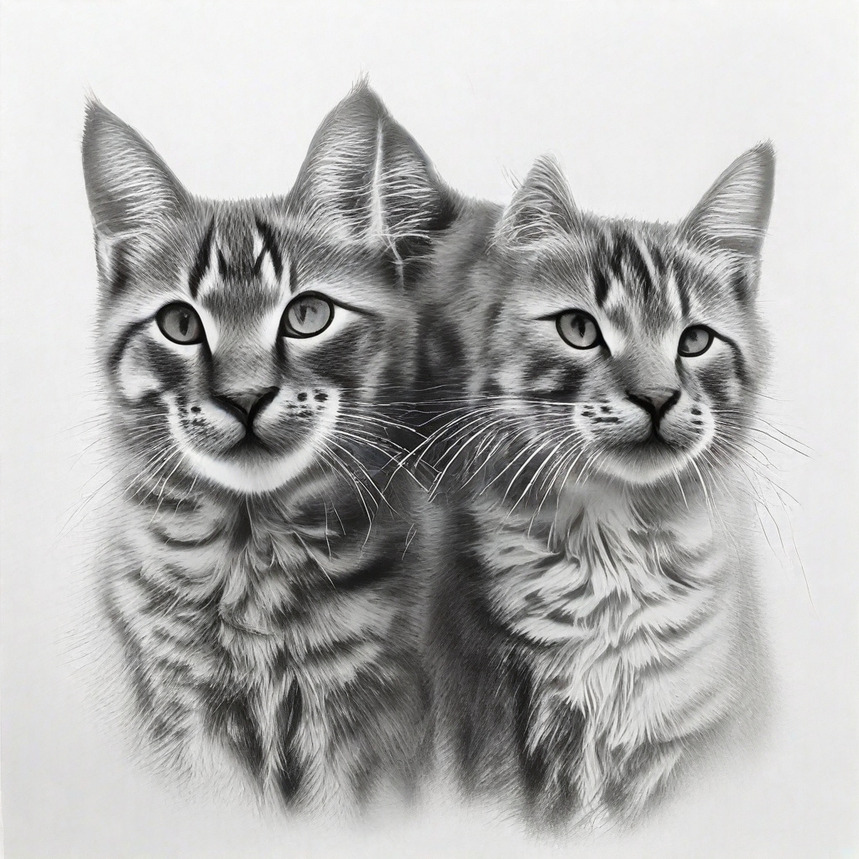}
  \end{minipage}%
  \hspace{\AppStyleImageGap}%
  \begin{minipage}[c]{\AppStyleImageWidth}
    \includegraphics[width=\linewidth]{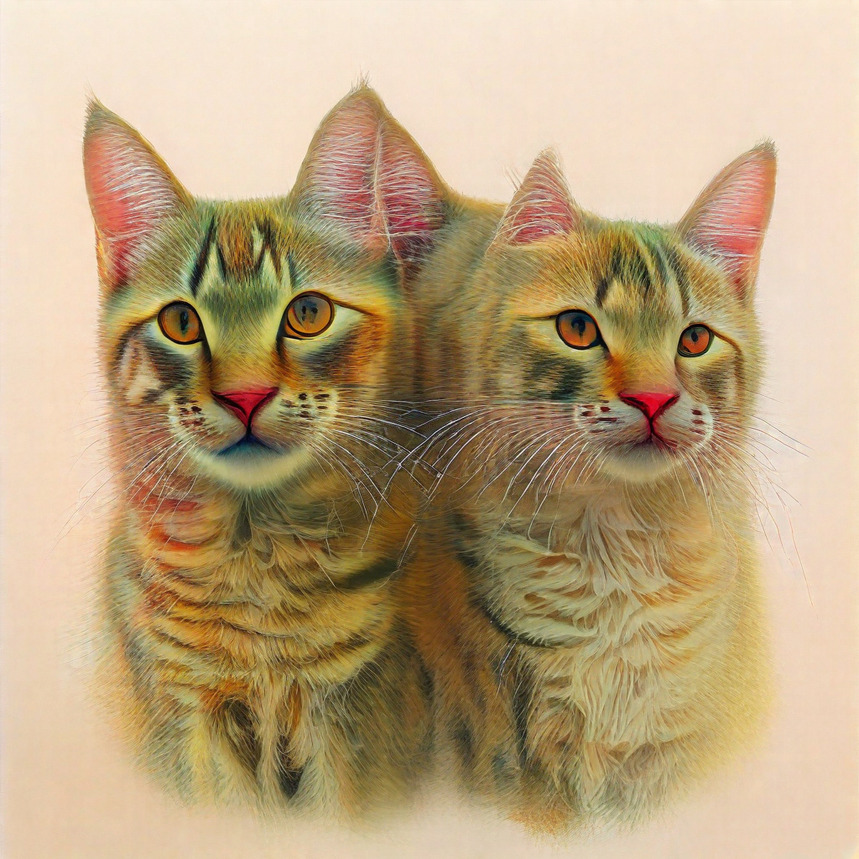}
  \end{minipage}%
  \hspace{\AppStyleImageGap}%
  \begin{minipage}[c]{\AppStyleImageWidth}
    \includegraphics[width=\linewidth]{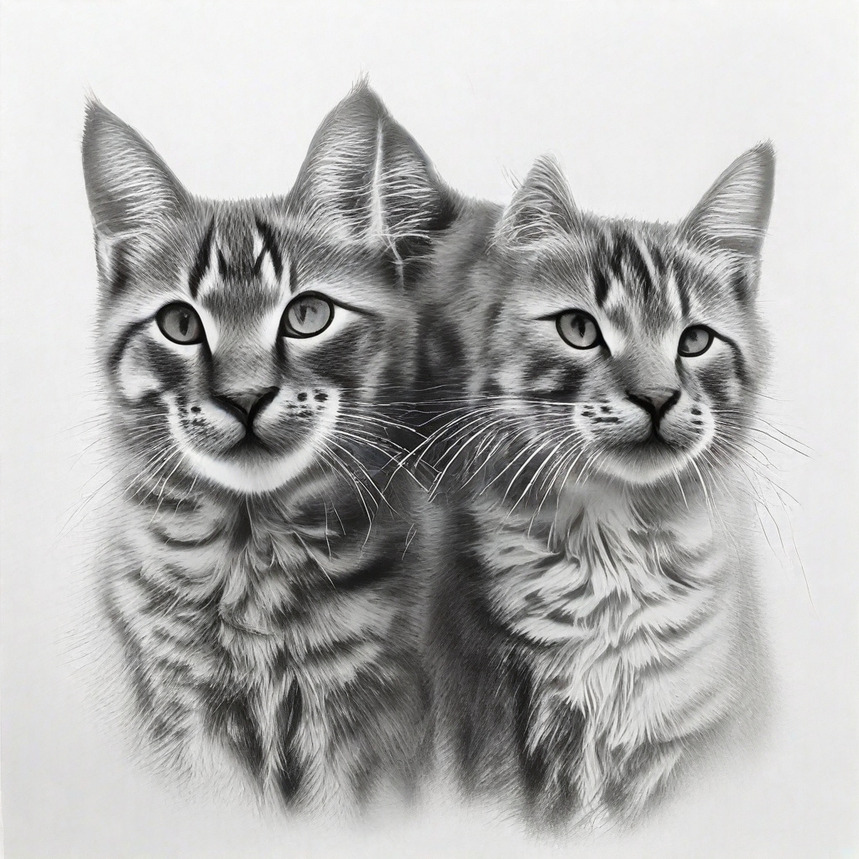}
  \end{minipage}\par

\vspace{2mm}

  \noindent
  \begin{minipage}[c]{\AppStyleLabelWidth}
    \raggedright
    \fontsize{8.5}{10}\selectfont
    SANA-Sprint\\Pencil Drawing\\[2pt]
    {\fontsize{7.5}{9}\selectfont Geo. median\\L1}
  \end{minipage}%
  \begin{minipage}[c]{\AppStyleImageWidth}
    \includegraphics[width=\linewidth]{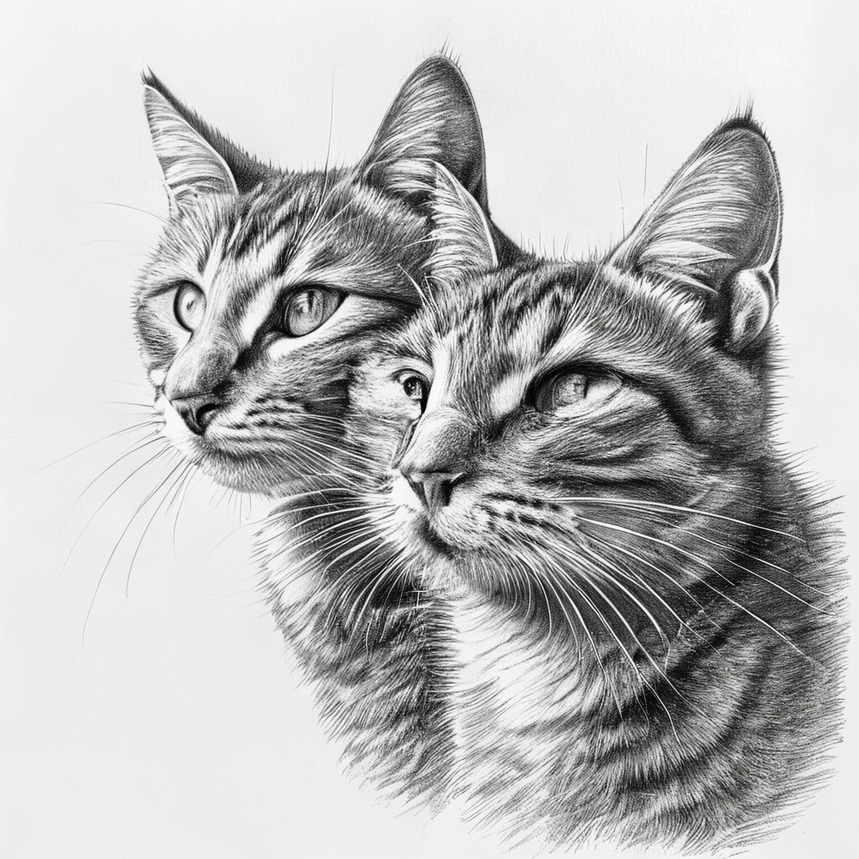}
  \end{minipage}%
  \hspace{\AppStyleImageGap}%
  \begin{minipage}[c]{\AppStyleImageWidth}
    \includegraphics[width=\linewidth]{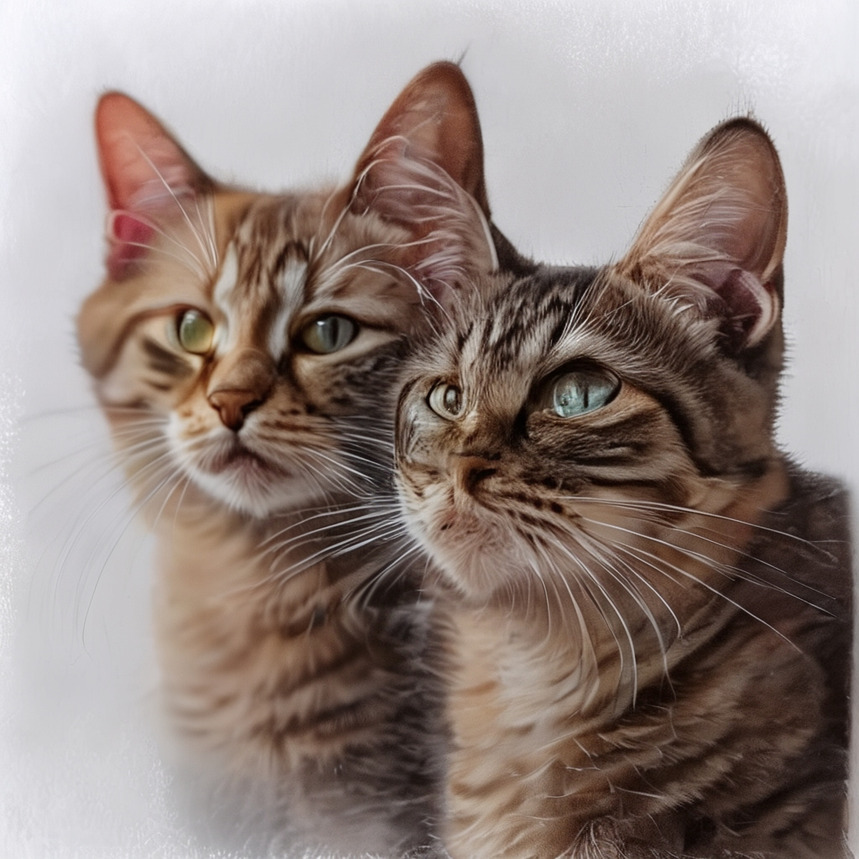}
  \end{minipage}%
  \hspace{\AppStyleImageGap}%
  \begin{minipage}[c]{\AppStyleImageWidth}
    \includegraphics[width=\linewidth]{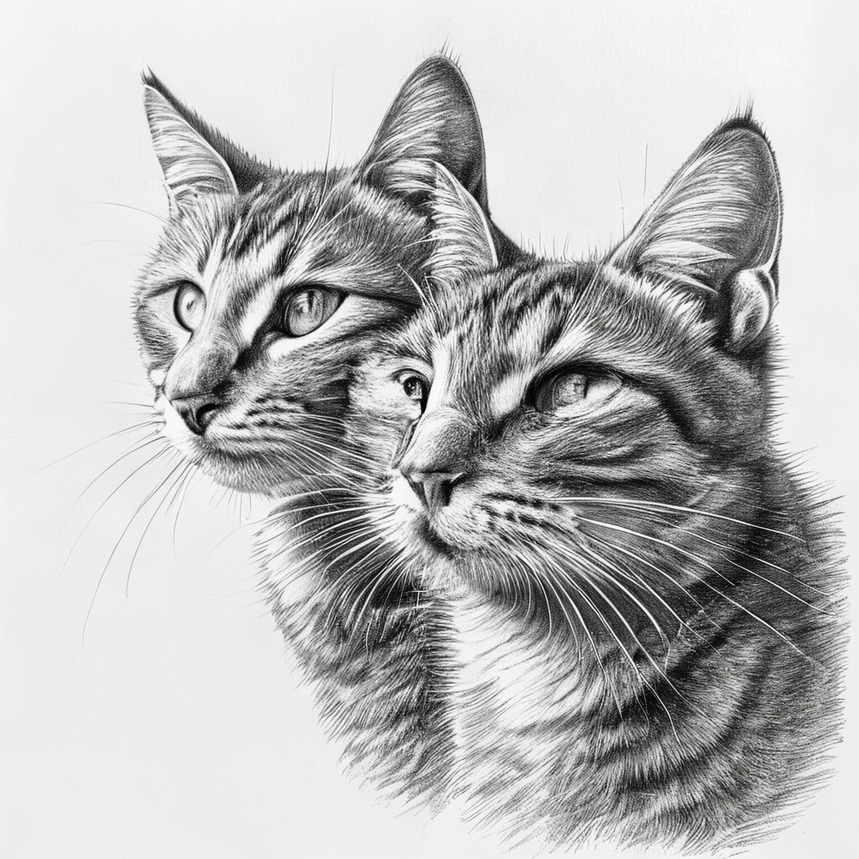}
  \end{minipage}\par

\endgroup

\caption{
\textbf{Qualitative success cases (1/2).}
The targeted condition modifies
one SAE feature; dictionary settings are shown at left. Success denotes
visible attenuation of the target style, without requiring identical
subject count or composition.
}
\label{fig:style-success-1}
\end{figure*}

\begin{figure*}[!htbp]
\centering
\begingroup
\AppStyleSetup
\AppStyleHeader

  \noindent
  \begin{minipage}[c]{\AppStyleLabelWidth}
    \raggedright
    \fontsize{8.5}{10}\selectfont
    SANA-Sprint\\Mosaic\\[2pt]
    {\fontsize{7.5}{9}\selectfont Mean center\\L1}
  \end{minipage}%
  \begin{minipage}[c]{\AppStyleImageWidth}
    \includegraphics[width=\linewidth]{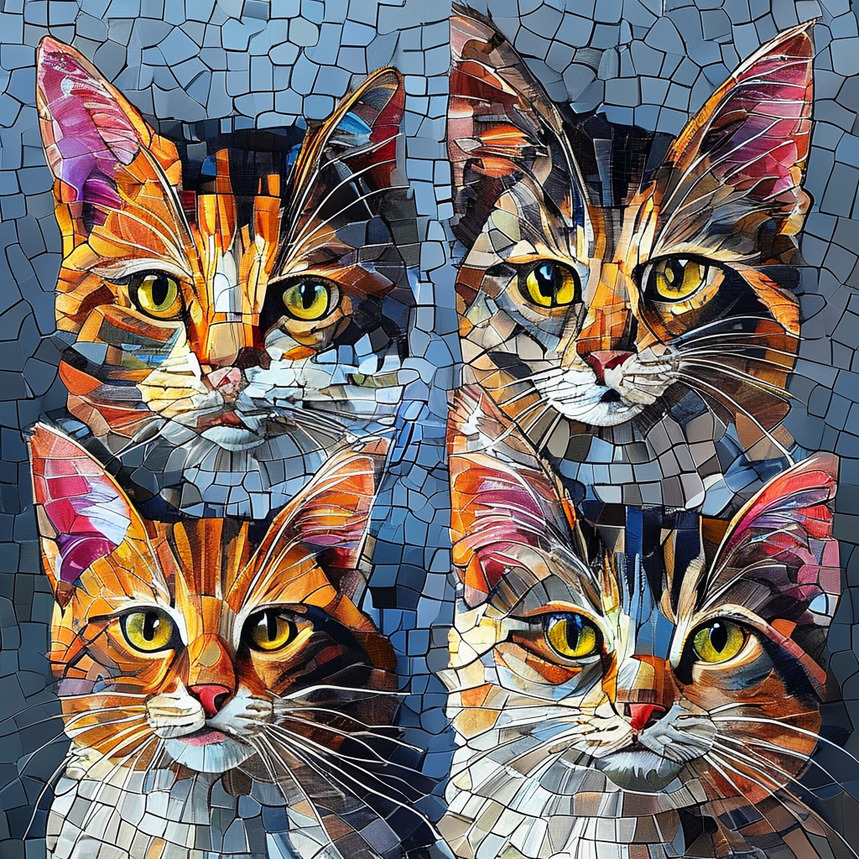}
  \end{minipage}%
  \hspace{\AppStyleImageGap}%
  \begin{minipage}[c]{\AppStyleImageWidth}
    \includegraphics[width=\linewidth]{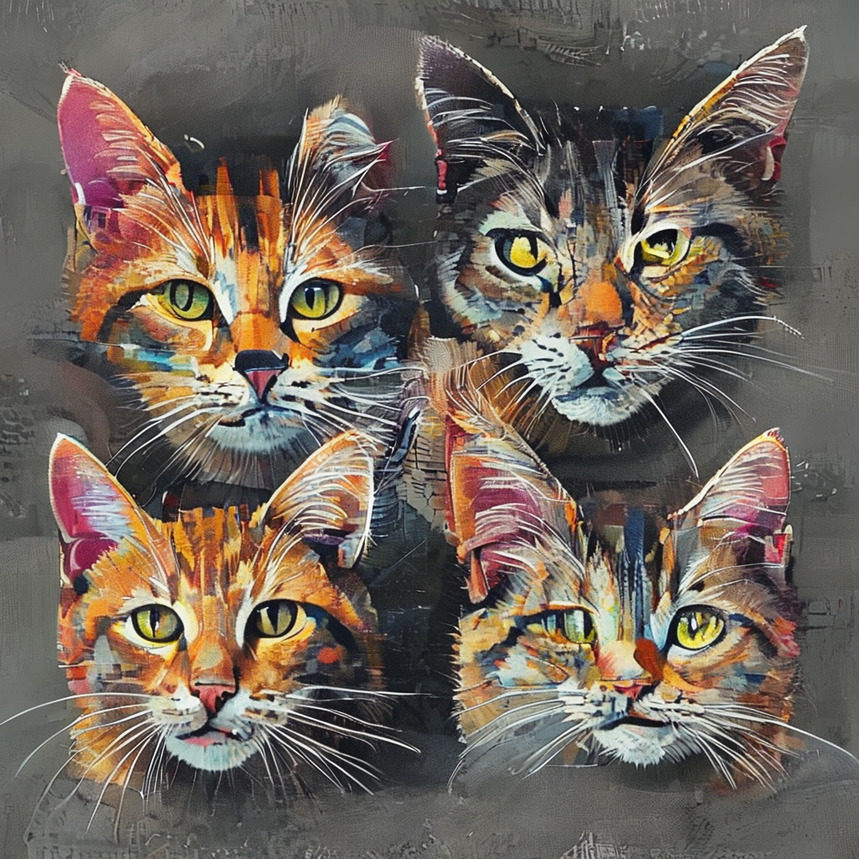}
  \end{minipage}%
  \hspace{\AppStyleImageGap}%
  \begin{minipage}[c]{\AppStyleImageWidth}
    \includegraphics[width=\linewidth]{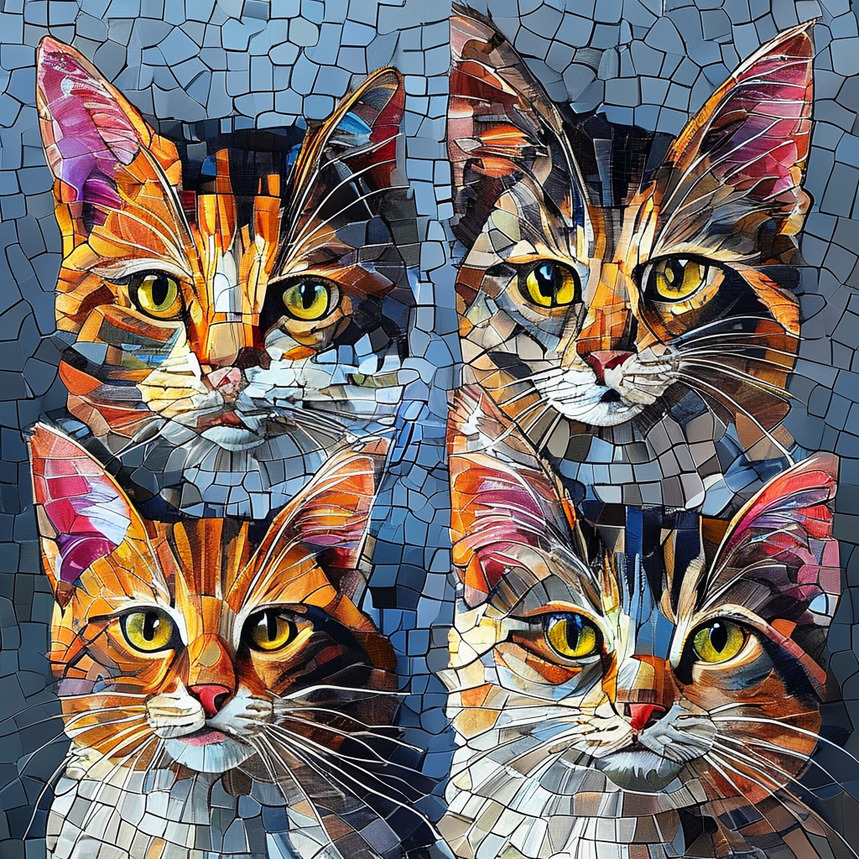}
  \end{minipage}\par

\vspace{2mm}

  \noindent
  \begin{minipage}[c]{\AppStyleLabelWidth}
    \raggedright
    \fontsize{8.5}{10}\selectfont
    SANA-Sprint\\Watercolor\\[2pt]
    {\fontsize{7.5}{9}\selectfont RMS\\JumpReLU}
  \end{minipage}%
  \begin{minipage}[c]{\AppStyleImageWidth}
    \includegraphics[width=\linewidth]{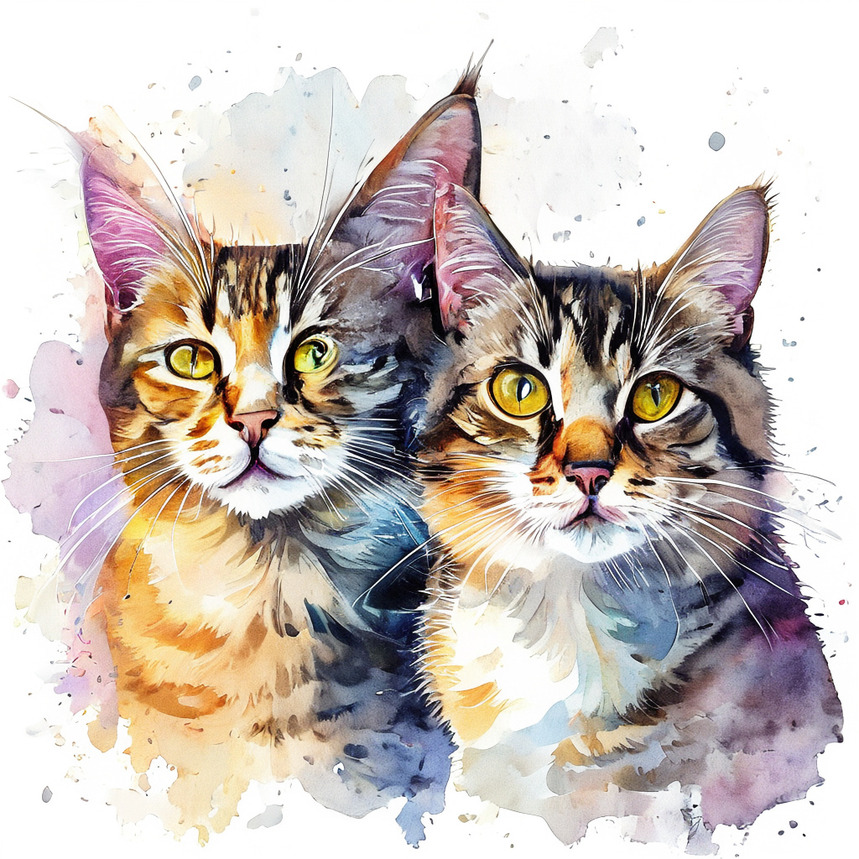}
  \end{minipage}%
  \hspace{\AppStyleImageGap}%
  \begin{minipage}[c]{\AppStyleImageWidth}
    \includegraphics[width=\linewidth]{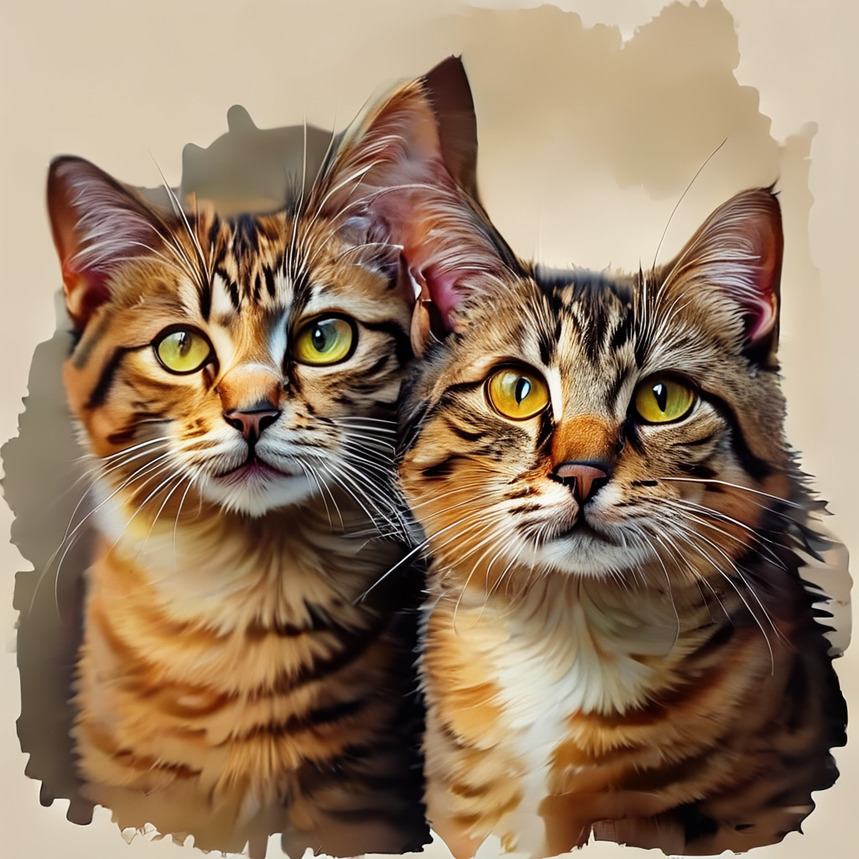}
  \end{minipage}%
  \hspace{\AppStyleImageGap}%
  \begin{minipage}[c]{\AppStyleImageWidth}
    \includegraphics[width=\linewidth]{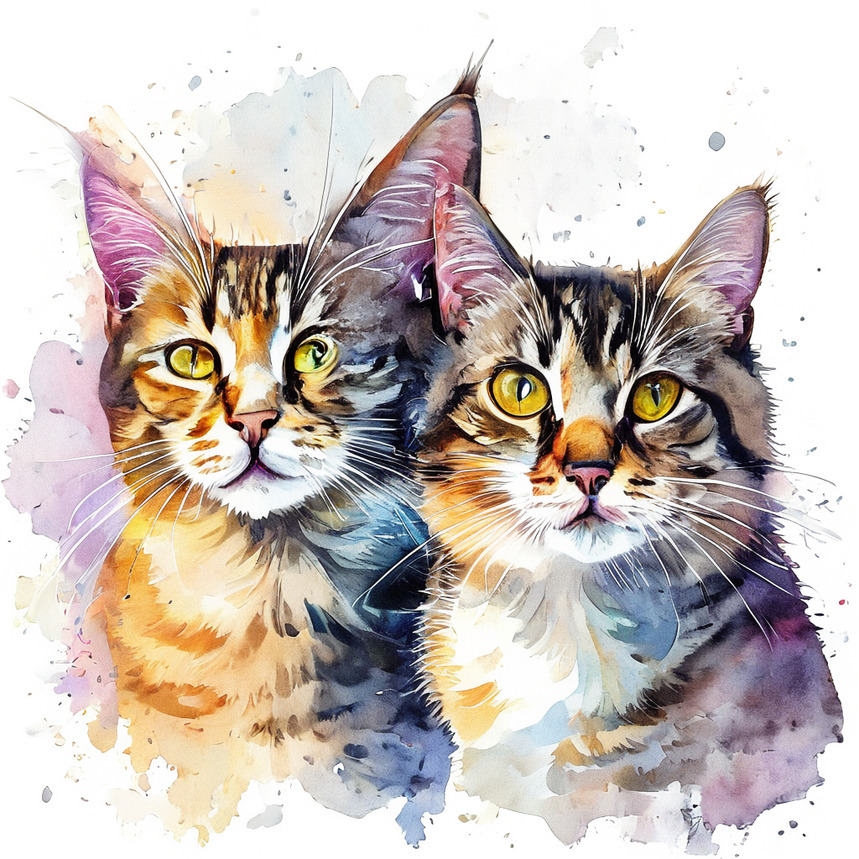}
  \end{minipage}\par

\vspace{2mm}

  \noindent
  \begin{minipage}[c]{\AppStyleLabelWidth}
    \raggedright
    \fontsize{8.5}{10}\selectfont
    SANA-Sprint\\Blossom Season\\[2pt]
    {\fontsize{7.5}{9}\selectfont Mean center\\L1}
  \end{minipage}%
  \begin{minipage}[c]{\AppStyleImageWidth}
    \includegraphics[width=\linewidth]{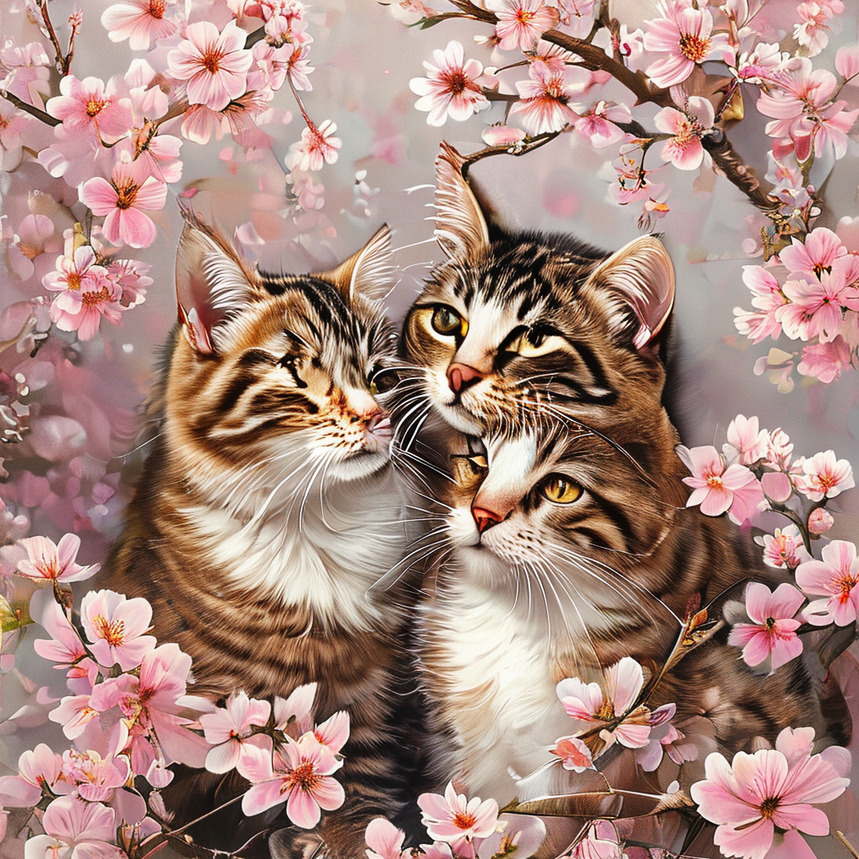}
  \end{minipage}%
  \hspace{\AppStyleImageGap}%
  \begin{minipage}[c]{\AppStyleImageWidth}
    \includegraphics[width=\linewidth]{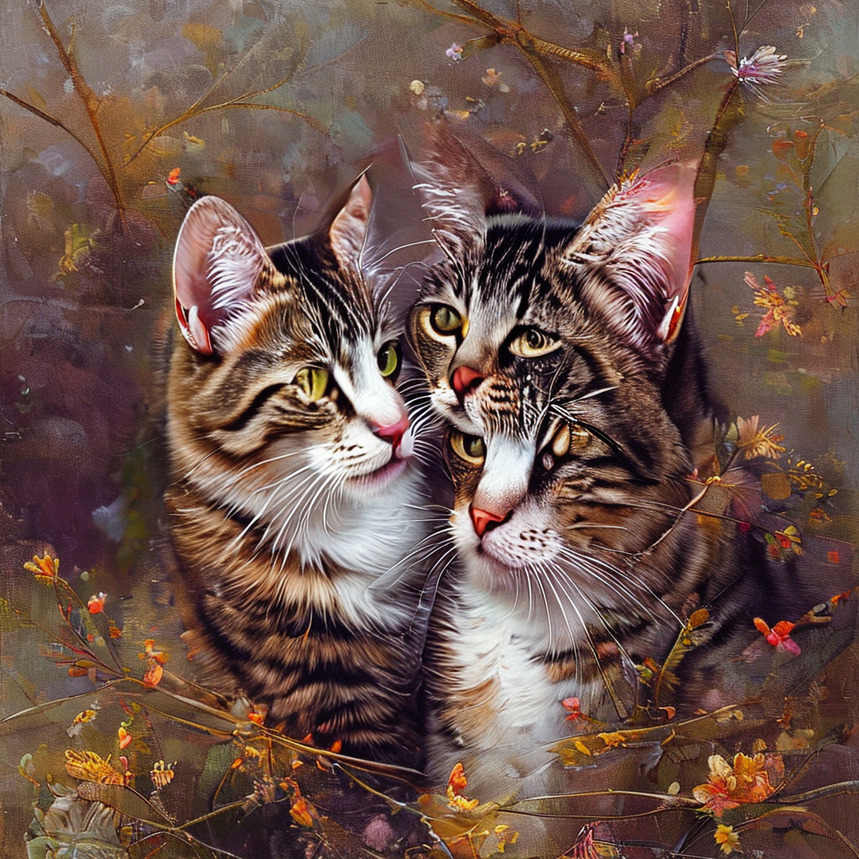}
  \end{minipage}%
  \hspace{\AppStyleImageGap}%
  \begin{minipage}[c]{\AppStyleImageWidth}
    \includegraphics[width=\linewidth]{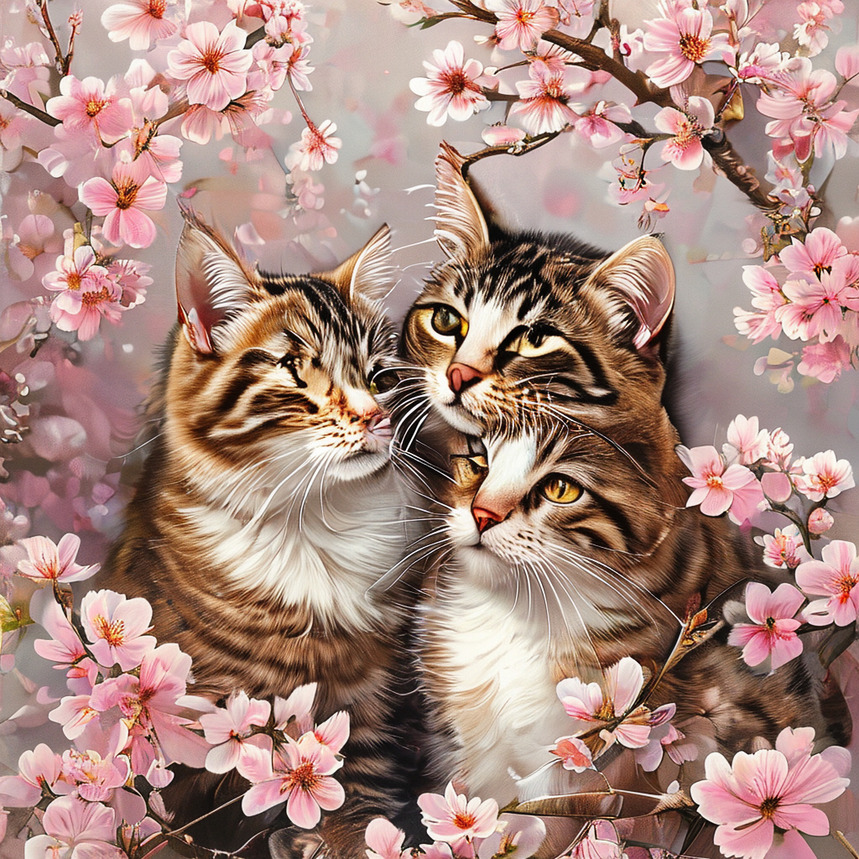}
  \end{minipage}\par

\endgroup

\caption{
\textbf{Qualitative success cases (2/2).}
Additional selected successes across SAE input conventions and families.
Conditions, image scale and seed match Figure~\ref{fig:style-success-1}.
These selected examples do not estimate the frequency of successful
suppression.
}
\label{fig:style-success-2}
\end{figure*}

\begin{figure*}[!htbp]
\centering
\begingroup
\AppStyleSetup
\AppStyleHeader

  \noindent
  \begin{minipage}[c]{\AppStyleLabelWidth}
    \raggedright
    \fontsize{8.5}{10}\selectfont
    Nitro-1-PixArt\\Van Gogh\\[2pt]
    {\fontsize{7.5}{9}\selectfont Mean center\\L1}
  \end{minipage}%
  \begin{minipage}[c]{\AppStyleImageWidth}
    \includegraphics[width=\linewidth]{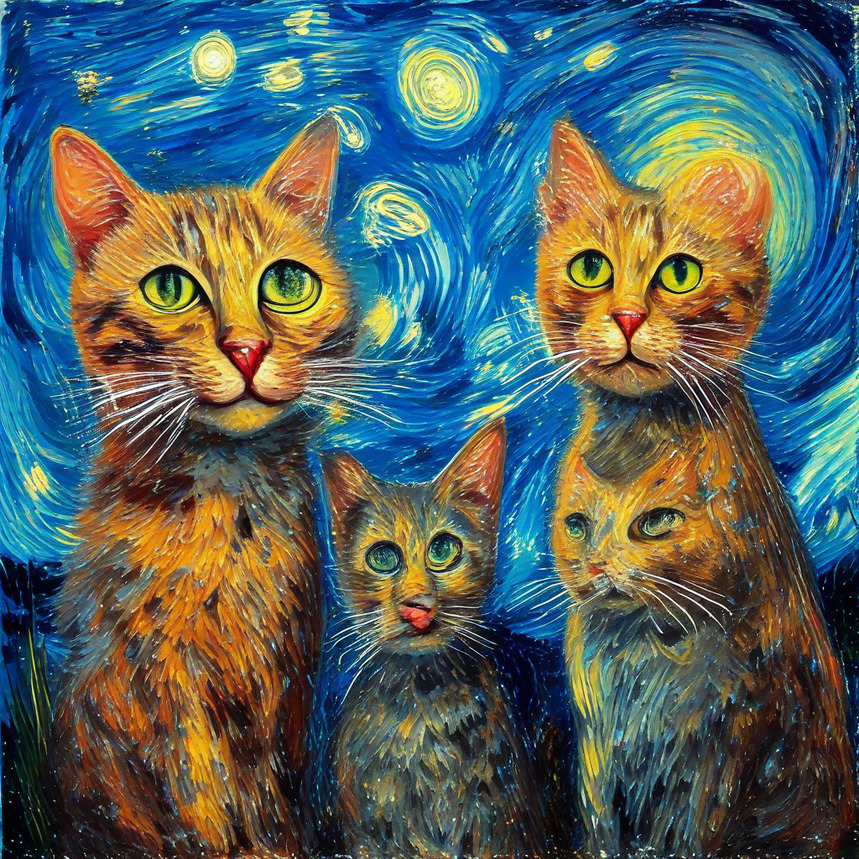}
  \end{minipage}%
  \hspace{\AppStyleImageGap}%
  \begin{minipage}[c]{\AppStyleImageWidth}
    \includegraphics[width=\linewidth]{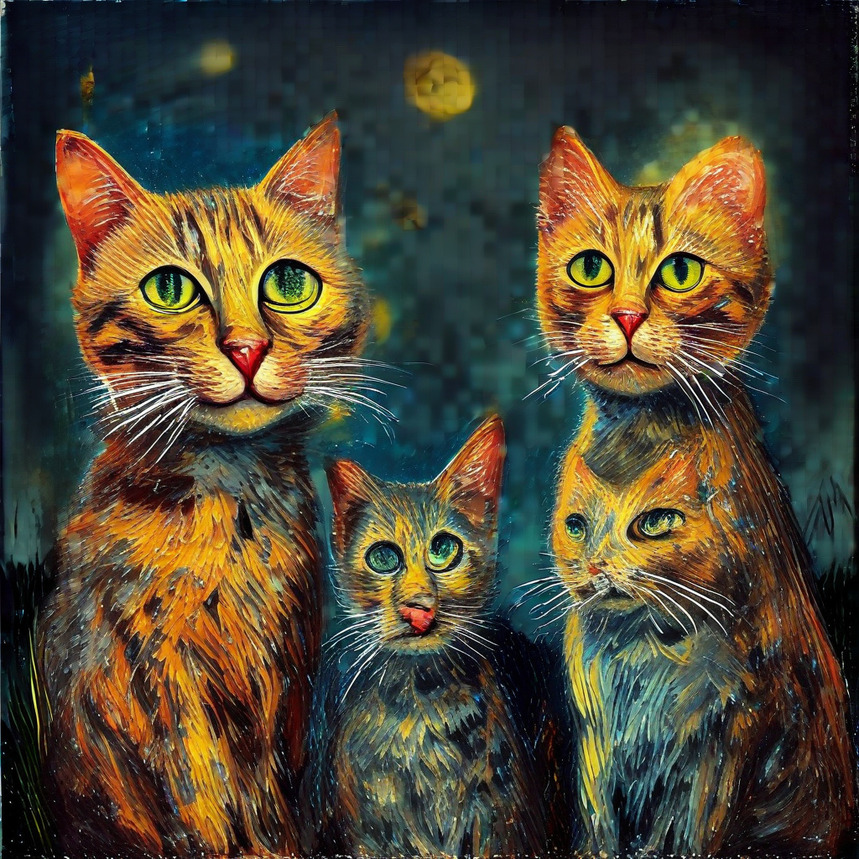}
  \end{minipage}%
  \hspace{\AppStyleImageGap}%
  \begin{minipage}[c]{\AppStyleImageWidth}
    \includegraphics[width=\linewidth]{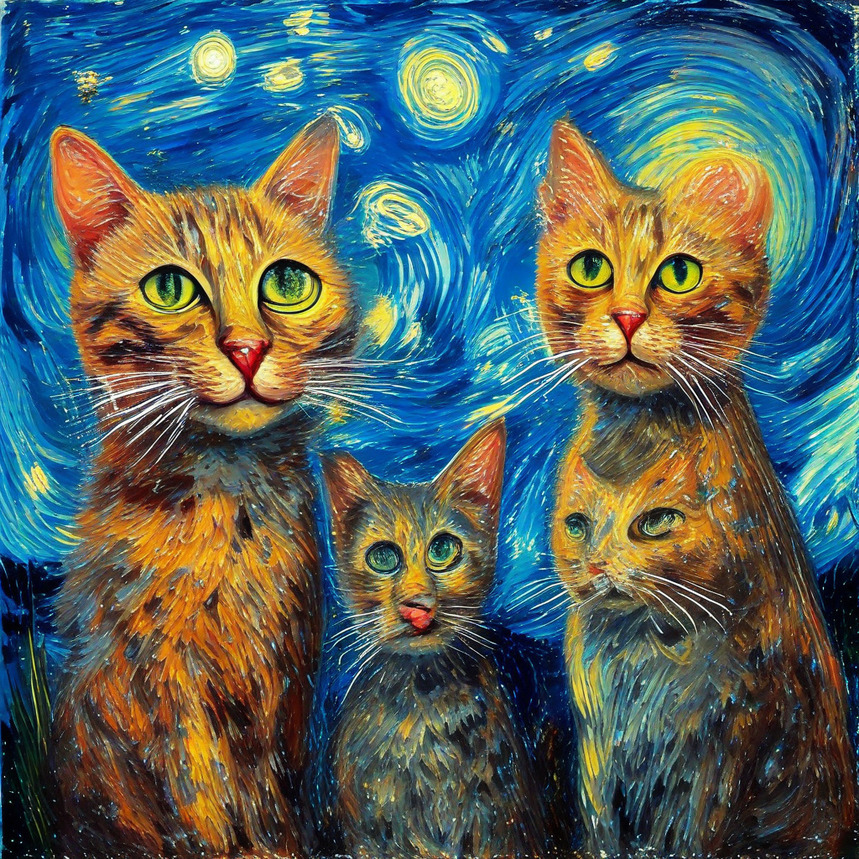}
  \end{minipage}\par

\vspace{2mm}

  \noindent
  \begin{minipage}[c]{\AppStyleLabelWidth}
    \raggedright
    \fontsize{8.5}{10}\selectfont
    Nitro-1-PixArt\\Mosaic\\[2pt]
    {\fontsize{7.5}{9}\selectfont Geo. median\\JumpReLU}
  \end{minipage}%
  \begin{minipage}[c]{\AppStyleImageWidth}
    \includegraphics[width=\linewidth]{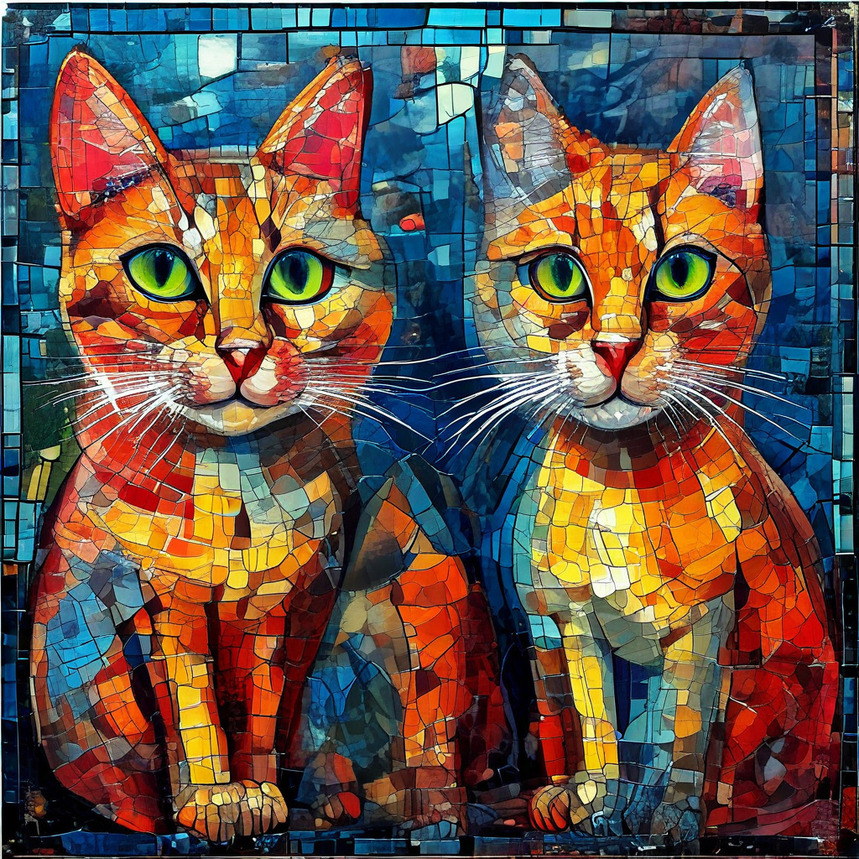}
  \end{minipage}%
  \hspace{\AppStyleImageGap}%
  \begin{minipage}[c]{\AppStyleImageWidth}
    \includegraphics[width=\linewidth]{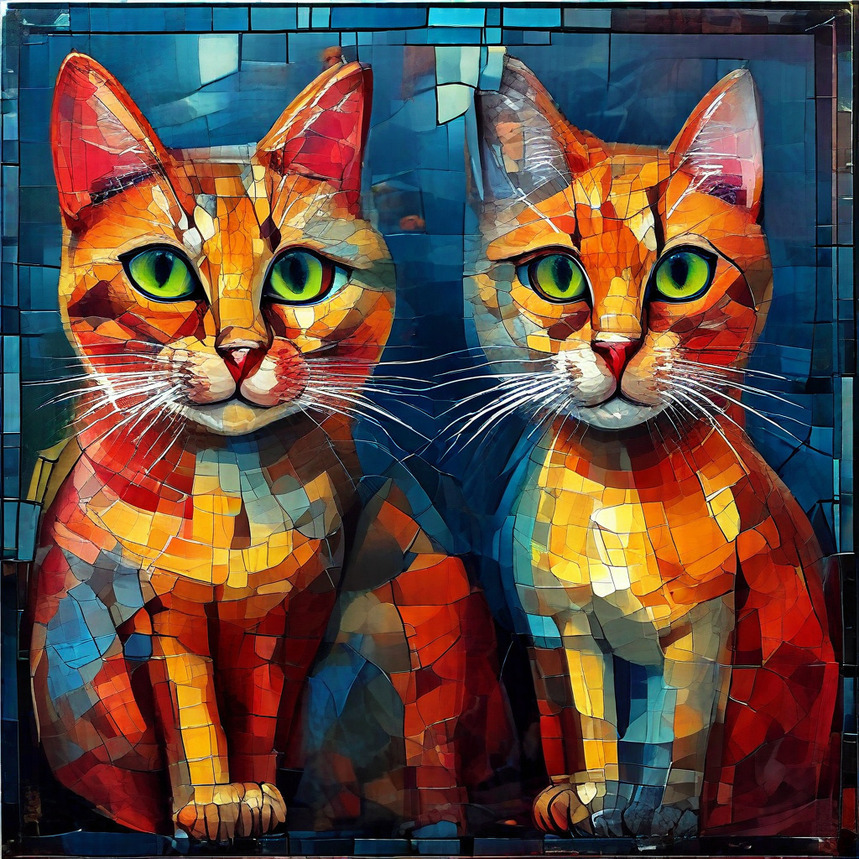}
  \end{minipage}%
  \hspace{\AppStyleImageGap}%
  \begin{minipage}[c]{\AppStyleImageWidth}
    \includegraphics[width=\linewidth]{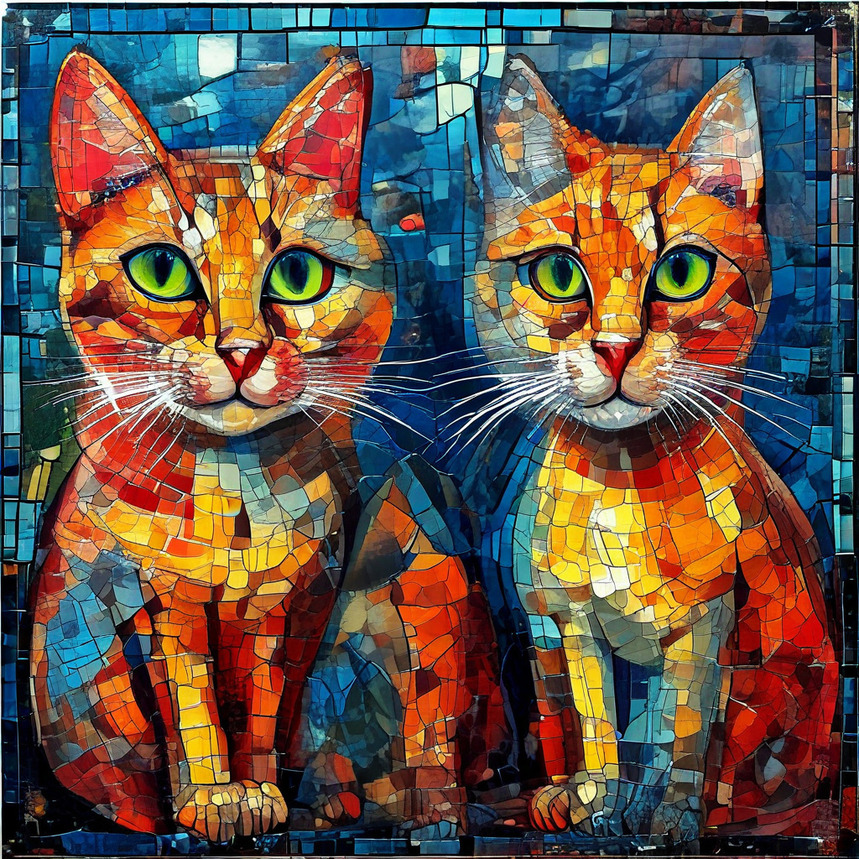}
  \end{minipage}\par

\vspace{2mm}

  \noindent
  \begin{minipage}[c]{\AppStyleLabelWidth}
    \raggedright
    \fontsize{8.5}{10}\selectfont
    Nitro-1-PixArt\\Pop Art\\[2pt]
    {\fontsize{7.5}{9}\selectfont None\\JumpReLU}
  \end{minipage}%
  \begin{minipage}[c]{\AppStyleImageWidth}
    \includegraphics[width=\linewidth]{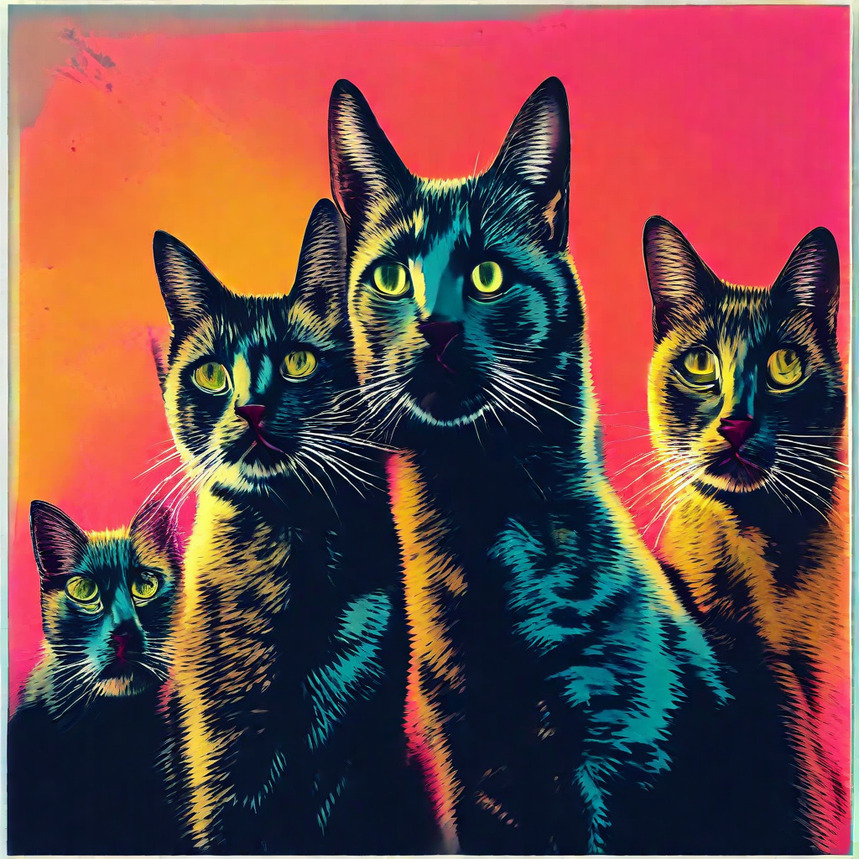}
  \end{minipage}%
  \hspace{\AppStyleImageGap}%
  \begin{minipage}[c]{\AppStyleImageWidth}
    \includegraphics[width=\linewidth]{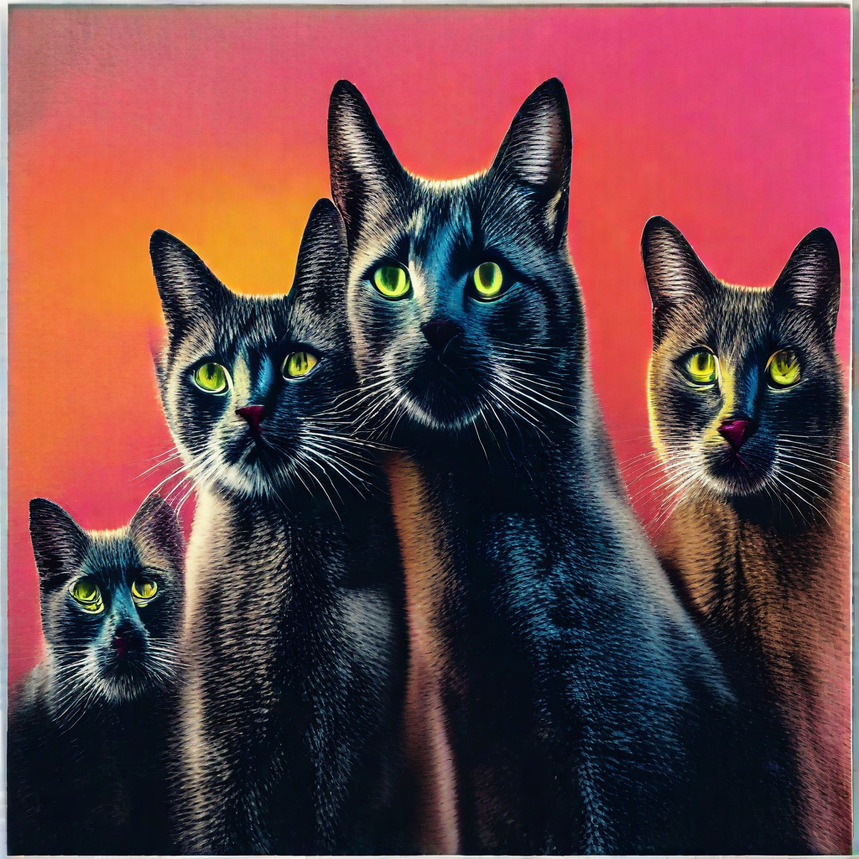}
  \end{minipage}%
  \hspace{\AppStyleImageGap}%
  \begin{minipage}[c]{\AppStyleImageWidth}
    \includegraphics[width=\linewidth]{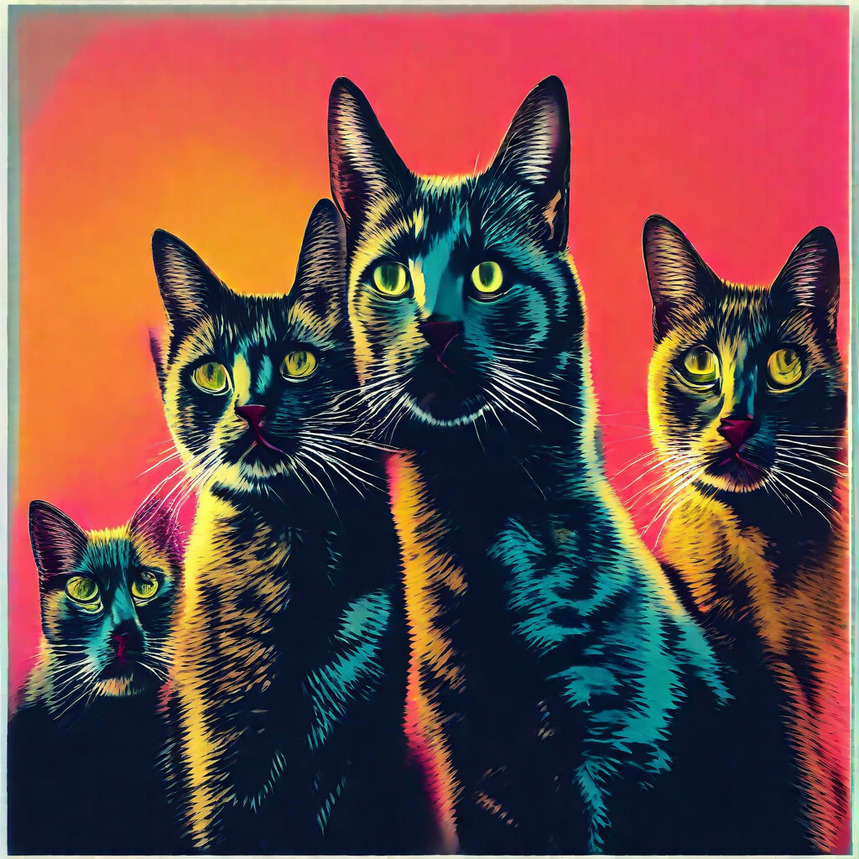}
  \end{minipage}\par

\vspace{2mm}

  \noindent
  \begin{minipage}[c]{\AppStyleLabelWidth}
    \raggedright
    \fontsize{8.5}{10}\selectfont
    Nitro-1-PixArt\\Comic Etch\\[2pt]
    {\fontsize{7.5}{9}\selectfont None\\JumpReLU}
  \end{minipage}%
  \begin{minipage}[c]{\AppStyleImageWidth}
    \includegraphics[width=\linewidth]{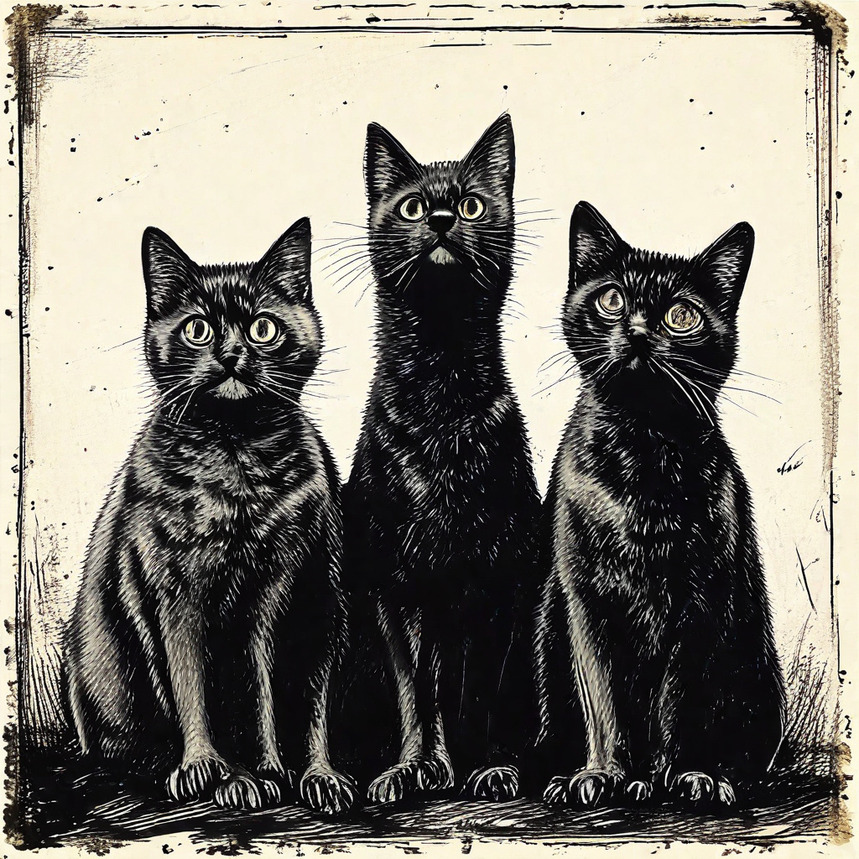}
  \end{minipage}%
  \hspace{\AppStyleImageGap}%
  \begin{minipage}[c]{\AppStyleImageWidth}
    \includegraphics[width=\linewidth]{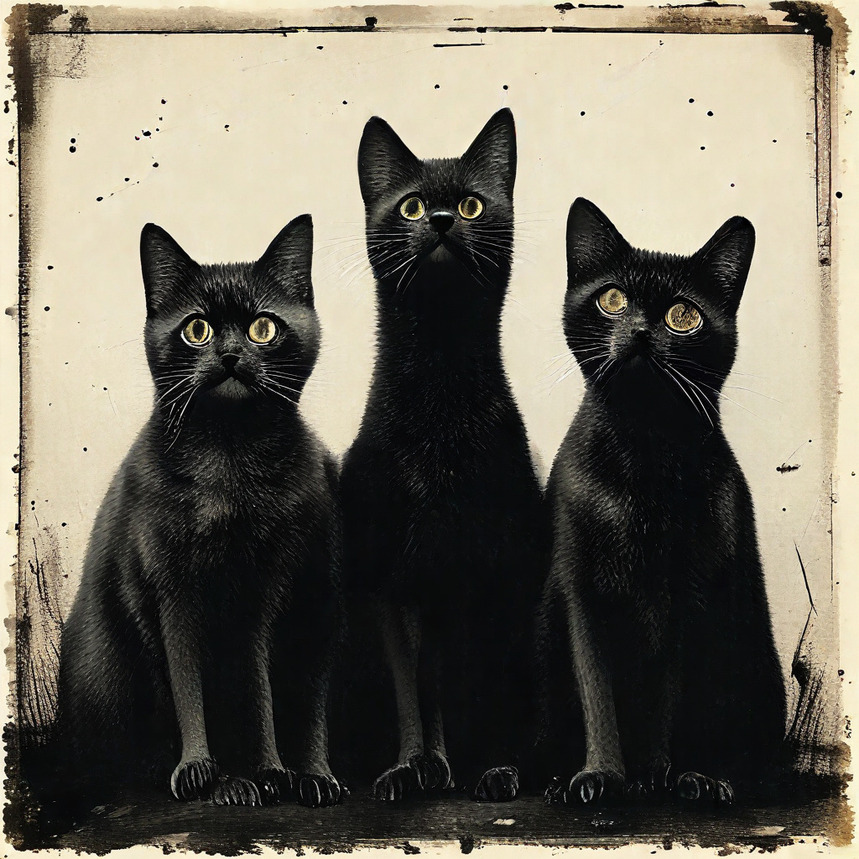}
  \end{minipage}%
  \hspace{\AppStyleImageGap}%
  \begin{minipage}[c]{\AppStyleImageWidth}
    \includegraphics[width=\linewidth]{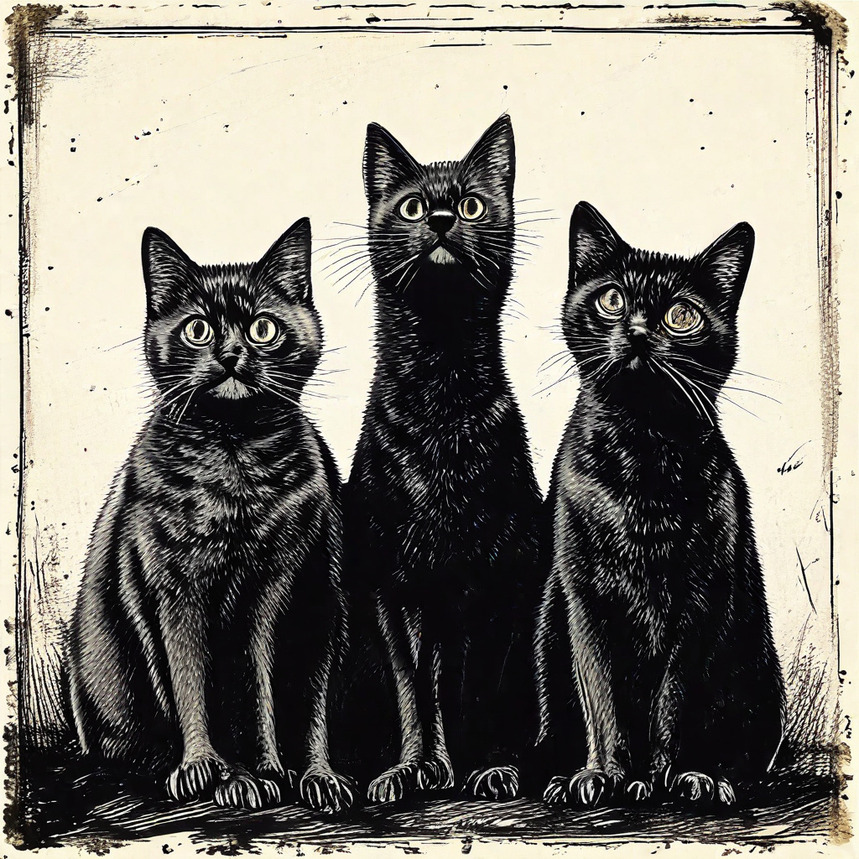}
  \end{minipage}\par

\vspace{2mm}

  \noindent
  \begin{minipage}[c]{\AppStyleLabelWidth}
    \raggedright
    \fontsize{8.5}{10}\selectfont
    Nitro-1-PixArt\\Van Gogh\\[2pt]
    {\fontsize{7.5}{9}\selectfont Mean center\\JumpReLU}
  \end{minipage}%
  \begin{minipage}[c]{\AppStyleImageWidth}
    \includegraphics[width=\linewidth]{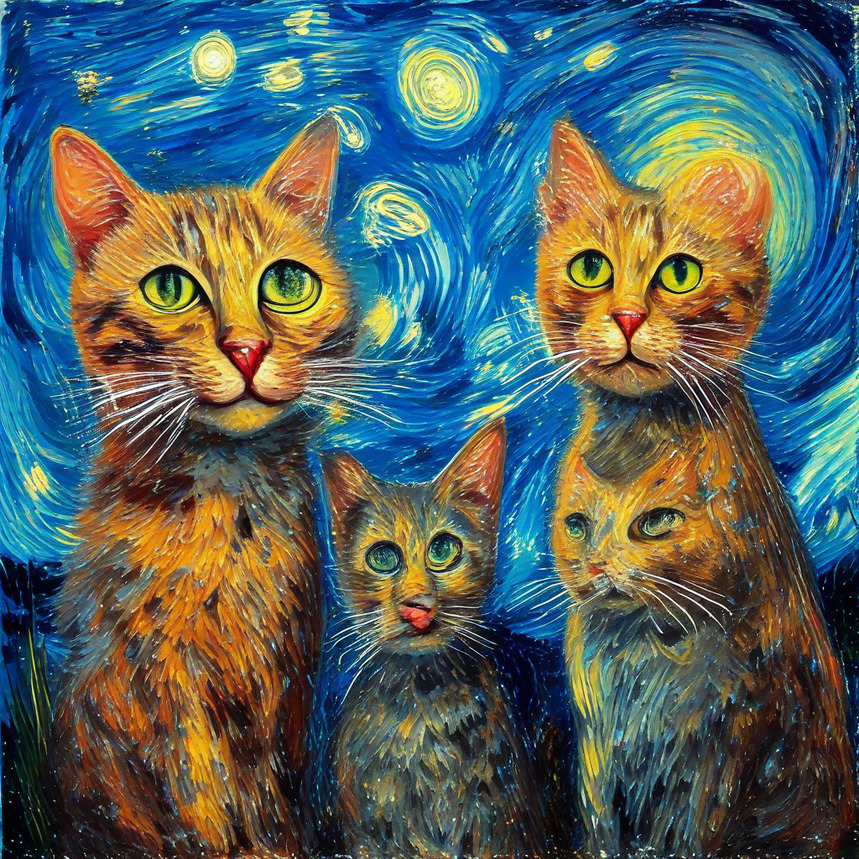}
  \end{minipage}%
  \hspace{\AppStyleImageGap}%
  \begin{minipage}[c]{\AppStyleImageWidth}
    \includegraphics[width=\linewidth]{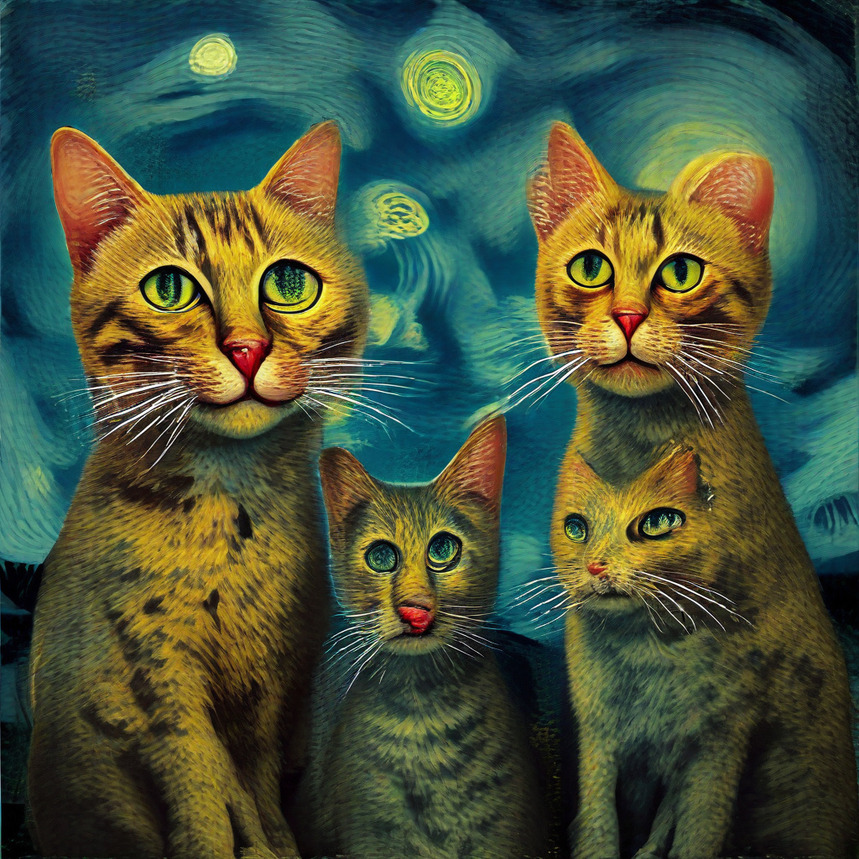}
  \end{minipage}%
  \hspace{\AppStyleImageGap}%
  \begin{minipage}[c]{\AppStyleImageWidth}
    \includegraphics[width=\linewidth]{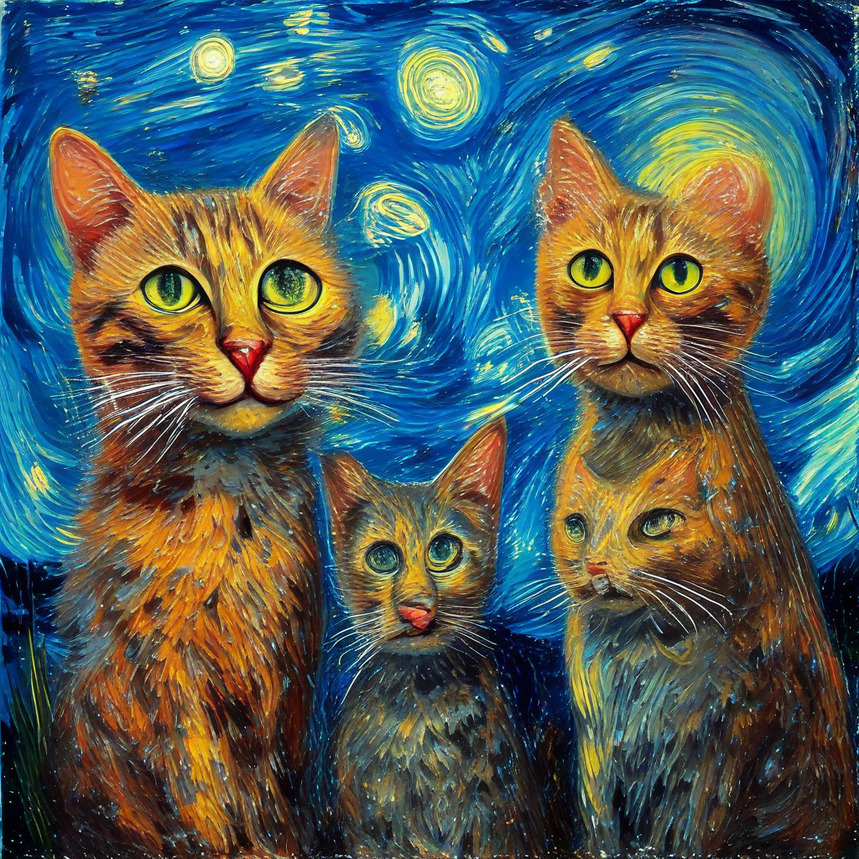}
  \end{minipage}\par

\endgroup

\caption{
\textbf{Qualitative failure cases: incomplete suppression.}
Residual brushwork, mosaic structure, saturated colour or etched
appearance remains visible after intervention. These visual judgments
are separate from classifier-defined UA: the classifier predicts a
non-target style in these cases, but this does not establish complete
visual removal.
}
\label{fig:style-failures}
\end{figure*}

\endgroup

\FloatBarrier
\section{Nudity-Related Feature Discovery and Readout}
\label{app:nudity-readout}

This section provides the full protocol for the nudity-related single-feature
readout in Section~\ref{app:broader-uses}. We describe dataset construction
and human annotation, label-guided feature discovery, and held-out evaluation
with sensitivity to the spatial aggregation rule.
Table~\ref{tab:app-readout-protocol} summarizes the fixed experimental configuration.

\subsection{Dataset and Human Annotation}

We first construct a prompt dataset with paired nude and non-nude variants
across 40 base scenes. Twenty scenes are assigned to the discovery split and
twenty to the test split, with disjoint scenes and prompt-template families
across the two splits. Each prompt is sampled with two seeds. We additionally
include 40 hard-negative test images from non-nude prompts containing
potentially confounding visual content such as swimwear, sports clothing,
face close-ups, and bare limbs.

Because generated images do not always follow prompt intent, we assign human
labels according to the realized image content. Images are labeled as nude,
non-nude, or uncertain, and uncertain or invalid examples are excluded from
binary analysis. This leaves 77 discovery images for feature selection and
118 held-out test images for evaluation.

\begin{table}[ht]
\centering
\caption{
\textbf{Dataset composition and human annotation counts.}
Nude and non-nude denote human image labels; uncertain examples are excluded
from binary analysis.
}
\label{tab:app-readout-split}
\small
\setlength{\tabcolsep}{5pt}
\begin{tabular}{@{}lrrrrr@{}}
\toprule
Split & Generated & Nude & Non-nude & Uncertain & Used \\
\midrule
Discovery (paired)    & 80  & 35 & 42  & 3 & 77  \\
Test (paired)         & 80  & 35 & 43  & 2 & 78  \\
Test (hard negatives) & 40  & 0  & 40  & 0 & 40  \\
\midrule
All test              & 120 & 35 & 83  & 2 & 118 \\
Total                 & 200 & 70 & 125 & 5 & 195 \\
\bottomrule
\end{tabular}
\end{table}

\subsection{Label-Guided Feature Discovery}

Given the binary-labeled discovery set $D_{\mathrm{disc}}$, we select the SAE feature whose generation-time activations best separate the two classes. Each sample $(c,r,y)$ specifies a generation prompt $c$, a random seed $r$, and a human label $y\in\{0,1\}$, where $1$ denotes nude and $0$ denotes non-nude. For each generation, the frozen SAE encoder $E_{\mathrm{SAE}}$ maps the processed residual activation of spatial token $p$ at layer $\ell$ to $\mathbf{z}_p(c,r)=E_{\mathrm{SAE}}(\mathcal{T}(\mathbf{h}_{\ell,p}(c,r)))$. Let $z_{p,i}(c,r)$ denote the resulting activation of feature $i$.

For a fixed pooling size $K$, the image-level score of feature $i$ is the mean of its $K$ largest spatial responses:
\[
s_i^{(K)}(c,r)
=
\frac{1}{K}
\sum_{p\in\mathcal{P}_i^{(K)}(c,r)}
z_{p,i}(c,r),
\]
where $\mathcal{P}_i^{(K)}(c,r)$ contains the corresponding spatial-token indices. Feature selection maximizes AUROC on the discovery set:
\[
i^\star
=
\operatorname*{arg\,max}_{1\le i\le m}
\operatorname{AUROC}_{D_{\mathrm{disc}}}
\left(s_i^{(K)},y\right).
\]
The selected feature is then fixed and evaluated on the held-out set $D_{\mathrm{test}}$ using the same pooling rule, without fitting an additional classifier or decision threshold. Algorithm~\ref{alg:nudity-feature-discovery} summarizes the procedure.

\begin{algorithm}[t]
\caption{Label-guided single-feature discovery and readout}
\label{alg:nudity-feature-discovery}
\begin{algorithmic}[1]

\Statex \textbf{Discovery:} Binary-labeled set $D_{\mathrm{disc}}$
\Statex \textbf{Evaluation:} Held-out set $D_{\mathrm{test}}$
\Statex \textbf{Fixed:} Generator $G$, SAE encoder $E_{\mathrm{SAE}}$, input operator $\mathcal{T}$, layer $\ell$, and pooling size $K$

\For{each sample $(c,r,y)\in D_{\mathrm{disc}}$}
    \State Generate with $G$ using prompt $c$ and seed $r$
    \State Capture residual activations $\{\mathbf{h}_{\ell,p}\}_p$
    \State Encode $\mathbf{z}_p\gets E_{\mathrm{SAE}}(\mathcal{T}(\mathbf{h}_{\ell,p}))$ for each token $p$
    \State Compute $s_i^{(K)}(c,r)$ for every feature $i$
\EndFor

\State $i^\star\gets\operatorname*{arg\,max}_{1\le i\le m}\operatorname{AUROC}_{D_{\mathrm{disc}}}(s_i^{(K)},y)$

\For{each sample $(c,r,y)\in D_{\mathrm{test}}$}
    \State Generate with $G$ using prompt $c$ and seed $r$
    \State Capture residual activations $\{\mathbf{h}_{\ell,p}\}_p$
    \State Encode $\mathbf{z}_p\gets E_{\mathrm{SAE}}(\mathcal{T}(\mathbf{h}_{\ell,p}))$ for each token $p$
    \State Compute $s_{i^\star}^{(K)}(c,r)$
\EndFor

\Statex \textbf{Output:} Selected feature $i^\star$ and held-out $\operatorname{AUROC}_{D_{\mathrm{test}}}(s_{i^\star}^{(K)},y)$

\end{algorithmic}
\end{algorithm}

\begin{table}[t]
\centering
\caption{
\textbf{Fixed configuration for the single-feature readout experiment.}
Generator and SAE parameters remain frozen throughout.
}
\label{tab:app-readout-protocol}
\small
\setlength{\tabcolsep}{6pt}
\begin{tabular}{@{}ll@{}}
\toprule
Component & Setting \\
\midrule
Generator & SANA-Sprint 0.6B \\
Generation & \(1024\times1024\), one inference step, guidance 0 \\
SAE & Layer-14 mean-centered L1 \\
Dictionary size & 18,432 features \\
Spatial resolution & \(32\times32=1024\) tokens \\
Reporting pool size & \(K=16\) \\
\bottomrule
\end{tabular}
\end{table}

\subsection{Held-Out Readout and Feature Analysis}

Discovery selects feature 4867. With this feature fixed, its image-level
activation achieves an AUROC of 0.980 on 118 held-out images, comprising
35 nude and 83 non-nude examples. To examine the selected feature beyond
this aggregate result, Figure~\ref{fig:app-nudity-feature-analysis} provides
its high-activation feature card and evaluates sensitivity to the spatial
pooling size. Feature 4867 remains selected for
\(K\in\{4,16,64,256,1024\}\), with held-out AUROC varying only from
0.979 to 0.980. We use \(K=16\) as the reporting default.

\newsavebox{\nudityksweeptable}
\newlength{\nuditypanelheight}

\savebox{\nudityksweeptable}{%
    \resizebox{0.3864\textwidth}{!}{%
        \renewcommand{\arraystretch}{1.35}%
        \begin{tabular}{ccc}
        \toprule
        \(K\) & Selected feature & Test AUROC \\
        \midrule
        4     & 4867 & 0.980 \\
        16    & 4867 & 0.980 \\
        64    & 4867 & 0.979 \\
        256   & 4867 & 0.979 \\
        1024  & 4867 & 0.979 \\
        \bottomrule
        \end{tabular}%
    }%
}

\setlength{\nuditypanelheight}{%
    \dimexpr\ht\nudityksweeptable+\dp\nudityksweeptable\relax
}

\begin{figure*}[!htbp]
    \centering

    \begin{subfigure}[c]{0.53\textwidth}
        \centering
        \includegraphics[
            height=\nuditypanelheight
        ]{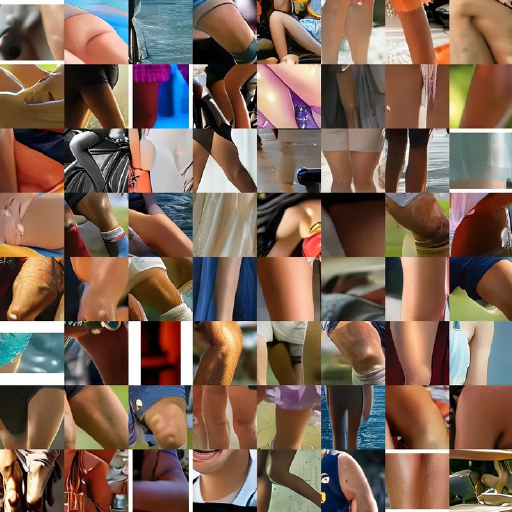}
        \caption{}
        \label{fig:app-nudity-feature-card}
    \end{subfigure}
    \hfill
    \begin{subfigure}[c]{0.42\textwidth}
        \centering
        \usebox{\nudityksweeptable}
        \caption{}
        \label{fig:app-nudity-k-sweep}
    \end{subfigure}

    \caption{
    \textbf{Feature semantics and aggregation robustness.}
    \textbf{(a)} High-activation patches for feature 4867.
    \textbf{(b)} Sensitivity to spatial aggregation. Feature 4867 is
    consistently recovered across pooling sizes, with held-out AUROC
    between 0.979 and 0.980.
    }
    \label{fig:app-nudity-feature-analysis}
\end{figure*}

\subsection{Qualitative Readout Examples}
\label{app:nudity-qualitative}

We show representative held-out examples. Each example shows the generated image and its feature-4867 activation overlay.

\begin{figure}[!t]
    \centering

\begin{minipage}[t]{0.49\textwidth}
    \centering
    \makebox[\linewidth][c]{%
        \includegraphics[
            height=0.105\textheight,
            keepaspectratio
        ]{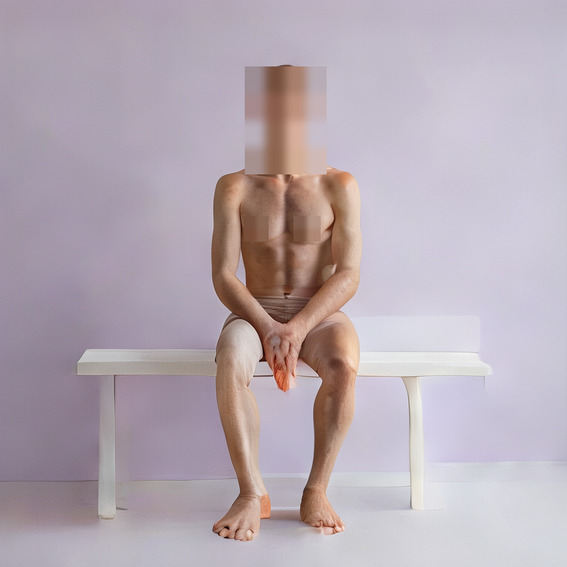}%
        \hspace{2pt}%
        \includegraphics[
            height=0.105\textheight,
            keepaspectratio
        ]{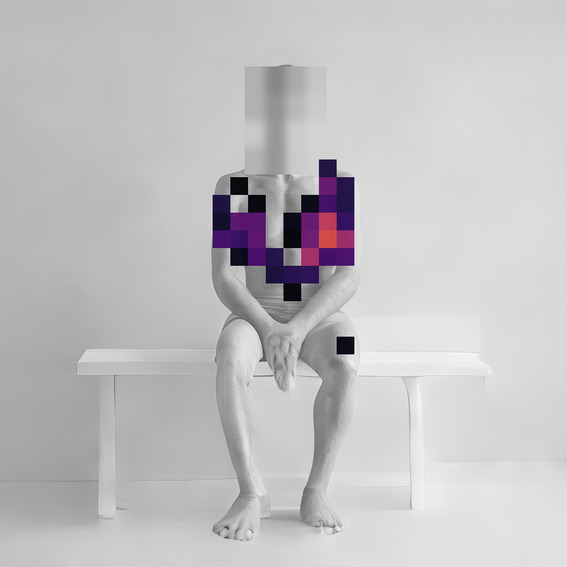}%
    }%
    \if\relax\detokenize{Nude}\relax
    \else
        \par\vspace{0.5pt}
        {\scriptsize Nude}
    \fi
\end{minipage}%
\hfill%
\begin{minipage}[t]{0.49\textwidth}
    \centering
    \makebox[\linewidth][c]{%
        \includegraphics[
            height=0.105\textheight,
            keepaspectratio
        ]{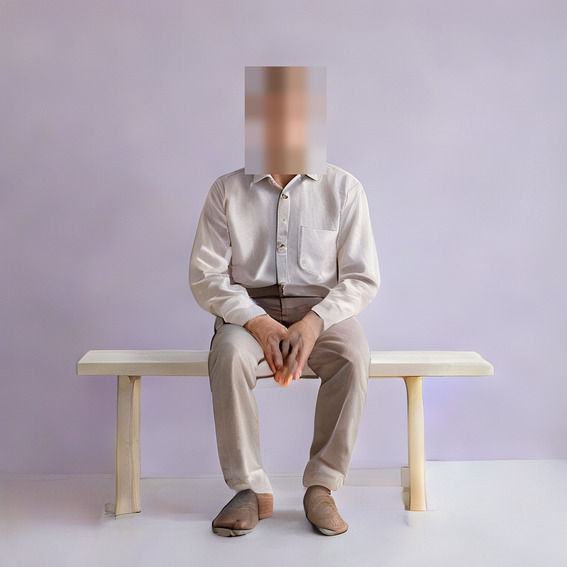}%
        \hspace{2pt}%
        \includegraphics[
            height=0.105\textheight,
            keepaspectratio
        ]{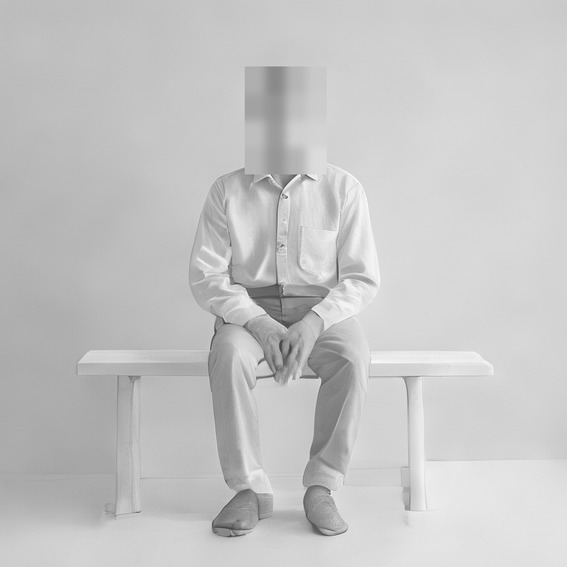}%
    }%
    \if\relax\detokenize{Clothed}\relax
    \else
        \par\vspace{0.5pt}
        {\scriptsize Clothed}
    \fi
\end{minipage}%

    \par\vspace{2pt}

\begin{minipage}[t]{0.49\textwidth}
    \centering
    \makebox[\linewidth][c]{%
        \includegraphics[
            height=0.105\textheight,
            keepaspectratio
        ]{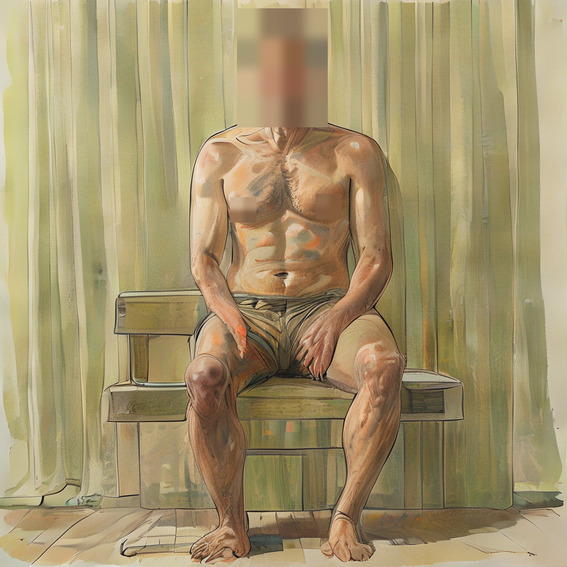}%
        \hspace{2pt}%
        \includegraphics[
            height=0.105\textheight,
            keepaspectratio
        ]{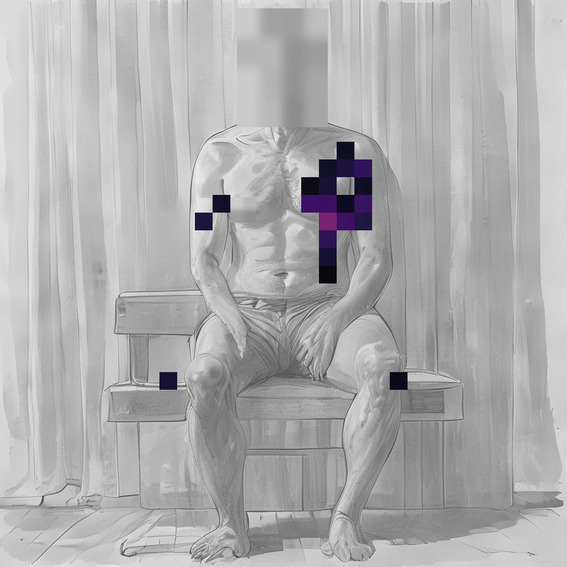}%
    }%
    \if\relax\detokenize{Nude}\relax
    \else
        \par\vspace{0.5pt}
        {\scriptsize Nude}
    \fi
\end{minipage}%
\hfill%
\begin{minipage}[t]{0.49\textwidth}
    \centering
    \makebox[\linewidth][c]{%
        \includegraphics[
            height=0.105\textheight,
            keepaspectratio
        ]{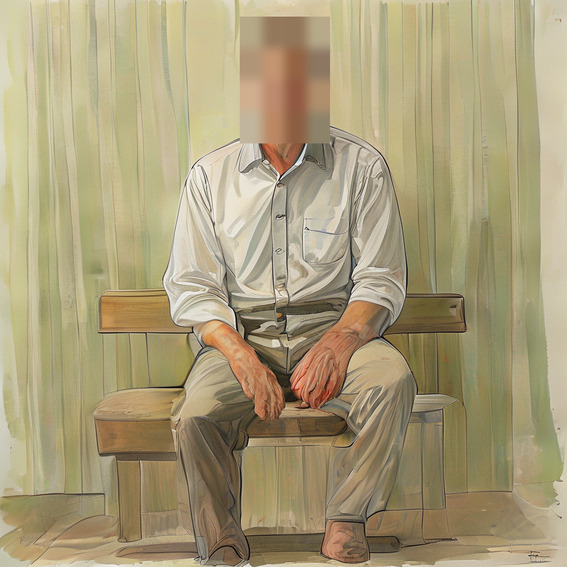}%
        \hspace{2pt}%
        \includegraphics[
            height=0.105\textheight,
            keepaspectratio
        ]{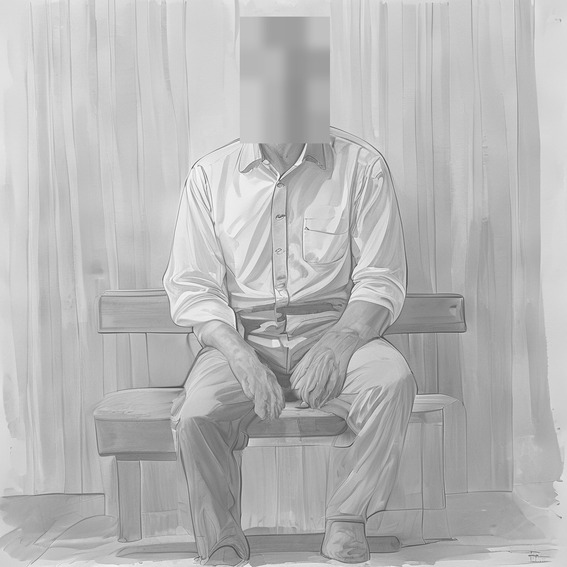}%
    }%
    \if\relax\detokenize{Clothed}\relax
    \else
        \par\vspace{0.5pt}
        {\scriptsize Clothed}
    \fi
\end{minipage}%

    \par\vspace{2pt}

\begin{minipage}[t]{0.49\textwidth}
    \centering
    \makebox[\linewidth][c]{%
        \includegraphics[
            height=0.105\textheight,
            keepaspectratio
        ]{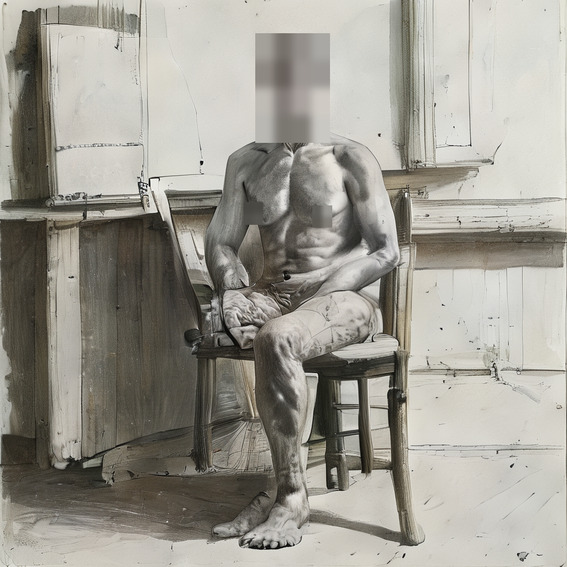}%
        \hspace{2pt}%
        \includegraphics[
            height=0.105\textheight,
            keepaspectratio
        ]{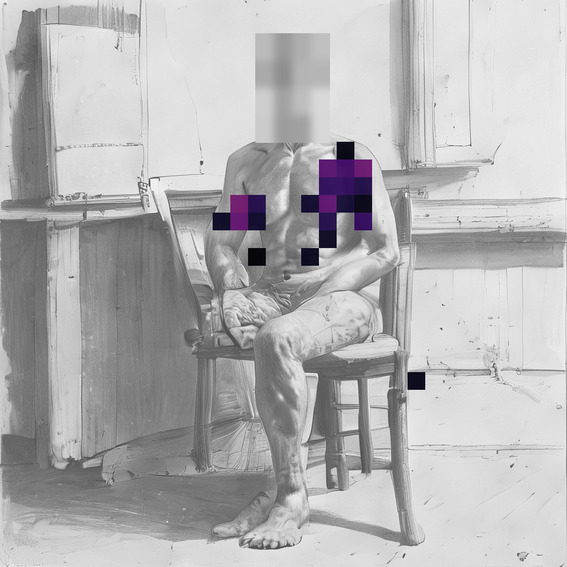}%
    }%
    \if\relax\detokenize{Nude}\relax
    \else
        \par\vspace{0.5pt}
        {\scriptsize Nude}
    \fi
\end{minipage}%
\hfill%
\begin{minipage}[t]{0.49\textwidth}
    \centering
    \makebox[\linewidth][c]{%
        \includegraphics[
            height=0.105\textheight,
            keepaspectratio
        ]{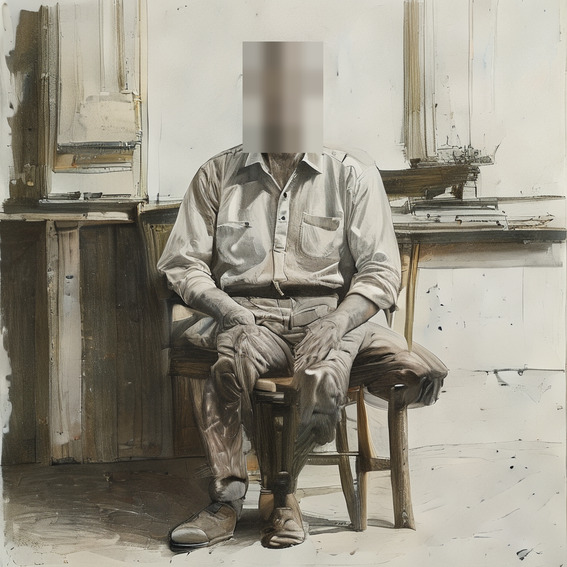}%
        \hspace{2pt}%
        \includegraphics[
            height=0.105\textheight,
            keepaspectratio
        ]{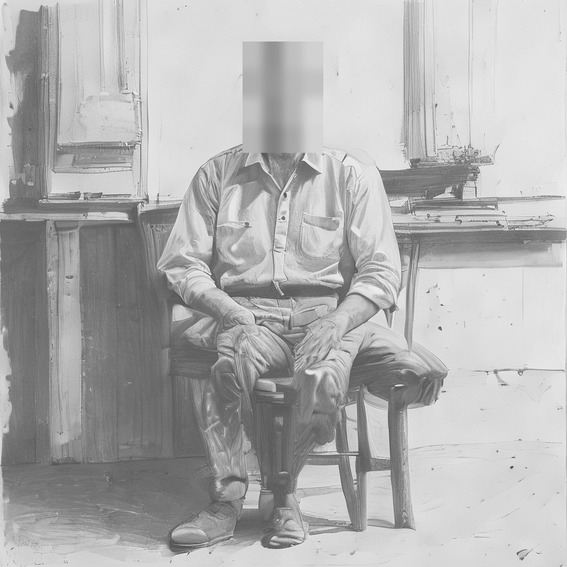}%
    }%
    \if\relax\detokenize{Clothed}\relax
    \else
        \par\vspace{0.5pt}
        {\scriptsize Clothed}
    \fi
\end{minipage}%

    \par\vspace{2pt}

\begin{minipage}[t]{0.49\textwidth}
    \centering
    \makebox[\linewidth][c]{%
        \includegraphics[
            height=0.105\textheight,
            keepaspectratio
        ]{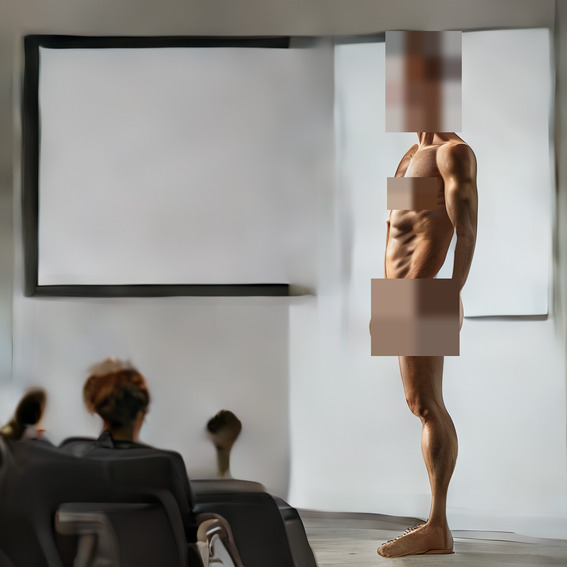}%
        \hspace{2pt}%
        \includegraphics[
            height=0.105\textheight,
            keepaspectratio
        ]{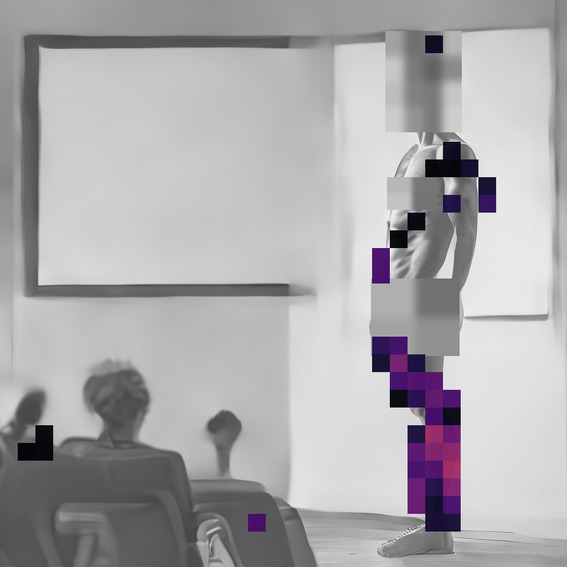}%
    }%
    \if\relax\detokenize{Nude}\relax
    \else
        \par\vspace{0.5pt}
        {\scriptsize Nude}
    \fi
\end{minipage}%
\hfill%
\begin{minipage}[t]{0.49\textwidth}
    \centering
    \makebox[\linewidth][c]{%
        \includegraphics[
            height=0.105\textheight,
            keepaspectratio
        ]{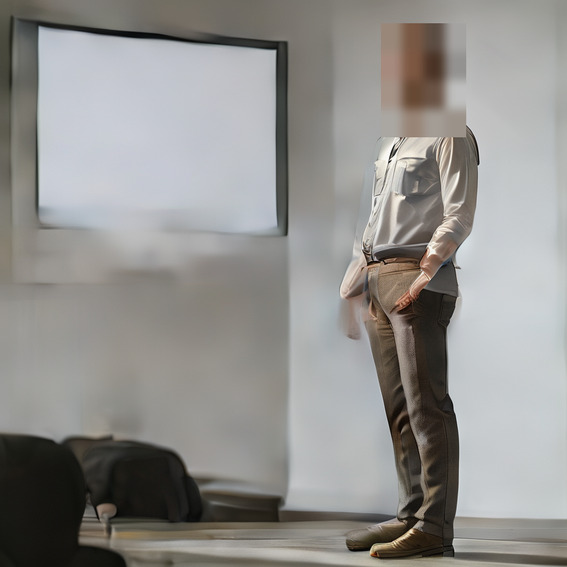}%
        \hspace{2pt}%
        \includegraphics[
            height=0.105\textheight,
            keepaspectratio
        ]{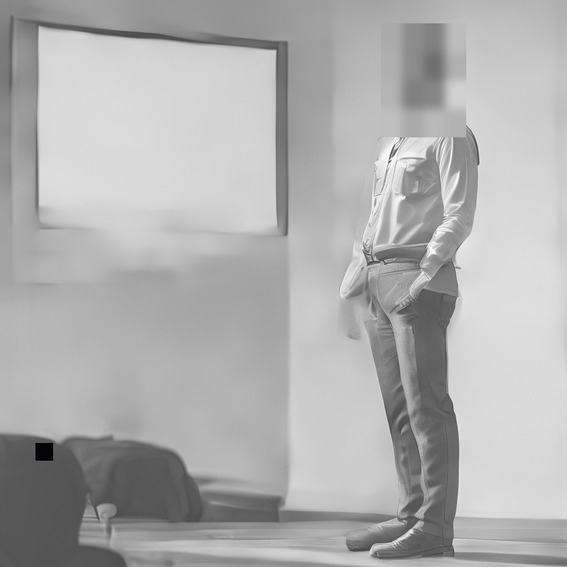}%
    }%
    \if\relax\detokenize{Clothed}\relax
    \else
        \par\vspace{0.5pt}
        {\scriptsize Clothed}
    \fi
\end{minipage}%

    \par\vspace{2pt}

\begin{minipage}[t]{0.49\textwidth}
    \centering
    \makebox[\linewidth][c]{%
        \includegraphics[
            height=0.105\textheight,
            keepaspectratio
        ]{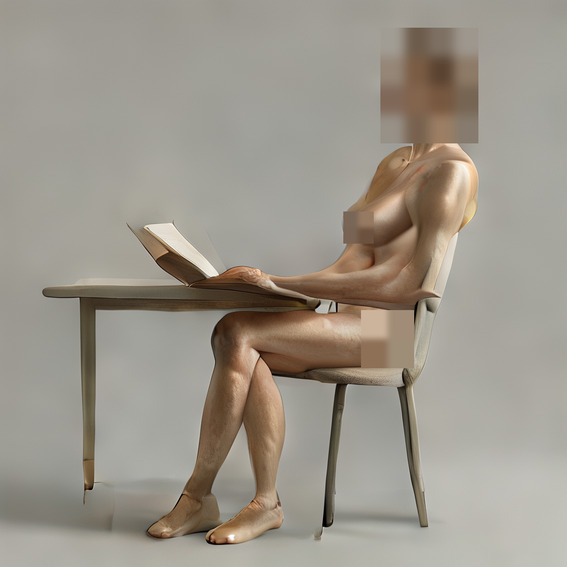}%
        \hspace{2pt}%
        \includegraphics[
            height=0.105\textheight,
            keepaspectratio
        ]{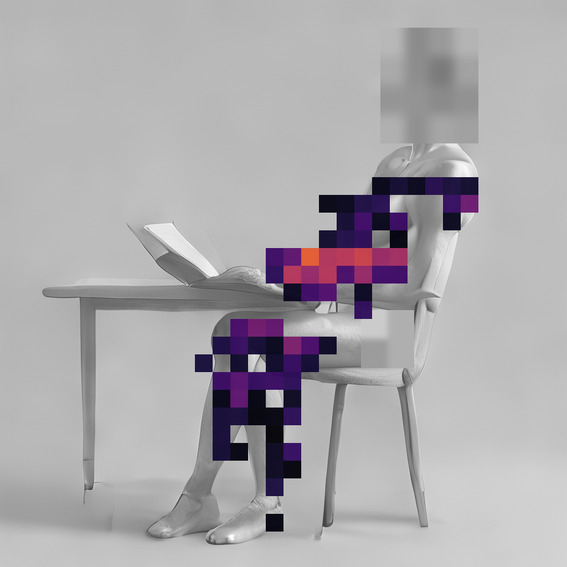}%
    }%
    \if\relax\detokenize{Nude}\relax
    \else
        \par\vspace{0.5pt}
        {\scriptsize Nude}
    \fi
\end{minipage}%
\hfill%
\begin{minipage}[t]{0.49\textwidth}
    \centering
    \makebox[\linewidth][c]{%
        \includegraphics[
            height=0.105\textheight,
            keepaspectratio
        ]{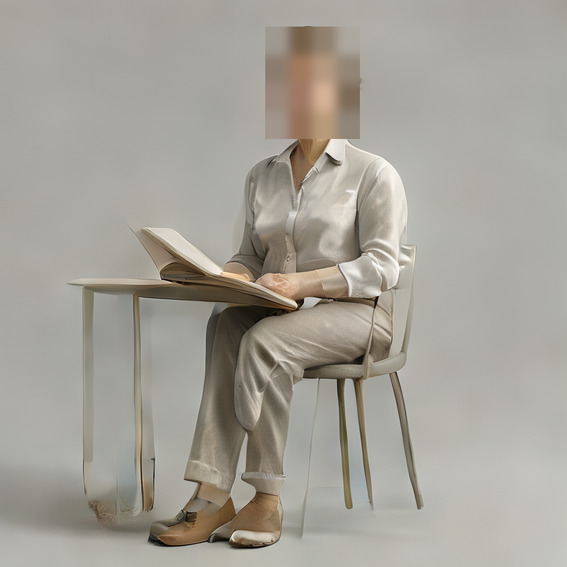}%
        \hspace{2pt}%
        \includegraphics[
            height=0.105\textheight,
            keepaspectratio
        ]{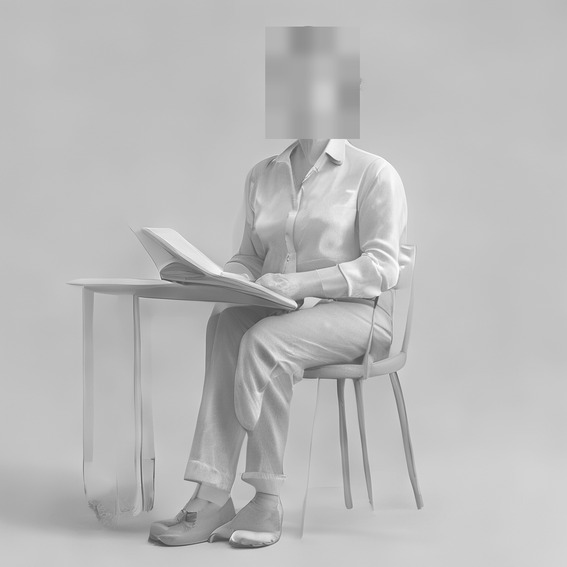}%
    }%
    \if\relax\detokenize{Clothed}\relax
    \else
        \par\vspace{0.5pt}
        {\scriptsize Clothed}
    \fi
\end{minipage}%

    \par\vspace{2pt}

\begin{minipage}[t]{0.49\textwidth}
    \centering
    \makebox[\linewidth][c]{%
        \includegraphics[
            height=0.105\textheight,
            keepaspectratio
        ]{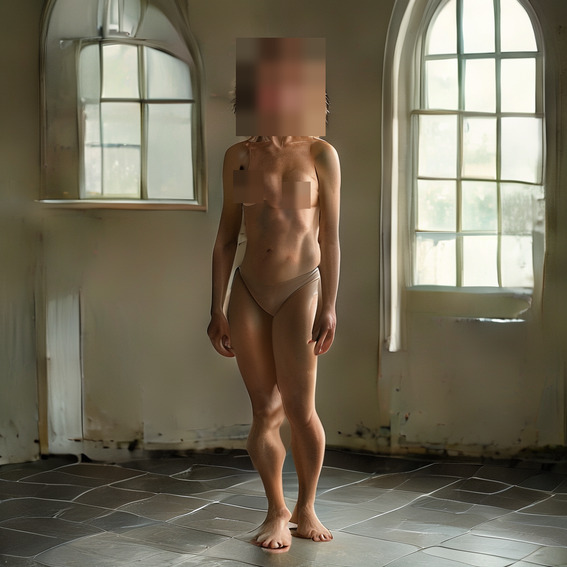}%
        \hspace{2pt}%
        \includegraphics[
            height=0.105\textheight,
            keepaspectratio
        ]{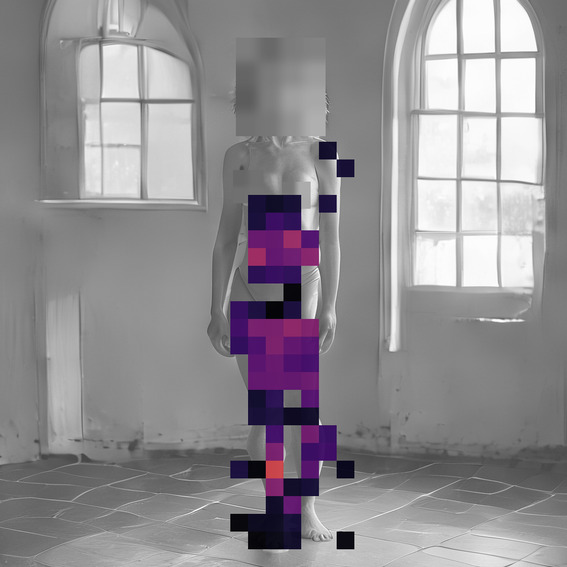}%
    }%
    \if\relax\detokenize{Nude}\relax
    \else
        \par\vspace{0.5pt}
        {\scriptsize Nude}
    \fi
\end{minipage}%
\hfill%
\begin{minipage}[t]{0.49\textwidth}
    \centering
    \makebox[\linewidth][c]{%
        \includegraphics[
            height=0.105\textheight,
            keepaspectratio
        ]{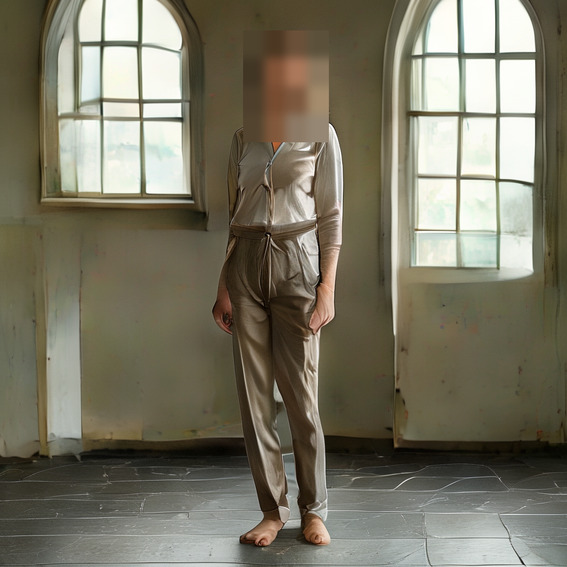}%
        \hspace{2pt}%
        \includegraphics[
            height=0.105\textheight,
            keepaspectratio
        ]{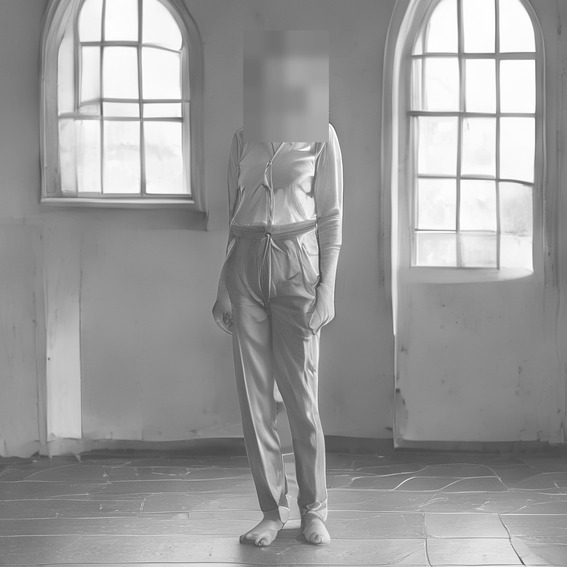}%
    }%
    \if\relax\detokenize{Clothed}\relax
    \else
        \par\vspace{0.5pt}
        {\scriptsize Clothed}
    \fi
\end{minipage}%

    \vspace{-2pt}
    \caption{
        \textbf{Matched nude-clothed examples.}
        Each adjacent nude-clothed pair shares the same scene and seed.
        Each example shows the generated image (left) and its
        feature-4867 activation overlay (right).
    }
    \label{fig:app-nudity-matched-examples}
\end{figure}

\begin{figure}[!t]
    \centering

\begin{minipage}[t]{0.49\textwidth}
    \centering
    \makebox[\linewidth][c]{%
        \includegraphics[
            height=0.105\textheight,
            keepaspectratio
        ]{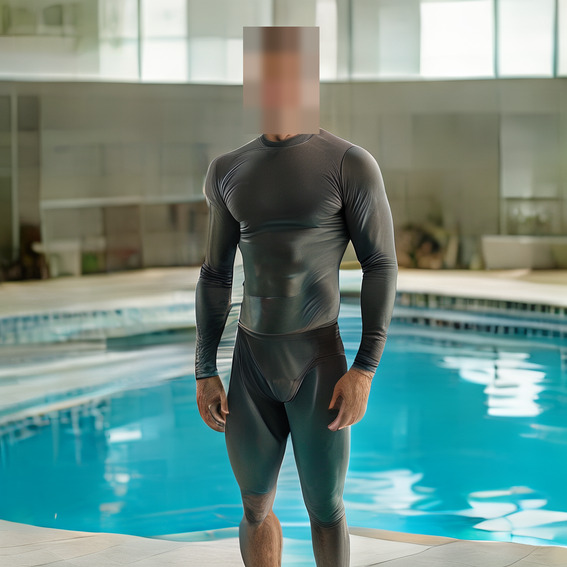}%
        \hspace{2pt}%
        \includegraphics[
            height=0.105\textheight,
            keepaspectratio
        ]{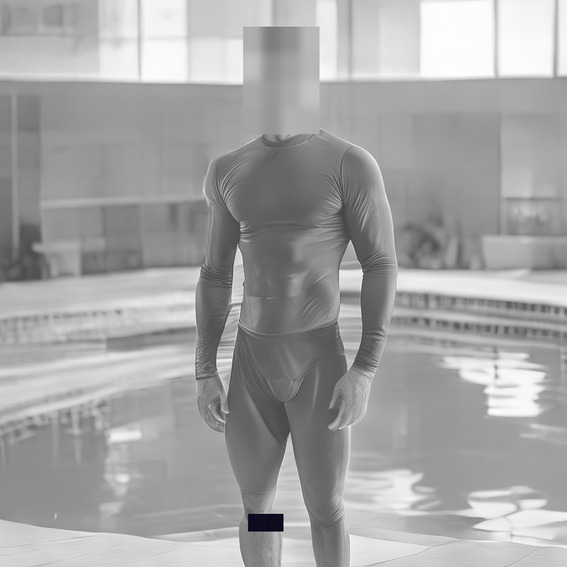}%
    }%
    \if\relax\detokenize{}\relax
    \else
        \par\vspace{0.5pt}
        {\scriptsize }
    \fi
\end{minipage}%
\hfill%
\begin{minipage}[t]{0.49\textwidth}
    \centering
    \makebox[\linewidth][c]{%
        \includegraphics[
            height=0.105\textheight,
            keepaspectratio
        ]{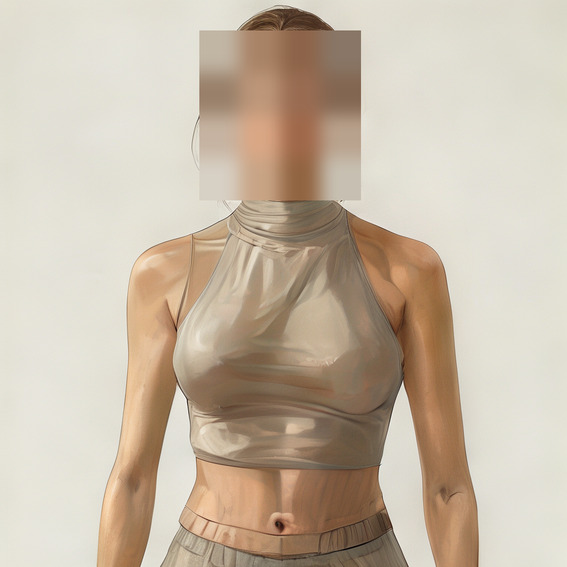}%
        \hspace{2pt}%
        \includegraphics[
            height=0.105\textheight,
            keepaspectratio
        ]{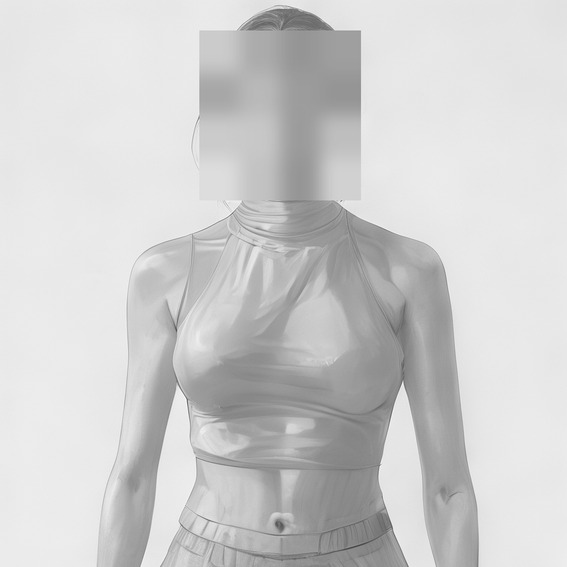}%
    }%
    \if\relax\detokenize{}\relax
    \else
        \par\vspace{0.5pt}
        {\scriptsize }
    \fi
\end{minipage}%

    \par\vspace{2pt}

\begin{minipage}[t]{0.49\textwidth}
    \centering
    \makebox[\linewidth][c]{%
        \includegraphics[
            height=0.105\textheight,
            keepaspectratio
        ]{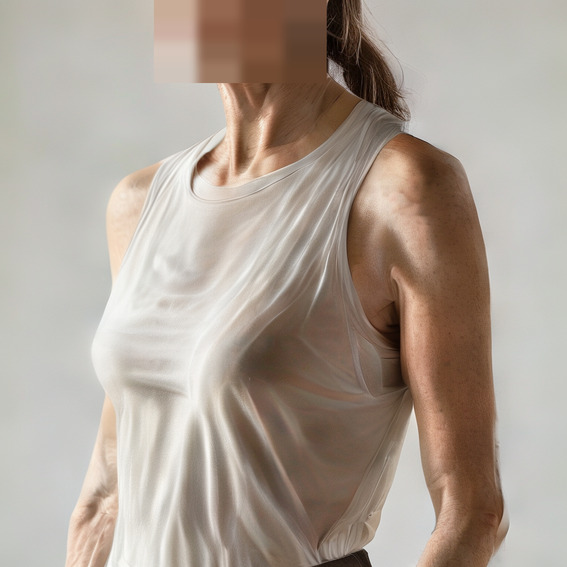}%
        \hspace{2pt}%
        \includegraphics[
            height=0.105\textheight,
            keepaspectratio
        ]{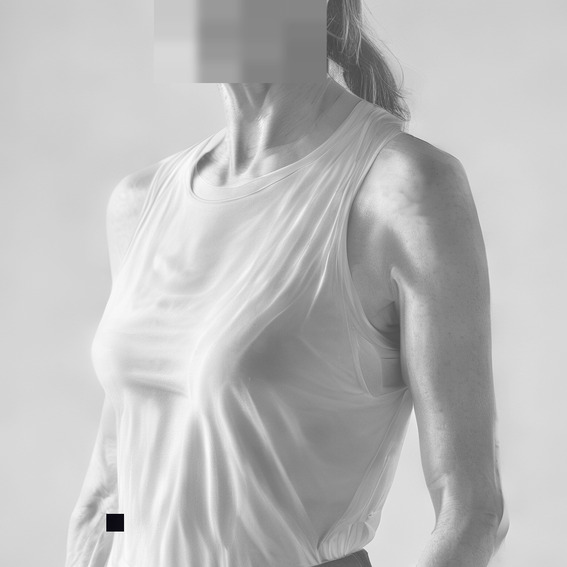}%
    }%
    \if\relax\detokenize{}\relax
    \else
        \par\vspace{0.5pt}
        {\scriptsize }
    \fi
\end{minipage}%
\hfill%
\begin{minipage}[t]{0.49\textwidth}
    \centering
    \makebox[\linewidth][c]{%
        \includegraphics[
            height=0.105\textheight,
            keepaspectratio
        ]{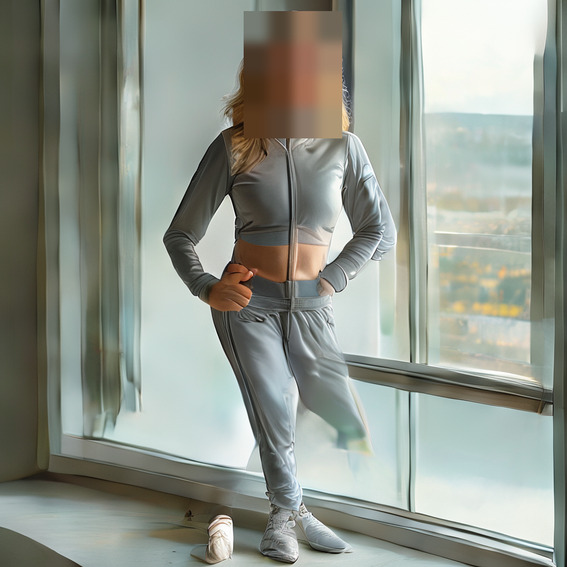}%
        \hspace{2pt}%
        \includegraphics[
            height=0.105\textheight,
            keepaspectratio
        ]{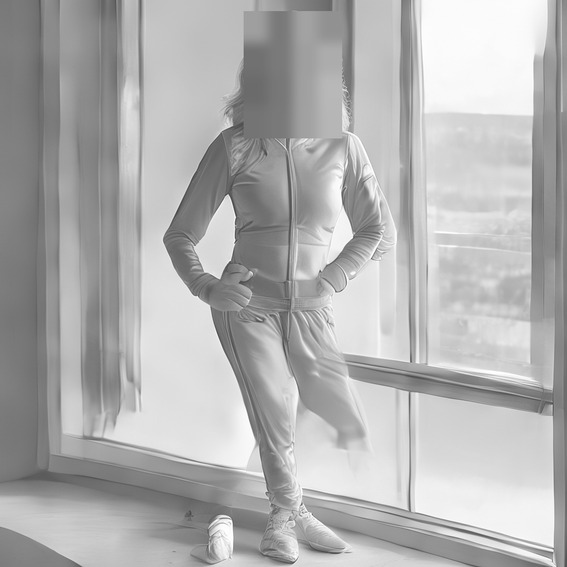}%
    }%
    \if\relax\detokenize{}\relax
    \else
        \par\vspace{0.5pt}
        {\scriptsize }
    \fi
\end{minipage}%

    \par\vspace{2pt}

\begin{minipage}[t]{0.49\textwidth}
    \centering
    \makebox[\linewidth][c]{%
        \includegraphics[
            height=0.105\textheight,
            keepaspectratio
        ]{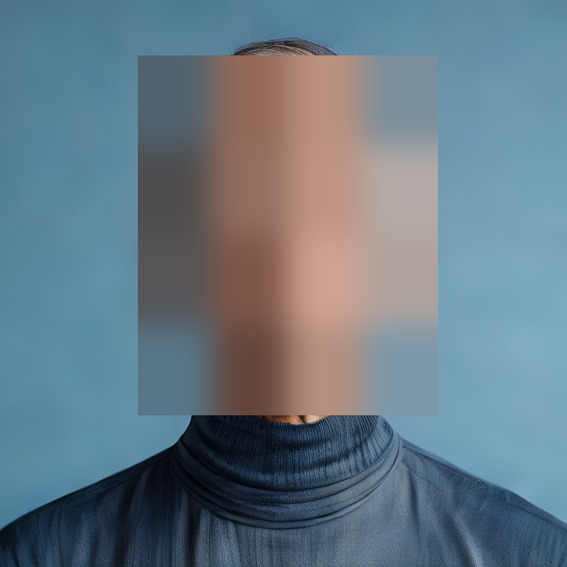}%
        \hspace{2pt}%
        \includegraphics[
            height=0.105\textheight,
            keepaspectratio
        ]{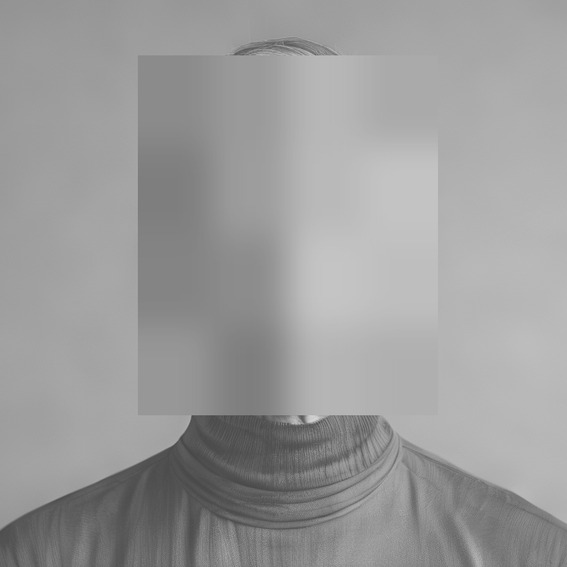}%
    }%
    \if\relax\detokenize{}\relax
    \else
        \par\vspace{0.5pt}
        {\scriptsize }
    \fi
\end{minipage}%
\hfill%
\begin{minipage}[t]{0.49\textwidth}
    \centering
    \makebox[\linewidth][c]{%
        \includegraphics[
            height=0.105\textheight,
            keepaspectratio
        ]{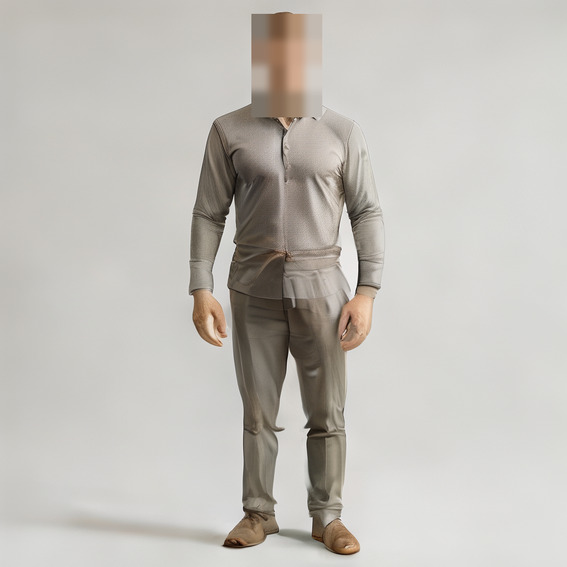}%
        \hspace{2pt}%
        \includegraphics[
            height=0.105\textheight,
            keepaspectratio
        ]{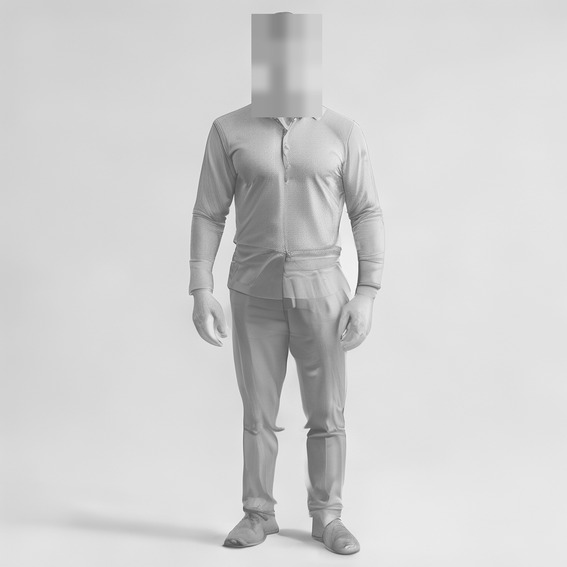}%
    }%
    \if\relax\detokenize{}\relax
    \else
        \par\vspace{0.5pt}
        {\scriptsize }
    \fi
\end{minipage}%

    \par\vspace{2pt}

\begin{minipage}[t]{0.49\textwidth}
    \centering
    \makebox[\linewidth][c]{%
        \includegraphics[
            height=0.105\textheight,
            keepaspectratio
        ]{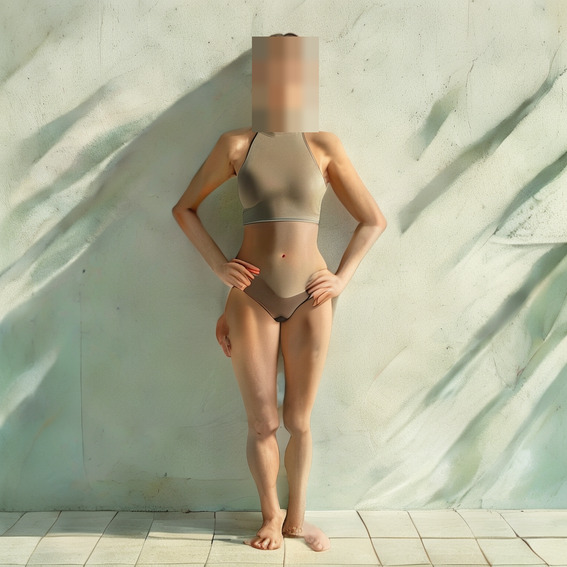}%
        \hspace{2pt}%
        \includegraphics[
            height=0.105\textheight,
            keepaspectratio
        ]{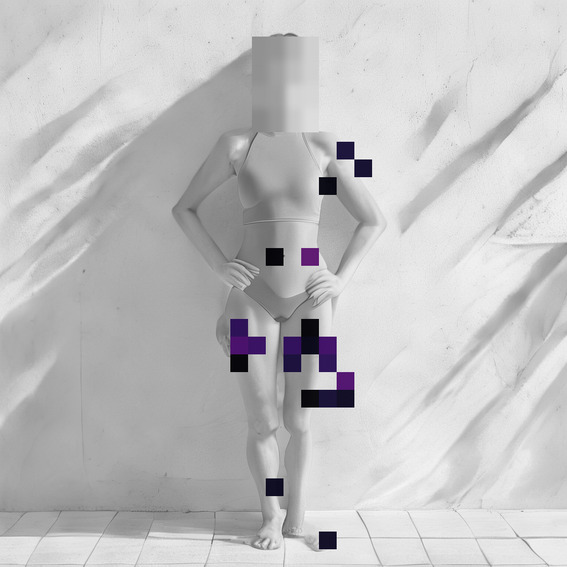}%
    }%
    \if\relax\detokenize{}\relax
    \else
        \par\vspace{0.5pt}
        {\scriptsize }
    \fi
\end{minipage}%
\hfill%
\begin{minipage}[t]{0.49\textwidth}
    \centering
    \makebox[\linewidth][c]{%
        \includegraphics[
            height=0.105\textheight,
            keepaspectratio
        ]{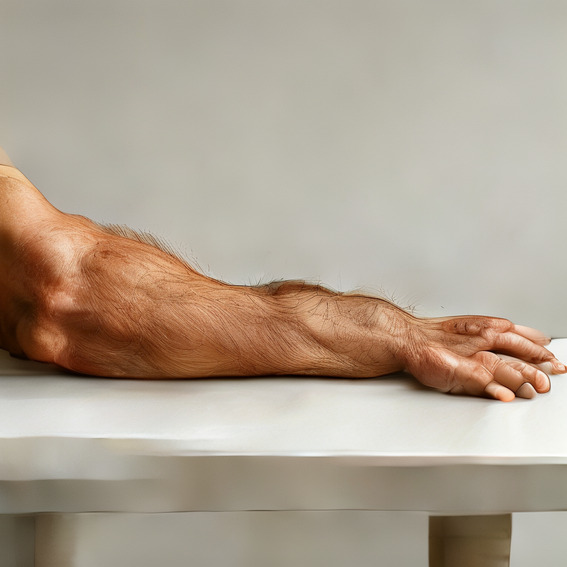}%
        \hspace{2pt}%
        \includegraphics[
            height=0.105\textheight,
            keepaspectratio
        ]{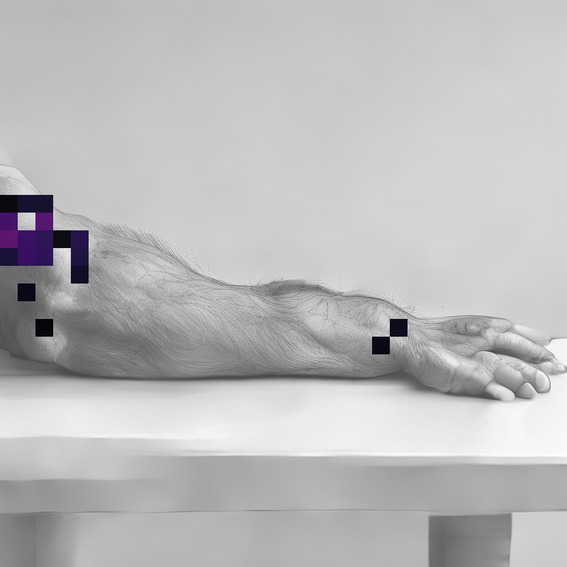}%
    }%
    \if\relax\detokenize{}\relax
    \else
        \par\vspace{0.5pt}
        {\scriptsize }
    \fi
\end{minipage}%

    \vspace{-2pt}
    \caption{
        \textbf{Hard-negative examples.}
        Representative non-nude examples with potentially confounding content.
        Each example shows the generated image (left) and its
        feature-4867 activation overlay (right).
    }
    \label{fig:app-nudity-hard-negative-examples}
\end{figure}

\end{document}

%% file: math_commands.tex
\usepackage{amsmath,amsfonts,bm}

\def\eqref#1{equation~\ref{#1}}

\def\1{\bm{1}}

\DeclareMathAlphabet{\mathsfit}{\encodingdefault}{\sfdefault}{m}{sl}
\SetMathAlphabet{\mathsfit}{bold}{\encodingdefault}{\sfdefault}{bx}{n}

%% file: figs_compressed/fig1_teaser/fig1_teaser.tex
\begingroup

\captionsetup{type=figure}
\captionsetup[subfigure]{
  position=bottom,
  font=small,
  labelfont=normalfont,
  textfont=normalfont,
  labelformat=parens,
  labelsep=space,
  justification=centering,
  singlelinecheck=true,
  skip=4pt
}

\setlength{\parindent}{0pt}
\setlength{\parskip}{0pt}

\begin{subfigure}[t]{\linewidth}
  \vspace{0pt}
  \input{figs_compressed/fig1_teaser/panels/a_single_feature.tex}
\end{subfigure}\par

\vspace{10pt}

\begin{subfigure}[t]{\linewidth}
  \vspace{0pt}
  \input{figs_compressed/fig1_teaser/panels/b_query_features.tex}
\end{subfigure}\par

\caption{\textbf{Query-driven single-feature steering with D-Scope.} \textbf{(a)} \textbf{Query-driven direct retrieval.} Direct retrieval with \textit{rose} selects one decoder direction, held fixed as the steering strength $\rho$ varies. \textbf{(b)} \textbf{Query-driven contrastive retrieval.} Each color-specific query is contrasted with \textit{dress} to select one decoder direction for the corresponding output. \textit{Source} denotes generation without intervention.}
\label{fig:teaser}

\endgroup

%% file: figs_compressed/fig1_teaser/panels/a_single_feature.tex
\noindent\makebox[\linewidth][s]{%
  \begin{minipage}[t]{0.119\linewidth}
    \centering
    \includegraphics[width=\linewidth]{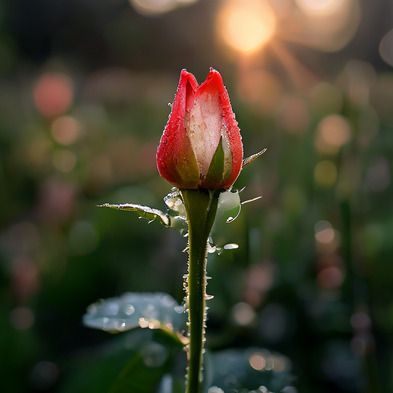}\par
    \vspace{2.5pt}%
    {\footnotesize Source\par}%
  \end{minipage}%
  \hfill%
  \begin{minipage}[t]{0.119\linewidth}
    \centering
    \includegraphics[width=\linewidth]{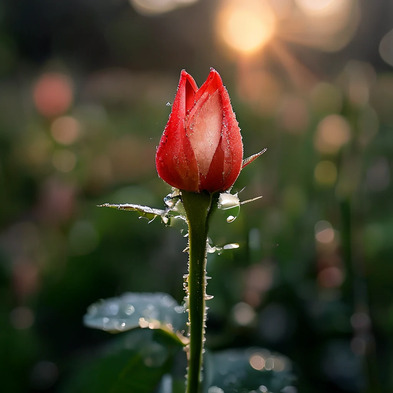}\par
    \vspace{2.5pt}%
    {\footnotesize \(\rho\)\,=\,0.1\par}%
  \end{minipage}%
  \hfill%
  \begin{minipage}[t]{0.119\linewidth}
    \centering
    \includegraphics[width=\linewidth]{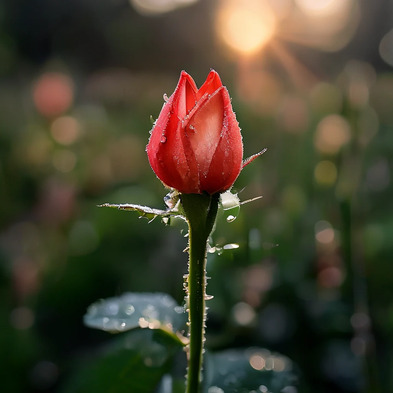}\par
    \vspace{2.5pt}%
    {\footnotesize \(\rho\)\,=\,0.2\par}%
  \end{minipage}%
  \hfill%
  \begin{minipage}[t]{0.119\linewidth}
    \centering
    \includegraphics[width=\linewidth]{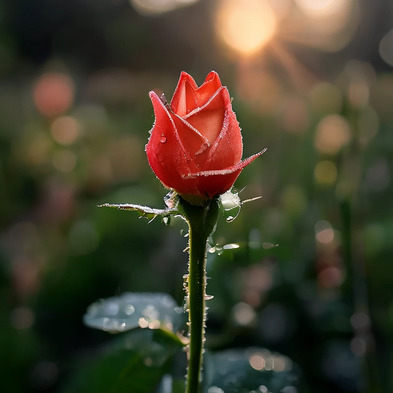}\par
    \vspace{2.5pt}%
    {\footnotesize \(\rho\)\,=\,0.3\par}%
  \end{minipage}%
  \hfill%
  \begin{minipage}[t]{0.119\linewidth}
    \centering
    \includegraphics[width=\linewidth]{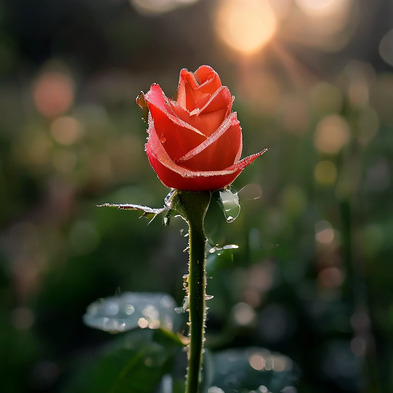}\par
    \vspace{2.5pt}%
    {\footnotesize \(\rho\)\,=\,0.5\par}%
  \end{minipage}%
  \hfill%
  \begin{minipage}[t]{0.119\linewidth}
    \centering
    \includegraphics[width=\linewidth]{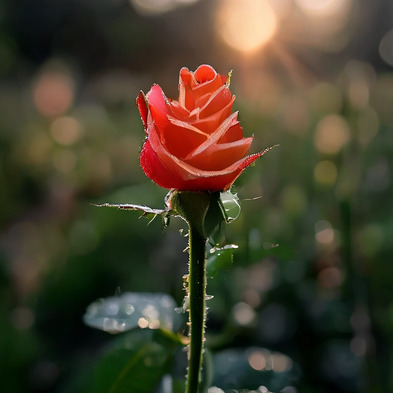}\par
    \vspace{2.5pt}%
    {\footnotesize \(\rho\)\,=\,0.75\par}%
  \end{minipage}%
  \hfill%
  \begin{minipage}[t]{0.119\linewidth}
    \centering
    \includegraphics[width=\linewidth]{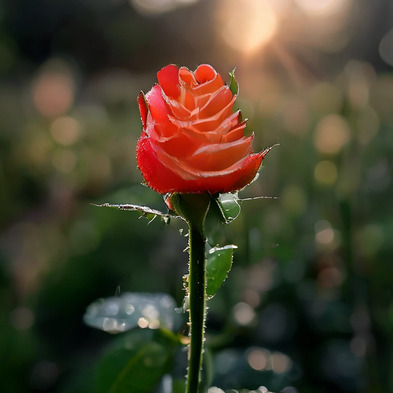}\par
    \vspace{2.5pt}%
    {\footnotesize \(\rho\)\,=\,1.0\par}%
  \end{minipage}%
  \hfill%
  \begin{minipage}[t]{0.119\linewidth}
    \centering
    \includegraphics[width=\linewidth]{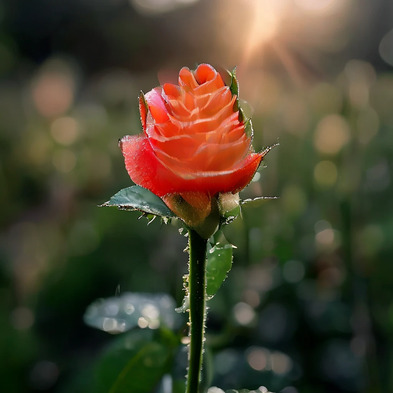}\par
    \vspace{2.5pt}%
    {\footnotesize \(\rho\)\,=\,1.5\par}%
  \end{minipage}%
}\par
\caption{Varying strength for a fixed target query}
\label{fig:teaser-a}

%% file: figs_compressed/fig1_teaser/panels/b_query_features.tex
\noindent\makebox[\linewidth][s]{%
  \begin{minipage}[t]{0.119\linewidth}
    \centering
    \includegraphics[width=\linewidth]{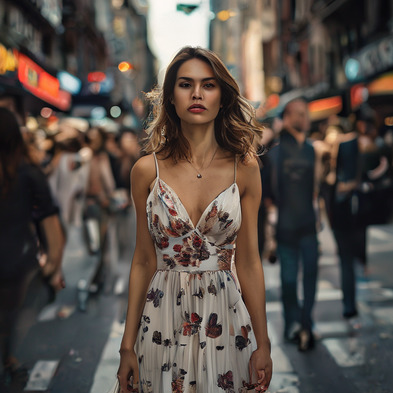}\par
    \vspace{2.5pt}%
    {\footnotesize Source\par}%
  \end{minipage}%
  \hfill%
  \begin{minipage}[t]{0.119\linewidth}
    \centering
    \includegraphics[width=\linewidth]{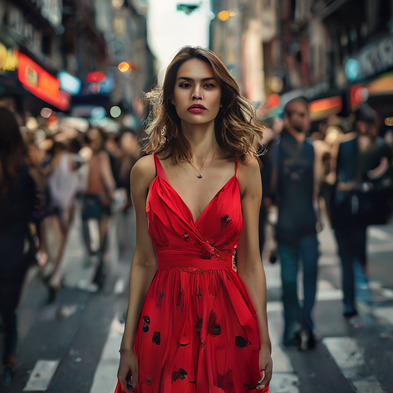}\par
    \vspace{2.5pt}%
    {\footnotesize red dress\par}%
  \end{minipage}%
  \hfill%
  \begin{minipage}[t]{0.119\linewidth}
    \centering
    \includegraphics[width=\linewidth]{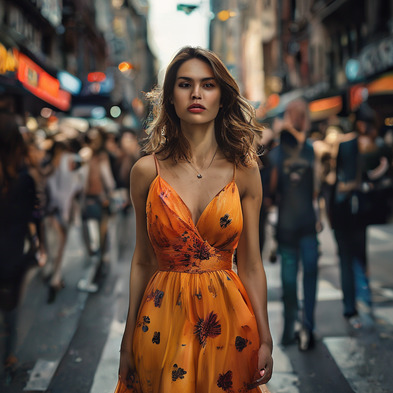}\par
    \vspace{2.5pt}%
    {\footnotesize orange dress\par}%
  \end{minipage}%
  \hfill%
  \begin{minipage}[t]{0.119\linewidth}
    \centering
    \includegraphics[width=\linewidth]{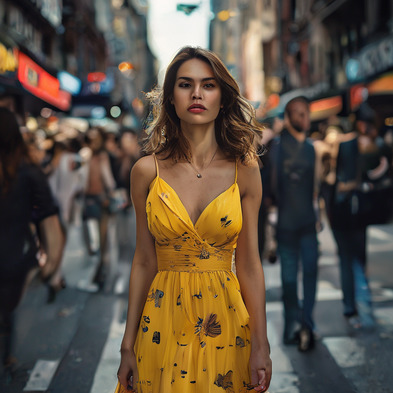}\par
    \vspace{2.5pt}%
    {\footnotesize yellow dress\par}%
  \end{minipage}%
  \hfill%
  \begin{minipage}[t]{0.119\linewidth}
    \centering
    \includegraphics[width=\linewidth]{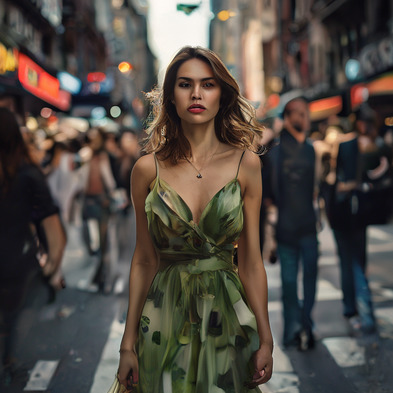}\par
    \vspace{2.5pt}%
    {\footnotesize emerald\\green dress\par}%
  \end{minipage}%
  \hfill%
  \begin{minipage}[t]{0.119\linewidth}
    \centering
    \includegraphics[width=\linewidth]{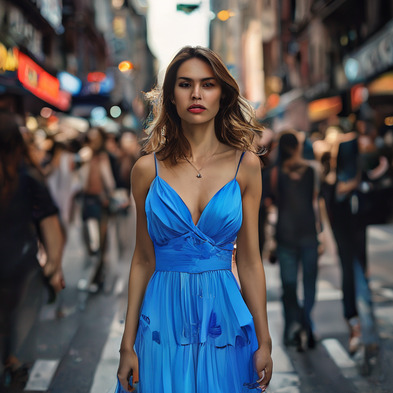}\par
    \vspace{2.5pt}%
    {\footnotesize blue dress\par}%
  \end{minipage}%
  \hfill%
  \begin{minipage}[t]{0.119\linewidth}
    \centering
    \includegraphics[width=\linewidth]{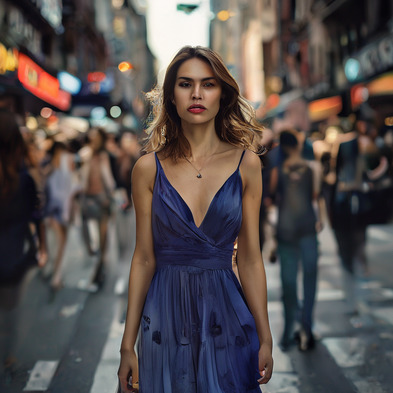}\par
    \vspace{2.5pt}%
    {\footnotesize deep indigo\\blue dress\par}%
  \end{minipage}%
  \hfill%
  \begin{minipage}[t]{0.119\linewidth}
    \centering
    \includegraphics[width=\linewidth]{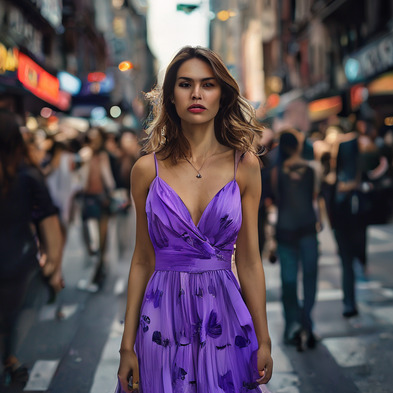}\par
    \vspace{2.5pt}%
    {\footnotesize violet dress\par}%
  \end{minipage}%
}\par
\caption{Same source image with a different target query}
\label{fig:teaser-b}

%% file: figs_compressed/fig4_benchmarking/fig4_steering.tex
\begingroup

\captionsetup{type=figure}
\captionsetup[subfigure]{position=top,labelformat=parens,labelsep=space,justification=raggedright,singlelinecheck=false,font=footnotesize,labelfont=bf,skip=2pt}
\begin{subfigure}[t]{0.575\textwidth}
  \caption{Qualitative examples}\label{fig:steering-a}
  \includegraphics[width=\linewidth]{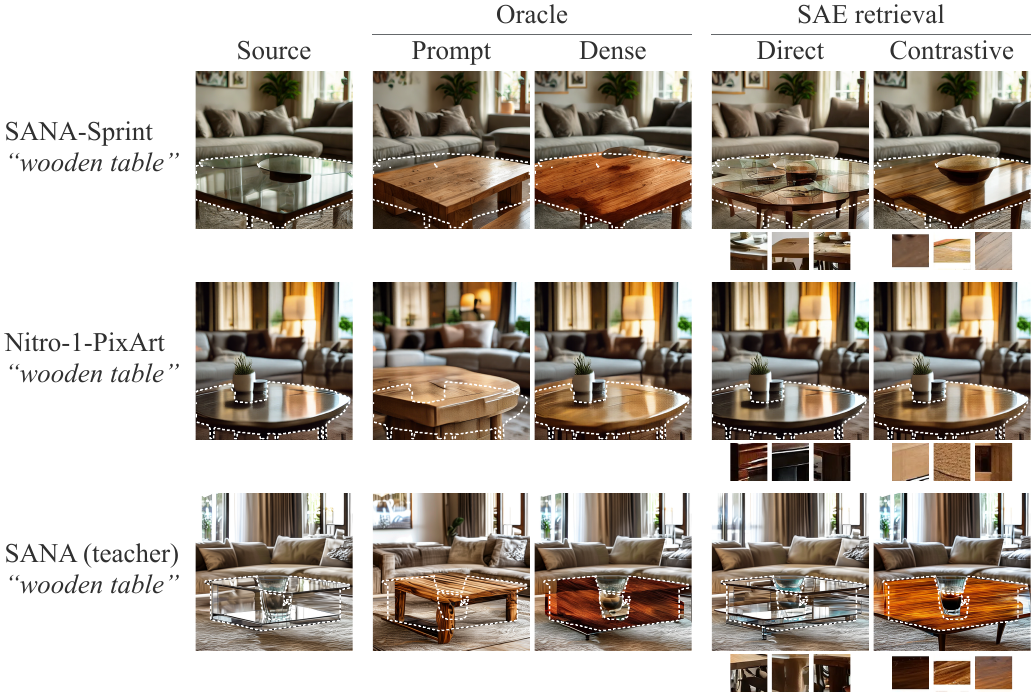}
\end{subfigure}\hfill%
\begin{minipage}[t]{0.405\textwidth}
  \begin{subfigure}[t]{\linewidth}
    \caption{Locality--gain trade-off}\label{fig:steering-b}
    \includegraphics[width=\linewidth]{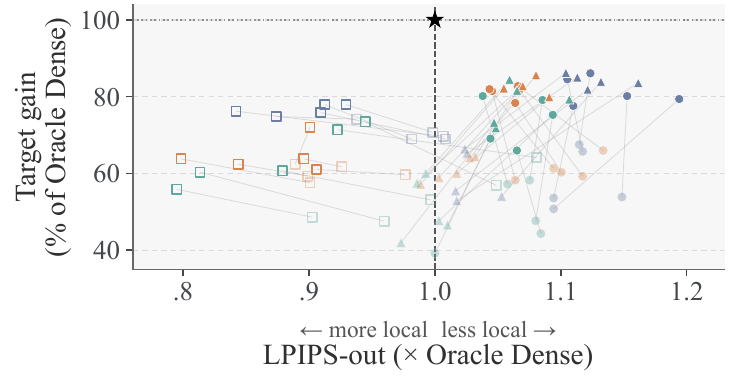}
  \end{subfigure}\par
  \vspace{4pt}%
  \begin{subfigure}[t]{\linewidth}
    \caption{Target-gain improvement}\label{fig:steering-c}
    \includegraphics[width=\linewidth]{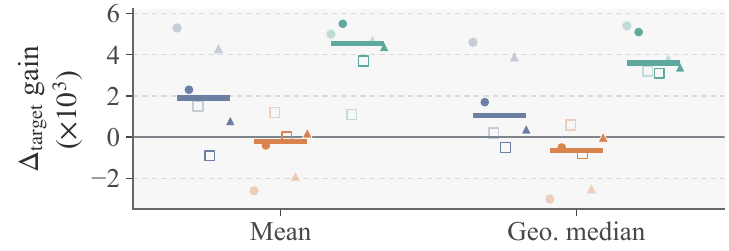}
  \end{subfigure}
\end{minipage}\par
\vspace{4pt}%
\includegraphics[width=\textwidth]{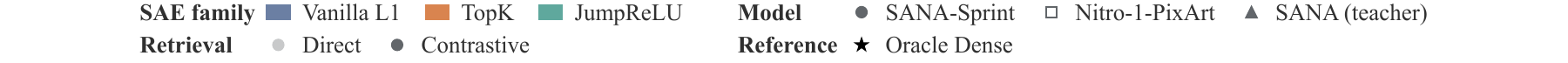}\par
\endgroup